\documentclass{article}

\usepackage{microtype}
\usepackage{graphicx}
\usepackage{subcaption}
\usepackage{booktabs} 
\usepackage{multirow}

\usepackage{hyperref}

\usepackage[accepted]{icml2025}

\usepackage{amsmath}
\usepackage{amssymb}
\usepackage{mathtools}
\usepackage{amsthm}
\usepackage{thm-restate}

\usepackage[capitalize,noabbrev]{cleveref}

\theoremstyle{plain}

\theoremstyle{definition}

\theoremstyle{remark}

\usepackage{xcolor}

\usepackage{tcolorbox}

\newcommand{\dx}{\mathrm{d}}
\newcommand{\R}{\mathbb{R}}

\newcommand{\E}{\mathbb{E}}

\newcommand{\ind}{\mathbb{I}}

\newcommand{\NN}{\mathrm{NN}}
\newcommand{\NND}{\mathrm{NND}}

\graphicspath{{../figures/}}

\icmltitlerunning{Position: All Current Generative Fidelity and Diversity Metrics are Flawed}

\begin{document}

\twocolumn[
\icmltitle{Position: All Current Generative Fidelity and Diversity Metrics are Flawed}

\icmlsetsymbol{equal}{*}

\begin{icmlauthorlist}
\icmlauthor{Ossi Räisä}{uh}
\icmlauthor{Boris van Breugel}{uc}
\icmlauthor{Mihaela van der Schaar}{uc}
\end{icmlauthorlist}

\icmlaffiliation{uh}{University of Helsinki}
\icmlaffiliation{uc}{University of Cambridge}

\icmlcorrespondingauthor{Ossi Räisä}{ossi.raisa@helsinki.fi}

\icmlkeywords{Generative model evaluation, synthetic data}

\vskip 0.3in
]



\printAffiliationsAndNotice{}  

\begin{abstract}
Any method's development and practical application is limited by our ability to measure its reliability. The popularity of generative modeling emphasizes the importance of good synthetic data metrics. Unfortunately, previous works have found many failure cases in current metrics, for example lack of outlier robustness and unclear lower and upper bounds. We propose a list of desiderata for synthetic data metrics, and a suite of sanity checks: carefully chosen simple experiments that aim to detect specific and known generative modeling failure modes. Based on these desiderata and the results of our checks, we arrive at our position: all current generative fidelity and diversity metrics are flawed. This significantly hinders practical use of synthetic data. Our aim is to convince the research community to spend more effort in developing metrics, instead of models. Additionally, through analyzing how current metrics fail, we provide practitioners with guidelines on how these metrics should (not) be used.
\end{abstract}

\section{Introduction}\label{sec:introduction}

Recent years have seen great advances in generative modeling,
from generating realistic facial 
images~\citep{karrasStyleBasedGeneratorArchitecture2019}
to generating tabular datasets with
large language models~\citep{borisovLanguageModelsAre2022}.
These advances are motivated by a promise that 
the synthetic data generated by these models could 
augment real data~\citep{dasConditionalSyntheticData2022},
maintain privacy of sensitive 
data~\citep{liewDataDistortionProbability1985,rubin1993statistical}, improve model evaluation~\citep{vanbreugelCanYouRely2023},
or improve fairness in downstream 
tasks~\citep{vanbreugelDECAFGeneratingFair2021}, among
some possible goals.

The quality of a generative model is typically evaluated 
using metrics that compare the real data with synthetic data 
produced by the model. Popular metrics include 
FID~\citep{heuselGANsTrainedTwo2017}, information divergences
like total variation distance and KL divergence, and the
performance of downstream machine learning tasks.

However, while these metrics can distinguish good and 
bad generators, they do not provide information on why 
a particular generator received a poor evaluation, or what 
can still be improved in a good generator. 
\citet{sajjadiAssessingGenerativeModels2018} proposed a 
precision/recall metric that distinguishes two failure 
modes: a generator that generates unrealistic samples
and a generator that does not cover all of the real 
distribution. These metrics were first improved 
by \citet{kynkaanniemiImprovedPrecisionRecall2019} and later
others, leading to several metrics, including 
density/coverage~\citep{naeemReliableFidelityDiversity2020}
and $\alpha$-precision/$\beta$-recall~\citep{alaaHowFaithfulYour2022}. The latter
two pairs are the most popular of these types of metrics 
today.

These metrics come in pairs. The first metrics of the pairs
measure how realistic synthetic samples are, so we 
call them \emph{fidelity} metrics. The second metrics of 
each pair measure how much of the real distribution the 
synthetic distribution covers, so we call them 
\emph{diversity} metrics.

While evaluating and improving generative models has received much 
attention, evaluating and improving the metrics themselves has received 
much less. Some recent works have found failure cases with the established 
fidelity and diversity metrics~\citep{cheemaPrecisionRecallCover2023,khayatkhoeiEmergentAsymmetryPrecision2023,parkProbabilisticPrecisionRecall2023}, and proposed
new metrics that fix the discovered problems. However, each of these
works only looks at a small number of problems and focuses on fixing those.
It remains an open question whether the fixed metrics suffer from 
the failure cases reported by the other works, or if the fixes cause new 
problems to emerge.

We aim to answer this question with a thorough evaluation of fidelity and
diversity metrics, consisting of three main contributions:
\begin{enumerate}
    \item We propose a list of six desiderata that any synthetic 
    data metric should fulfill in \Cref{sec:evaluation-metric-qualities}.
    \item We distill each failure case that has been reported in the 
    literature into a simple \emph{sanity check}, with precisely defined and automatically checked passing criteria, that are 
    linked to the desiderata, in \Cref{sec:sanity-checks}. 
    Our implementation code is available.\footnote{\url{https://github.com/vanderschaarlab/position-fidelity-diversity-metrics-flawed}}
    \item We evaluate whether each metric passes each check, and discuss the results in \Cref{sec:discussion}.
\end{enumerate}
We also add some novel
sanity checks that focus on tabular data, which is an important domain
where these metrics are used~\citep{kotelnikovTabDDPMModellingTabular2023,zhangGenerativeTablePretraining2023}, but has been neglected by existing work on metrics.

The results of our evaluation in Tables~\ref{table:passes-fails-fidelity}
and \ref{table:passes-fails-diversity} lead to our position. All of the 
fidelity and diversity metrics fail a large number of sanity checks,
in many cases failing to measure even the basic property that they are 
supposed to measure,
which we argue means that \textbf{all current generative fidelity and 
diversity metrics are flawed.}

We argue that our position has two main takeaways:
\begin{enumerate}
    \item Practitioners using fidelity and diversity metrics to evaluate
    synthetic data must be wary of what the metrics they use 
    can measure, and what they fail to measure.
    \item New metrics are needed to fix the failure cases in the current ones.
    These new metrics must be evaluated in a wide range of scenarios
    to uncover potential failure cases, including a benchmark of sanity checks
    like the one we present.
\end{enumerate}

The first takeaway is a consequence of the clear
unreliability of all metrics in our sanity checks.
A metric suffering from a failure case that is present in a given setting does not give useful
information on synthetic data quality. This is complicated by the 
complexities of real data: while our sanity checks each look at one 
potential failure case at a time, a real dataset will likely have multiple
potential failure cases at the same time. If a metric fails at 
one of these cases, it can also fail at the others, even if it 
does not fail on those in isolation.

The second takeaway follows from the first and the 
usefulness of the information fidelity and diversity metrics aim to measure. This information is useful for generative model evaluation, as evidenced 
by their wide-scale adoption in the 
community~\citep{pearceImitatingHumanBehaviour2022,anciukeviciusRenderDiffusionImageDiffusion2023,kotelnikovTabDDPMModellingTabular2023,qianSynthcityBenchmarkFramework2023,zhangGenerativeTablePretraining2023}.
Since the current metrics we have are not completely reliable, as the 
generative model research community, we must strive to develop 
reliable alternatives that can be used without worrying about a 
minefield of failure cases.

\subsection{Related Work}
\citet{borjiProsConsGAN2019,borjiProsConsGAN2022,xuEmpiricalStudyEvaluation2018} survey evaluation
metrics with a focus on GANs in the image domain. 
\citet{borjiProsConsGAN2019,xuEmpiricalStudyEvaluation2018} give lists of desiderata, which
partially overlap with our list in 
Section~\ref{sec:evaluation-metric-qualities}, and ask
which metrics fulfill each of the desiderata. However, their desiderata
focus on GANs, and demand that a single metric distinguishes 
several generator qualities like fidelity, diversity, and overfitting at 
the same time. 
We consider it sufficient for a metric to focus on a single quality at a time.
\citet{theisNoteEvaluationGenerative2016} looks at 
several classical metrics and finds that they can
result in conflicting and undesirable evaluations. 
\citet{theisPositionWhatMakes2024} considers what properties a fidelity,
or ``realism'' in their terminology, metric should have, and gives some
theoretical insight on how a metric with those properties could be 
implemented.

\section{Desiderata for Synthetic Data Evaluation Metrics}\label{sec:evaluation-metric-qualities}

Before we can evaluate how good different metrics are, we must 
establish what we want out of a metric. We use the lists given
by \citet{borjiProsConsGAN2019,xuEmpiricalStudyEvaluation2018}
as a starting point, but we do not require a single metric to 
evaluate multiple aspects of synthetic data quality. While we 
focus on fidelity and diversity metrics in our evaluation, we 
take a wider perspective in our desiderata list, and include
goals that other types of metrics may aim to meet.

Our desiderata for synthetic data metrics are:
\begin{enumerate}
    \item[D1] (purpose) One of the following: 
    \begin{enumerate}
        \item measure a quantity that is of direct practical interest, 
        \item give interpretable information on the difference between real and generated data,
        \item be a proxy-metric that is correlated
        with an impractical metric of interest.
    \end{enumerate}
    \item[D2] (hyperparameters) Have a minimal number of hyperparameters. The effect of hyperparameters
    should be as clear as possible.
    \item[D3] (data) Require less data than what is
    available in the problem of interest.
    \item[D4] (bounds) Have clear lower and upper bounds.
    \item[D5] (invariance) Be invariant to transformations of both real and synthetic data that
    do not affect data quality in the domain
    of interest.
    \item[D6] (computation) Be computationally efficient.
\end{enumerate}
These are not in any sort of priority order.

Examples of D1a (purpose) are the accuracy of predictive models or statistical inference
on synthetic data. Fidelity and diversity metrics that are the 
focus of this paper are examples of D1b.
Many works evaluate metrics by looking at D1c, for example
by comparing the metric values with human 
evaluations~\citep{salimansImprovedTechniquesTraining2016,jayasumanaRethinkingFIDBetter2024}. 
We include the condition that the target metric
is impractical in some way, or at least less practical than 
the proxy metric, since otherwise there is no need for the proxy metric. If the 
target metric can be practically computed, a practitioner could simple compute the 
target metric instead of a proxy metric that is merely correlated with it.

D2 (hyperparameters) aims to maximise the objectivity of synthetic data evaluation. A metric with 
many hyperparameters with unclear effects may have many seemingly equivalent choices 
for the hyperparameters, which nevertheless lead to different estimations of 
synthetic data quality. Hyperparameters with a standard value,
or method of choosing the value, should not be counted here,
since a default choice forbids the tweaking of the hyperparameter
to obtain a desired result from a comparison.

For D3 (data), an ideal metric would be completely invariant to the 
size of the real data, but this is an unreasonable
requirement for very small dataset sizes. Instead, we want the 
metric to be approximately invariant after some size, which we call
the required size. The required size should of course
be smaller than the amount of data that is available
in a given problem of interest. For this paper,
we set a threshold of 1000 datapoints, which is
small enough to cover the majority of datasets\footnote{
As of 29th of January, 2025, 55\% of datasets on
\url{https://openml.org/} have $> 1000$ datapoints.
}, but large enough for many statistical asymptotics 
to kick in.
Note that the size of the synthetic dataset is separate
from the real data, though the two are often set to be identical. 
Since the only constraint for the size of the 
synthetic data is computational, we consider it a hyperparameter
of the metric, and place it under desiderata D2 (hyperparameters).

D4 (bounds) allows gauging how good or bad the synthetic 
data is in absolute terms, not only in comparison
to other synthetic datasets. This is important,
since it is possible that in a hard setting,
even the best synthetic data is of poor 
quality.

The transformations that do not affect
data quality in D5 (invariance) are domain specific. 
For images, examples are small translation and 
rotations~\citep{borjiProsConsGAN2019}.
For tabular data, we can 
identify three types of transformation that clearly
do not affect data quality: scaling numerical
variables, permuting the categories of nominal discrete variables, and permuting the
order of variables. The latter two are clearly irrelevant transformations, since they 
are changes to orderings that are arbitrarily chosen in the first place. Scaling
a numerical attribute is equivalent to changing units. Changes of units
in both real and synthetic data do not change the quality of synthetic data, so
they should not change the values of evaluation metrics either.

A minimum threshold of computational efficiency for D6 (computation) is tractability: the 
metric should be possible to compute in reasonable time.
Of course, computational efficiency is always  useful for a metric, but for 
tractable  metrics, the other desiderata should be prioritised over pure computational
savings. 

\section{Evaluated Metrics}

\paragraph{Inclusion Criteria} We include metrics in our evaluation that:
\begin{enumerate}
    \item measure fidelity or diversity,
    \item produce a single number and
    \item run in reasonable time.
\end{enumerate}
Requirement (2) excludes the curve-valued metric
of \citet{sajjadiAssessingGenerativeModels2018} and improvements to 
it~\citep{simonRevisitingPrecisionRecall2019,djolongaPrecisionRecallCurvesUsing2020,siryTheoreticalEquivalenceSeveral2023,sykesUnifyingExtendingPrecision2024}. We only include single number metrics since they 
are easier to interpret and display than curve-values ones,
and are much more common in 
practice ~\citep{pearceImitatingHumanBehaviour2022,anciukeviciusRenderDiffusionImageDiffusion2023,kotelnikovTabDDPMModellingTabular2023,qianSynthcityBenchmarkFramework2023,zhangGenerativeTablePretraining2023}.
We also looked at the pair of metrics from \citet{kimTopPRRobustSupport2023} that 
are based on topological properties, but they took too much time to compute a single
evaluation.\footnote{
Specifically, computing the metrics of \citet{kimTopPRRobustSupport2023} for 1000 real 
and synthetic samples of 2-dimensional Gaussian distributions with the original implementation
did not finish in
30 minutes on an M1 MacBook Air. Computing all of I-Prec, I-Rec, density and coverage in the 
same setting took less than 0.1s.
}
We list the metrics that we include in \Cref{table:metrics}. 
We describe each metric in 
detail and list the implementations we used in \Cref{sec:metric-details}.

\begin{table*}
    \caption{Metrics in this work.}
    \label{table:metrics}
    \vskip 0.10in
    \centering
    \begin{tabular}{llr}
    \toprule
         Paper & Fidelity Metric & Diversity Metric \\
    \midrule
         \citep{kynkaanniemiImprovedPrecisionRecall2019} & Improved Precision (I-Prec) & Improved Recall (I-Rec) \\
         \citep{naeemReliableFidelityDiversity2020} & Density & Coverage \\
         \citep{alaaHowFaithfulYour2022} & Integrated $\alpha$-precision (IAP) & Integrated $\beta$-recall (IBR) \\
         \citep{cheemaPrecisionRecallCover2023} & Precision Cover (C-Prec) & Recall Cover (C-Rec) \\
         \citep{khayatkhoeiEmergentAsymmetryPrecision2023} & Symmetric Precision (symPrec) & Symmetric Recall (symRec) \\
         \citep{parkProbabilisticPrecisionRecall2023} & Probabilistic Precision (P-Prec) & Probabilistic Recall (P-Rec) \\
    \bottomrule
    \end{tabular}
    \vspace{-2mm}
\end{table*}

\paragraph{Embeddings}
All of the metrics we consider take a set of real data $X_r \sim P_r$ and a set of 
synthetic data $X_g \sim P_g$. They first embed $X_r$ and $X_g$ into a 
space with more suitable geometry than the original data space.
We denote the embedding of a single real or synthetic datapoint as 
$\phi_r$ or $\phi_g$, respectively, and denote the embeddings of the whole
real or synthetic datasets as $\Phi_r$ and $\Phi_g$, respectively.

For image datasets, the embedding is typically a pretrained neural 
network~\citep{kynkaanniemiImprovedPrecisionRecall2019,naeemReliableFidelityDiversity2020},
or a randomly initialised CNN~\citep{naeemReliableFidelityDiversity2020}.
For tabular data, \citet{alaaHowFaithfulYour2022} use a neural network embedding
that is trained on the real data. With the other metrics, we will use a simpler 
embedding that is appropriate for tabular data: we one-hot encode 
categorical variables, and normalise numerical 
variables to mean zero and unit standard deviation on the real
data.\footnote{We also apply this embedding before the neural network of \citet{alaaHowFaithfulYour2022} for $\alpha$-precision and $\beta$-recall.} 

This simple embedding is motivated by our 
desiderata D5 (invariance) and D2 (hyperparameters). Normalising numerical variables ensures that 
the resulting metric in invariant to scaling. One-hot encoding categorical
variables ensures that the metric is invariant to permuting the order
of categories, as long as the metric is invariant to permuting the components
of the embedding vector $\phi$, which is the case for all of the metrics we look
at. This also ensures that the metric is invariant to permuting the order of 
variables in the real data, so this embedding ensures invariance to all of the 
tabular data transformations that clearly do not affect data quality we discussed in
Section~\ref{sec:evaluation-metric-qualities}. This embedding also does not have
hyperparameters. In contrast, all architecture and training choices of 
a neural network embedding become hyperparameters of the resulting metric, unless there is a standard
choice of them that is widely used.

\section{Sanity Checks}\label{sec:sanity-checks}

\begin{table*}
\caption{
    Sanity Checks considered in this work. The Tab. column marks checks intended for 
    only tabular data, and the Fig. column contains the 
    figure number of the full results in the Appendix. The results are summarised in 
    \Cref{table:passes-fails-fidelity,table:passes-fails-diversity}.
    The checks are grouped as in \Cref{sec:sanity-check-summaries}.
}
\label{table:pathological-cases}
\vskip 0.15in
\centering
\begin{scriptsize}
\begin{sc}
\begin{tabular}{lcccr}
\toprule
Setting & Desiderata & Tab. & Fig. & Papers \\
\midrule
Gaussian Mean Difference & D1b, D4 & & \ref{fig:gaussian-mean-difference} & \citep{naeemReliableFidelityDiversity2020,alaaHowFaithfulYour2022,cheemaPrecisionRecallCover2023} \\
Gaussian Mean Difference + Outlier & D1b, D4 & & \ref{fig:gaussian-mean-difference-with-real-outlier}, \ref{fig:gaussian-mean-difference-with-syn-outlier} & \citep{naeemReliableFidelityDiversity2020,alaaHowFaithfulYour2022,parkProbabilisticPrecisionRecall2023} \\
Gaussian Std. Deviation Difference & D1b, D4 & & \ref{fig:gaussian-std-difference} & \citep{parkProbabilisticPrecisionRecall2023} \\
One Disjoint Dim. + Many Identical Dim. & D1b, D4 & & \ref{fig:gaussian-high-dim-one-disjoint-dim} & This work \\
Scaling One Dimension & D4, D5 & & \ref{fig:gaussian-scaling-one-dimension} & This work \\
\cmidrule[\lightrulewidth](lr){1-1}
Mode Collapse & D1b, D4 & & \ref{fig:one-vs-two-modes} & \citep{alaaHowFaithfulYour2022} \\
Mode Dropping + Invention & D1b, D4 & & \ref{fig:mode-dropping-invention} & \citep{kynkaanniemiImprovedPrecisionRecall2019} \\
Sequential / Simultaneous Mode dropping & D1b, D4 & & \ref{fig:mode-dropping-sequential}, \ref{fig:mode-dropping-simultaneous} & \citep{naeemReliableFidelityDiversity2020} \\
\cmidrule[\lightrulewidth](lr){1-1}
Hypercube, Varying Sample Size & D3, D1b & & \ref{fig:uniform-hypercube-varying-dataset-size} & \citep{cheemaPrecisionRecallCover2023} \\
Hypercube, Varying Syn. Size & D2, D1b & & \ref{fig:uniform-hypercube-varying-syn-dataset-size} & \citep{cheemaPrecisionRecallCover2023} \\
\cmidrule[\lightrulewidth](lr){1-1}
Hypersphere Surface & D1b, D4 & & \ref{fig:uniform-hypersphere-surface} & \citep{khayatkhoeiEmergentAsymmetryPrecision2023} \\
\cmidrule[\lightrulewidth](lr){1-1}
Sphere vs. Torus & D1b, D4 & & \ref{fig:sphere-torus} & \citep{cheemaPrecisionRecallCover2023} \\
\cmidrule[\lightrulewidth](lr){1-1}
Discrete Num. vs. Continuous Num. & D1b, D4 & \checkmark & \ref{fig:discrete-numerical-vs-continuous-numerical} & This work \\
Gaussian Mean Difference + Pareto & D1b, D4 & \checkmark & \ref{fig:gaussian-mean-difference-with-pareto} & This work \\
\bottomrule
\end{tabular}
\end{sc}
\end{scriptsize}
\vskip -0.1in
\vspace{-1mm}
\end{table*}

To evaluate the failure cases each metric may or may not have, we use
simple sanity checks, with artificial real and synthetic
distributions. This allows us to focus on one potential problem in each check,
and also allows evaluating extreme scenarios where we know that a metric
meeting D4 (bounds) should have a value of 0 or 1. Most of our checks
are from existing work, but we do not always use identical setups in these
cases. A handful of checks are new to this work.
\Cref{table:pathological-cases} lists the sanity checks we look at 
and previous works using them. \Cref{sec:sanity-check-details}
provides detailed descriptions of each check and the setup we use.
Next, we very briefly summarise each check.

\subsection{Sanity Check Summaries}\label{sec:sanity-check-summaries}

\paragraph{Gaussian Checks}
To evaluate how metrics distinguish simple differences 
between distributions, we have five sanity checks looking at 
two otherwise identical Gaussian distributions that differ in 
one way. In the first three checks, the difference is
either their mean, their standard deviation, or their mean and 
the presence of an outlier. In the fourth and fifth check, the 
difference is their mean in one dimension. In the fourth check,
there is one additional identical dimension, which is scaled,
testing whether the metrics are invariant to the scaling.
In the fifth check, there are many additional identical 
dimensions, testing whether metrics can find the difference 
in only one of the dimensions.

\paragraph{Gaussian Mixture Checks}
These checks evaluate how the metrics see mode collapse, invention
and dropping. One check drops 9 out of 10 modes either one mode 
at a time, or by reducing the weight of all 9 dropped modes
simultaneously. A second check evaluates both mode dropping
and invention by increasing the number of modes in the synthetic
distribution, first including modes in the real data, and later
inventing new modes. A third check evaluates mode collapse using
a real distribution with two modes, and a synthetic distribution
with one wide component that covers both real modes.

\paragraph{Varying Dataset Sizes on Uniform Hypercubes}
These checks evaluate the effect of both the real and 
synthetic dataset size on two uniform distributions on 
partially overlapping hypercubes. One experiment varies 
both $|\Phi_r| = |\Phi_g|$, while one fixes $|\Phi_r|$ and
varies $|\Phi_g|$.
\citet{cheemaPrecisionRecallCover2023} 
argue that the theoretically correct values for fidelity and diversity metrics in 
this case are the volume of the overlapping space, which we have set to 0.2.
For D1b (purpose), we check that the metrics are close to this value, and for 
D2 (hyperparameters) or D3 (data), we check that the metrics simply converge after 1000 datapoints.

\paragraph{Uniform Hypersphere Surface Check}
This check looks at the problem that inspired the symPrecision
and symRecall metrics~\citep{khayatkhoeiEmergentAsymmetryPrecision2023}.
Both real and synthetic distributions are uniform on the surface 
of a hypersphere, with different radii. The distributions are 
disjoint, so all metrics should return low values, but 
\citet{khayatkhoeiEmergentAsymmetryPrecision2023} observed that 
in high dimensions, many metrics only behave correctly when 
the synthetic radius is either smaller or larger than the real 
radius, but not in the opposite case.

\paragraph{Torus vs. Sphere Check}
This check from \citet{cheemaPrecisionRecallCover2023} 
evaluates metrics on distributions with non-trivial
geometry. One of the distributions is a uniform distribution 
on a sphere, and the other is on a torus that surrounds the 
sphere, but is disjoint from it. See \Cref{fig:sphere-torus-samples}
in the Appendix for an illustration.
We vary which of the two 
distributions is considered real and which is considered synthetic.

\paragraph{Tabular Data Focused Checks}
We include two sanity checks that focus on issues found in tabular
data. The first check repeats the Gaussian mean difference 
check, but with an additional Pareto-distributed random
variable with an identical distribution in real and 
synthetic data. This tests how the metrics function in the 
presence of a heavy-tailed power law distribution, which are 
common in tabular data. The second check has data from a 
Gaussian distribution, and data from the same Gaussian that has 
been rounded to an integer, varying the scale of both 
distributions. This evaluates whether the 
metrics can distinguish a discrete numerical distribution from a 
continuous one, which is important since integer-valued
variables are common in tabular data.

\subsection{Success Criteria}\label{sec:success-criteria}
We evaluate the success or failure of each metric on the sanity 
checks with precisely defined criteria that we check 
programmatically. We detail the criteria of each check in
\Cref{sec:sanity-check-details}. 

\looseness=-1
In general, we 
connect the criteria to one of desiderata D1b (purpose), D2 (hyperparameters), D3 (data), D4 (bounds) or D5 (invariance),
depending on the check. D1b is evaluated in almost 
every check, with criteria that require the metric to generally
behave correctly, but does not require
specific values in most cases. D4 is also evaluated
in almost every check, with criteria checking that
the metrics have values close to 0 or 1 in extreme cases. We include
these more difficult criteria because making absolute 
evaluations of generative models, like ``this model is 
good'' requires a metric with clear bounds that the metric 
actually follows in practice. D3 is evaluated in 
the two checks looking at dataset sizes, and requires that the 
metrics behave consistently after 1000 datapoints.
D5 is evaluated in the scale invariance check, and 
checks whether the metrics are scale invariant.
We do not include specific checks for D6 (computation),
since all of the metrics we include are computationally tractable, so they pass the 
only sanity check that would make sense for D6.

\paragraph{High and Low Diversity Metrics}
In some cases for diversity metrics, it can 
be argued that both high and low values are 
acceptable results, depending on how ``covering''
the real distribution is interpreted. These are
cases where the synthetic distribution covers the real 
distribution, but is so wide that the probability (density) of getting
a value from the synthetic distribution with high probability (density)
under the real distribution is very low.
An example of a sanity check like this is the 
two Gaussians with different standard deviations,
with the synthetic standard deviation being much 
larger.
On one hand, the synthetic distribution completely covers the 
real distribution, which suggests
that diversity metrics should be high. On the other hand, the probability
of sampling a synthetic datapoint in the high-density region of the real 
distribution is low, which suggests that diversity metrics should have low
values. We designate diversity metrics that adhere to the former
argument as ``high'', and designate metrics following the latter argument as
``low''. 
We require each metric to be consistent with ``low'' or ``high'' across
different variations of a sanity check.
We make these explicit in the results, so if only one
of these viewpoints is valid in a given application,
it is clear which metrics are suited to that 
application.

\begin{table*}[ht]
    \caption{Passes (T) and fails (F) of each fidelity metric on each sanity check. 
    The Tab. column marks checks intended for only tabular data.}
    \label{table:passes-fails-fidelity}
    \vskip 0.15in 
    \centering
    \resizebox{\linewidth}{!}{
    \begin{small}
        \begin{tabular}{lllllllll}
\toprule
 &  & Tab. & I-Prec & Density & IAP & C-Prec & symPrec & P-Prec \\
Desiderata & Sanity Check &  &  &  &  &  &  &  \\
\midrule
\multirow[c]{14}{*}{D1b (purpose)} & Discrete Num. vs. Continuous Num. & \checkmark & \color{red} F & \color{red} F & \color{red} F & \color{red} F & \color{red} F & \color{red} F \\
 & Gaussian Mean Difference &   & T & T & T & T & T & T \\
 & Gaussian Mean Difference + Outlier &   & \color{red} F & T & T & T & \color{red} F & T \\
 & Gaussian Mean Difference + Pareto & \checkmark & T & T & T & T & T & T \\
 & Gaussian Std. Deviation Difference &   & T & T & \color{red} F & \color{red} F & \color{red} F & T \\
 & Hypercube, Varying Sample Size &   & \color{red} F & \color{red} F & \color{red} F & \color{red} F & \color{red} F & \color{red} F \\
 & Hypercube, Varying Syn. Size &   & \color{red} F & \color{red} F & \color{red} F & \color{red} F & \color{red} F & \color{red} F \\
 & Hypersphere Surface &   & \color{red} F & \color{red} F & T & \color{red} F & T & \color{red} F \\
 & Mode Collapse &   & T & T & T & T & T & T \\
 & Mode Dropping + Invention &   & T & T & \color{red} F & \color{red} F & \color{red} F & T \\
 & One Disjoint Dim. + Many Identical Dim. &   & \color{red} F & \color{red} F & \color{red} F & \color{red} F & \color{red} F & \color{red} F \\
 & Sequential Mode Dropping &   & T & T & \color{red} F & \color{red} F & \color{red} F & T \\
 & Simultaneous Mode Dropping &   & T & \color{red} F & \color{red} F & \color{red} F & \color{red} F & T \\
 & Sphere vs. Torus &   & T & T & T & \color{red} F & T & T \\
\cline{1-9}
D2 (hyperparameters) & Hypercube, Varying Syn. Size &   & T & T & T & \color{red} F & \color{red} F & T \\
\cline{1-9}
D3 (data) & Hypercube, Varying Sample Size &   & \color{red} F & \color{red} F & \color{red} F & \color{red} F & \color{red} F & \color{red} F \\
\cline{1-9}
\multirow[c]{13}{*}{D4 (bounds)} & Discrete Num. vs. Continuous Num. & \checkmark & \color{red} F & \color{red} F & \color{red} F & \color{red} F & \color{red} F & \color{red} F \\
 & Gaussian Mean Difference &   & \color{red} F & T & \color{red} F & T & \color{red} F & \color{red} F \\
 & Gaussian Mean Difference + Outlier &   & \color{red} F & \color{red} F & \color{red} F & T & \color{red} F & \color{red} F \\
 & Gaussian Mean Difference + Pareto & \checkmark & \color{red} F & T & T & T & T & \color{red} F \\
 & Gaussian Std. Deviation Difference &   & \color{red} F & \color{red} F & \color{red} F & \color{red} F & \color{red} F & \color{red} F \\
 & Hypersphere Surface &   & \color{red} F & \color{red} F & \color{red} F & \color{red} F & T & \color{red} F \\
 & Mode Collapse &   & \color{red} F & T & \color{red} F & T & \color{red} F & \color{red} F \\
 & Mode Dropping + Invention &   & T & T & \color{red} F & \color{red} F & \color{red} F & \color{red} F \\
 & One Disjoint Dim. + Many Identical Dim. &   & \color{red} F & \color{red} F & T & \color{red} F & \color{red} F & \color{red} F \\
 & Scaling One Dimension &   & \color{red} F & T & T & T & T & T \\
 & Sequential Mode Dropping &   & \color{red} F & T & \color{red} F & \color{red} F & \color{red} F & \color{red} F \\
 & Simultaneous Mode Dropping &   & \color{red} F & T & \color{red} F & \color{red} F & \color{red} F & \color{red} F \\
 & Sphere vs. Torus &   & T & T & \color{red} F & \color{red} F & T & T \\
\cline{1-9}
D5 (invariance) & Scaling One Dimension &   & \color{red} F & T & T & T & T & T \\
\cline{1-9}
\bottomrule
\end{tabular}

    \end{small}
    }
\end{table*}

\subsection{Results}
The passes and fails of each metric on each check are shown in 
\Cref{table:passes-fails-fidelity} for fidelity metrics and 
\Cref{table:passes-fails-diversity} for diversity metrics.
Plots showing the full results are included in 
\Cref{sec:sanity-check-figures}. We discuss the implications 
of these results in the next section.

\begin{table*}[ht]
    \caption{Passes (T, H or L) and fails (F) of each diversity metric on each sanity check. H and L refer to high and low metric types described in 
    Section~\ref{sec:success-criteria}. The Tab. column marks checks intended for 
    only tabular data.}
    \label{table:passes-fails-diversity}
    \vskip 0.15in 
    \centering
    \resizebox{\linewidth}{!}{
    \begin{small}
        \begin{tabular}{lllllllll}
\toprule
 &  & Tab. & I-Rec & Coverage & IBR & C-Rec & symRec & P-Rec \\
Desiderata & Sanity Check &  &  &  &  &  &  &  \\
\midrule
\multirow[c]{14}{*}{D1b (purpose)} & Discrete Num. vs. Continuous Num. & \checkmark & \color{red} F & \color{red} F & \color{red} F & \color{red} F & \color{red} F & \color{red} F \\
 & Gaussian Mean Difference &   & T & T & T & T & T & T \\
 & Gaussian Mean Difference + Outlier &   & \color{red} F & T & T & T & T & T \\
 & Gaussian Mean Difference + Pareto & \checkmark & T & T & T & T & T & T \\
 & Gaussian Std. Deviation Difference &   & H & \color{red} F & L & \color{red} F & \color{red} F & H \\
 & Hypercube, Varying Sample Size &   & \color{red} F & \color{red} F & \color{red} F & \color{red} F & \color{red} F & \color{red} F \\
 & Hypercube, Varying Syn. Size &   & \color{red} F & \color{red} F & \color{red} F & \color{red} F & \color{red} F & \color{red} F \\
 & Hypersphere Surface &   & \color{red} F & \color{red} F & \color{red} F & \color{red} F & T & \color{red} F \\
 & Mode Collapse &   & \color{red} F & \color{red} F & \color{red} F & \color{red} F & \color{red} F & \color{red} F \\
 & Mode Dropping + Invention &   & T & \color{red} F & \color{red} F & \color{red} F & \color{red} F & T \\
 & One Disjoint Dim. + Many Identical Dim. &   & \color{red} F & \color{red} F & \color{red} F & \color{red} F & \color{red} F & \color{red} F \\
 & Sequential Mode Dropping &   & T & T & \color{red} F & \color{red} F & T & T \\
 & Simultaneous Mode Dropping &   & \color{red} F & T & \color{red} F & T & \color{red} F & T \\
 & Sphere vs. Torus &   & \color{red} F & \color{red} F & T & \color{red} F & \color{red} F & \color{red} F \\
\cline{1-9}
D2 (hyperparameters) & Hypercube, Varying Syn. Size &   & \color{red} F & \color{red} F & \color{red} F & \color{red} F & \color{red} F & \color{red} F \\
\cline{1-9}
D3 (data) & Hypercube, Varying Sample Size &   & \color{red} F & \color{red} F & \color{red} F & \color{red} F & \color{red} F & \color{red} F \\
\cline{1-9}
\multirow[c]{13}{*}{D4 (bounds)} & Discrete Num. vs. Continuous Num. & \checkmark & \color{red} F & \color{red} F & \color{red} F & \color{red} F & \color{red} F & \color{red} F \\
 & Gaussian Mean Difference &   & \color{red} F & T & \color{red} F & T & \color{red} F & \color{red} F \\
 & Gaussian Mean Difference + Outlier &   & \color{red} F & T & \color{red} F & T & \color{red} F & \color{red} F \\
 & Gaussian Mean Difference + Pareto & \checkmark & T & T & \color{red} F & T & T & \color{red} F \\
 & Gaussian Std. Deviation Difference &   & \color{red} F & \color{red} F & \color{red} F & \color{red} F & \color{red} F & \color{red} F \\
 & Hypersphere Surface &   & \color{red} F & \color{red} F & \color{red} F & \color{red} F & T & \color{red} F \\
 & Mode Collapse &   & \color{red} F & T & \color{red} F & T & \color{red} F & \color{red} F \\
 & Mode Dropping + Invention &   & T & \color{red} F & \color{red} F & \color{red} F & \color{red} F & \color{red} F \\
 & One Disjoint Dim. + Many Identical Dim. &   & \color{red} F & \color{red} F & T & \color{red} F & \color{red} F & \color{red} F \\
 & Scaling One Dimension &   & T & T & T & T & T & T \\
 & Sequential Mode Dropping &   & \color{red} F & T & \color{red} F & T & \color{red} F & \color{red} F \\
 & Simultaneous Mode Dropping &   & \color{red} F & T & \color{red} F & T & \color{red} F & \color{red} F \\
 & Sphere vs. Torus &   & T & \color{red} F & T & T & \color{red} F & T \\
\cline{1-9}
D5 (invariance) & Scaling One Dimension &   & \color{red} F & T & T & T & T & T \\
\cline{1-9}
\bottomrule
\end{tabular}

    \end{small}
    }
\end{table*}

\section{Discussion}\label{sec:discussion}
In this section, we discuss the main takeaways of our results.
We start by looking at practical considerations in \Cref{sec:practical-implications}, and consider future research
directions in \Cref{sec:future-research}. 

\subsection{Implications for Practitioners}\label{sec:practical-implications}
Next, we go through the most important practical takeaways
from our results. We present a checklist summarising these, along with more specific takeaways,
in \Cref{sec:practical-advice-checklist}.

\paragraph{No Metric is Suitable for Absolute Evaluations}
By absolute evaluation of a generative model, we mean answering 
the question ``Is this model good / bad?''. Research papers
often focus on relative evaluation, that is the question
``Which model is the best?'', however, practical deployments
require a model that is good in absolute terms, not just relative
to other models. 

Absolute evaluation requires a metric with 
clearly defined lower and upper bounds, which can be chosen 
to be 0 and 1 without loss of generality. While most of the 
metrics we examine have these bounds in theory, our results 
in checks of D4 (bounds) show
that they often do not have them in practice. For example,
in \crefrange{fig:gaussian-mean-difference}{fig:gaussian-std-difference} with $d = 64$, an integrated $\alpha$-precision
of around 0.5 is the best one can get, but with $d < 64$,
0.5 indicates bad synthetic data.

Another example is density, which behaves 
strangely in some cases. In most settings, the best density 
is close to 1, but in 
\Cref{fig:uniform-hypersphere-surface,fig:gaussian-std-difference}
in high dimensions, density goes up to around 200. However, 
in these cases with identical distributions, density is around 1,
so the high values do not correspond to better synthetic data.

\paragraph{Investigate Effect of Real Dataset Size on Evaluation}
The results of our D3 (data) check in 
\Cref{fig:uniform-hypercube-varying-dataset-size} show that all 
metrics are affected by the dataset size, many even at 
sizes around 10000. In many benchmarks, there are choices that affect
the dataset size, such as the handling of missing values,
or the choice of a subset from a larger dataset. Since these choices 
also affect the metrics through the dataset size, one should investigate
how large this effect is in any given evaluation, and whether it could 
affect the conclusions of the evaluation.

\paragraph{Using Metrics with Real Data Requires Care}
A real dataset is much more complex than any of our checks,
and will likely have many of the potential failure cases that the checks
look at simultaneously. If one uses a metric that fails
in the presence of any one of these cases, the metric could 
fail as a whole and not provide any meaningful evaluation.
As a result, it if one wants to use a metric in a 
particular setting, one should be able to justify that none
of the cases the metric fails on are present in the setting,
or otherwise justify that the failures of the metric are not
a concern in the setting.

\newtcolorbox{mycolorbox}[1]{colback=black!5!white,colframe=blue!50!black,fonttitle=\bfseries,title=#1}

\begin{mycolorbox}{Takeaway for Practitioners}
Practitioners using fidelity and diversity metrics must be 
wary that: 
\begin{itemize}
    \item No metric is suitable for absolute evaluations.
    \item The effect of dataset size on metrics should be investigated
    in each evaluation.
    \item Using metrics with real data requires care, see
    checklist in \Cref{sec:practical-advice-checklist}.
\end{itemize}
\end{mycolorbox}

\subsection{Practical Advice Checklist}\label{sec:practical-advice-checklist}

This checklist summarises the main takeaways from our results for
practical applications of fidelity and diversity metrics.
The list is intended for practitioners designing an experiment evaluating generative models. For each experiment, the designer should answer the questions, and, if advice is given for their answer, take that into account in their design. 
If there are multiple
preferred metrics after answering all questions,
we encourage computing all of them to see whether they all lead to 
the same conclusion, or to different conclusions.
\begin{enumerate}
    \item Is your evaluation absolute (is the model good) or relative (which model is best)?
    \begin{itemize}
        \item Absolute: all current metrics are bad
    \end{itemize}
    \item Does the size of your evaluation dataset depend on choices
    you have made during data preparation, such as how missing data
    is handled?
    \begin{itemize}
        \item Yes: Investigate how the metrics you are using behave with different dataset sizes, and take this into account when interpreting your results.
    \end{itemize}
    \item Is detecting mode collapse / dropping / invention important?
    \begin{itemize}
        \item Yes: prefer I-Prec, P-Prec and P-Rec
    \end{itemize}
    \item Is robustness to outliers important?
    \begin{itemize}
        \item Yes: avoid I-Prec, symPrec and I-Rec
    \end{itemize}
    \item Is it important to find differences in one dimension among many?
    \begin{itemize}
        \item Yes: all current metrics are bad
    \end{itemize}
    \item Is it important to find a difference between continuous and discrete numerical values?
    \begin{itemize}
        \item Yes: all current metrics are bad
    \end{itemize}
    \item Is the distinguishing between distributions on close surfaces important (uniform hypersphere surface check)?
    \begin{itemize}
        \item Yes: prefer symPrec, IAP and symRec
    \end{itemize}
    \item Is it important to distinguish distributions on complex shapes, with possibly surrounding the other (sphere vs. torus check)?
    \begin{itemize}
        \item Yes: avoid C-Prec among fidelity metrics, prefer IBR among diversity metrics
    \end{itemize}
\end{enumerate}

\paragraph{Reasons for Checklist Advice}
The first two questions on the checklist are discussed in 
\Cref{sec:practical-implications}. The advice in the checklist for 
these questions summarise the recommendations from the corresponding
discussion.
For the rest of the questions, we look at relevant checks
in \Cref{table:passes-fails-fidelity,table:passes-fails-diversity}, and
recommend metrics that pass all or most relevant checks, and recommend
avoiding metrics that fail many relevant checks.
In general, we avoid giving any hard recommendations of a
particular metric due to the fact that there is no clearly
best metric in any case. 

\subsection{Hardest Sanity Checks}\label{sec:hardest-sanity-checks}
In this section, we examine the hardest sanity checks that most or 
all metrics failed, and discuss possible reasons for the failures.

\paragraph{One Disjoint Dimension}
This check is about finding a large difference in one dimension among many identical
dimensions. All metrics failed, and only $\alpha$-precision and $\beta$-recall are even
close to passing in \Cref{fig:gaussian-high-dim-one-disjoint-dim}.
We conjecture that their partial success is due to the neural network embedding they use,
which proved capable of learning an embedding where the difference in one dimension is 
apparent even with a large number of identical dimensions. The other metrics are based on Euclidean distance, where the large difference
in one dimension gets drowned out by the other dimensions.

\paragraph{Uniform Hypercube}
These checks test how the metrics behave as the sizes of the real and synthetic 
datasets changes. The real and synthetic distributions are uniform distributions
on two overlapping $d$-dimensional hypercubes. All metrics fail to even converge,
as required for D3 (data), when the sizes of both datasets change. When only
the synthetic dataset size changes, many metrics are almost invariant, or at least 
converge reasonably, but none are close to the theoretical fidelity and diversity 
values from \citet{cheemaPrecisionRecallCover2023}. We conjecture that this is due
to a non-intuitive feature of high-dimensional geometry. If the length of the overlapping
sides between the two hypercubes were fixed, the overlap volume would decrease exponentially
with dimension, so to keep the volume fixed, the overlapping length must quickly approach
1.\footnote{Both hypercubes have side length 1.} The distance between points that are
a fixed length apart in all dimensions only increases polynomially with increased
dimension, so the distances between the points in the real and synthetic hypercubes
decrease with increased dimension in this setting.

\paragraph{Hypersphere Surface}
This check evaluates a problem discovered by \citet{khayatkhoeiEmergentAsymmetryPrecision2023}
that motivated their symPrecand symRec metrics. However,
this still remains an open problem, since those are the only metrics
that pass this check, but they fail many basic D1b (purpose)
checks. Future work should consider if the insights from 
\citet{khayatkhoeiEmergentAsymmetryPrecision2023} on this problem
can be applied to a fix with other metrics.

\paragraph{Discrete vs. Continuous Numerical Variables}
This check evaluates whether the metrics can distinguish a continuous numerical
distribution from a similar rounded distribution with discrete values. This 
was especially hard for current metrics, since they all failed.
However, most of the failures happen only at a very fine discretisation,
as seen in \Cref{fig:discrete-numerical-vs-continuous-numerical}.

\subsection{Future Directions for Metrics}\label{sec:future-research}
Since all metrics failed so many sanity checks that using any of them requires 
great care, it is clear that new fidelity and diversity metrics are needed that 
avoid as many of these failure cases as possible. 

\paragraph{Going Beyond Euclidean Geometry}
Many of the hard checks we have highlighted in \Cref{sec:hardest-sanity-checks} 
have a common theme:
the check requires distinguishing things that are not easily distinguished in
Euclidean geometry. As all of the metrics we evaluated are based on Euclidean
distances in some way,\footnote{$\alpha$-precision and $\beta$-recall have learned 
non-Euclidean embedding, but still use Euclidean distances after applying the embedding.}
it is not surprising that they fail these checks. We conjecture that 
future work must go beyond Euclidean distances in some way, such as by using
topological features~\citep{barannikovManifoldTopologyDivergence2021,southernCurvatureFiltrationsGraph2023,kimTopPRRobustSupport2023},
to solve these issues.

\begin{mycolorbox}{Takeaway for Researchers}
More research on
fidelity and diversity metrics is needed. This research should be evaluated
on a wide range of sanity checks that aim to reveal as many failure cases as possible.
\end{mycolorbox}

\section{Alternative Views}\label{sec:alternative-views}
\paragraph{Sanity checks are unrealistic since they do not use real data}
All of our sanity checks use very simple distributions,
so they are toy settings that lack the complexities of 
real data. As a result, one may argue that these checks
are not representative of real settings, and are 
thus not useful to evaluate how metrics perform in 
practice.

We use simple artificial distributions in our sanity checks to 
isolate a single property that is tested in each check. This also allows
us to set up extreme scenarios where we know what value a metric should 
have to meet D4 (bounds), which would be much harder with real data.

In addition, our position is that all current metrics are flawed. 
We argue that even simple artificial sanity checks are 
enough to show this, since a metric that does not work even in many simple
settings in unlikely to work in more complex settings.

\paragraph{Sanity check x/y/z is not relevant or is flawed}
Not all of our checks are relevant in all settings, and it is 
possible that there are flaws in their design or implementation, despite 
our efforts to make sure all the tests are 
correctly implemented and their criteria make sense. 
However, all metrics
fail much more than just one check. Arguing that some metric would pass
all, or almost all, checks if irrelevant and flawed ones were removed 
would require there to be many of these removed checks, which we 
find very unlikely.

\paragraph{A metric does not need to be perfect to be useful}
It is true that an imperfect metric can be very useful, and perfect 
metrics may not even exist at all. As a result, one may 
worry that our sanity checks raise a too high of a bar:
reviewers may be inclined to reject papers proposing new
metrics if those metrics fail any checks, even though 
the new papers might have genuinely useful insights.
Setting the bar too high in this way could stall the 
development of metrics.

It is not our intention to claim
that a metric must pass all of our checks to be useful. However, users 
of metrics must 
still know if if their metrics are imperfect, and how the flaws 
could affect their use case. In addition, as we argue in 
\Cref{sec:practical-implications}, a metric that fails on any potential failure case 
that is present in a setting could cause it to not provide any useful 
information, so the many failures of all current metrics means that 
their use needs to be considered very carefully.

\section{Conclusion}
In this work, we evaluated generative fidelity and diversity metrics using simple sanity checks that we collected from the literature,
or proposed ourselves. Unlike the previous literature, we defined precise
passing criteria for each check, allowing checking the criteria 
programmatically. 

We found that all of the metrics fail many of the sanity checks, which
lead to our position: all current fidelity and diversity metrics are flawed in
one way or the other. We presented two takeaways from this position.
First, practitioners using these metrics must be wary of their flaws,
and second, more research into metrics is needed to correct the flaws
of the current ones.
We presented several alternative views to our position, and gave our 
arguments against each alternative.

We hope to encourage better informed use of fidelity and diversity metrics
in future works, and to highlight the need for new metrics with less
potential issues. We think that these will lead to more accurate 
evaluations of generative models, and help continue the great advances
in generative modeling of recent years into the future.

\section*{Acknowledgements}
OR was supported by the Researchers Abroad (Tutkijat maailmalle)-program 
(project 20240109).

\newpage

\section*{Impact Statement}

The goal of this paper is to encourage informed use of fidelity 
and diversity metrics when evaluating generative models, and the 
development of better metrics. We think the societal consequences 
of this work are likely to be positive, since naive use of poor 
metrics may lead to incorrect 
conclusions about the models the metrics are measuring.

\bibliography{Synthetic_Data_Metrics}

\begin{thebibliography}{36}
\providecommand{\natexlab}[1]{#1}
\providecommand{\url}[1]{\texttt{#1}}
\expandafter\ifx\csname urlstyle\endcsname\relax
  \providecommand{\doi}[1]{doi: #1}\else
  \providecommand{\doi}{doi: \begingroup \urlstyle{rm}\Url}\fi

\bibitem[Alaa et~al.(2022)Alaa, Breugel, Saveliev, and van~der Schaar]{alaaHowFaithfulYour2022}
Alaa, A., Breugel, B.~V., Saveliev, E.~S., and van~der Schaar, M.
\newblock How {{Faithful}} is your {{Synthetic Data}}? {{Sample-level Metrics}} for {{Evaluating}} and {{Auditing Generative Models}}.
\newblock In \emph{Proceedings of the 39th {{International Conference}} on {{Machine Learning}}}, pp.\  290--306. PMLR, 2022.

\bibitem[Anciukevi{\v c}ius et~al.(2023)Anciukevi{\v c}ius, Xu, Fisher, Henderson, Bilen, Mitra, and Guerrero]{anciukeviciusRenderDiffusionImageDiffusion2023}
Anciukevi{\v c}ius, T., Xu, Z., Fisher, M., Henderson, P., Bilen, H., Mitra, N.~J., and Guerrero, P.
\newblock {{RenderDiffusion}}: {{Image Diffusion}} for {{3D Reconstruction}}, {{Inpainting}} and {{Generation}}.
\newblock In \emph{Proceedings of the {{IEEE}}/{{CVF Conference}} on {{Computer Vision}} and {{Pattern Recognition}}}, pp.\  12608--12618, 2023.

\bibitem[Barannikov et~al.(2021)Barannikov, Trofimov, Sotnikov, Trimbach, Korotin, Filippov, and Burnaev]{barannikovManifoldTopologyDivergence2021}
Barannikov, S., Trofimov, I., Sotnikov, G., Trimbach, E., Korotin, A., Filippov, A., and Burnaev, E.
\newblock Manifold {{Topology Divergence}}: A {{Framework}} for {{Comparing Data Manifolds}}.
\newblock In \emph{Advances in {{Neural Information Processing Systems}}}, volume~34, pp.\  7294--7305. Curran Associates, Inc., 2021.

\bibitem[Borisov et~al.(2022)Borisov, Sessler, Leemann, Pawelczyk, and Kasneci]{borisovLanguageModelsAre2022}
Borisov, V., Sessler, K., Leemann, T., Pawelczyk, M., and Kasneci, G.
\newblock Language {{Models}} are {{Realistic Tabular Data Generators}}.
\newblock In \emph{The {{Eleventh International Conference}} on {{Learning Representations}}}, 2022.

\bibitem[Borji(2019)]{borjiProsConsGAN2019}
Borji, A.
\newblock Pros and cons of {{GAN}} evaluation measures.
\newblock \emph{Computer Vision and Image Understanding}, 179:\penalty0 41--65, 2019.

\bibitem[Borji(2022)]{borjiProsConsGAN2022}
Borji, A.
\newblock Pros and cons of {{GAN}} evaluation measures: {{New}} developments.
\newblock \emph{Computer Vision and Image Understanding}, 215:\penalty0 103329, 2022.

\bibitem[Cheema \& Urner(2023)Cheema and Urner]{cheemaPrecisionRecallCover2023}
Cheema, F. and Urner, R.
\newblock Precision {{Recall Cover}}: {{A Method For Assessing Generative Models}}.
\newblock In \emph{Proceedings of {{The}} 26th {{International Conference}} on {{Artificial Intelligence}} and {{Statistics}}}, pp.\  6571--6594. PMLR, 2023.

\bibitem[Das et~al.(2022)Das, Tran, Singh, Yue, Tison, {Sangiovanni-Vincentelli}, and Spanos]{dasConditionalSyntheticData2022}
Das, H.~P., Tran, R., Singh, J., Yue, X., Tison, G., {Sangiovanni-Vincentelli}, A., and Spanos, C.~J.
\newblock Conditional {{Synthetic Data Generation}} for {{Robust Machine Learning Applications}} with {{Limited Pandemic Data}}.
\newblock In \emph{Proceedings of the {{AAAI Conference}} on {{Artificial Intelligence}}}, volume~36, pp.\  11792--11800, 2022.

\bibitem[Djolonga et~al.(2020)Djolonga, Lucic, Cuturi, Bachem, Bousquet, and Gelly]{djolongaPrecisionRecallCurvesUsing2020}
Djolonga, J., Lucic, M., Cuturi, M., Bachem, O., Bousquet, O., and Gelly, S.
\newblock Precision-{{Recall Curves Using Information Divergence Frontiers}}.
\newblock In \emph{Proceedings of the {{Twenty Third International Conference}} on {{Artificial Intelligence}} and {{Statistics}}}, pp.\  2550--2559. PMLR, 2020.

\bibitem[Heusel et~al.(2017)Heusel, Ramsauer, Unterthiner, Nessler, and Hochreiter]{heuselGANsTrainedTwo2017}
Heusel, M., Ramsauer, H., Unterthiner, T., Nessler, B., and Hochreiter, S.
\newblock {{GANs Trained}} by a {{Two Time-Scale Update Rule Converge}} to a {{Local Nash Equilibrium}}.
\newblock In \emph{Advances in {{Neural Information Processing Systems}}}, volume~30, 2017.

\bibitem[Jayasumana et~al.(2024)Jayasumana, Ramalingam, Veit, Glasner, Chakrabarti, and Kumar]{jayasumanaRethinkingFIDBetter2024}
Jayasumana, S., Ramalingam, S., Veit, A., Glasner, D., Chakrabarti, A., and Kumar, S.
\newblock Rethinking {{FID}}: {{Towards}} a {{Better Evaluation Metric}} for {{Image Generation}}.
\newblock In \emph{Proceedings of the {{IEEE}}/{{CVF Conference}} on {{Computer Vision}} and {{Pattern Recognition}}}, pp.\  9307--9315, 2024.

\bibitem[Karras et~al.(2019)Karras, Laine, and Aila]{karrasStyleBasedGeneratorArchitecture2019}
Karras, T., Laine, S., and Aila, T.
\newblock A {{Style-Based Generator Architecture}} for {{Generative Adversarial Networks}}.
\newblock In \emph{Proceedings of the {{IEEE}}/{{CVF Conference}} on {{Computer Vision}} and {{Pattern Recognition}}}, pp.\  4401--4410, 2019.

\bibitem[Khayatkhoei \& Abdalmageed(2023)Khayatkhoei and Abdalmageed]{khayatkhoeiEmergentAsymmetryPrecision2023}
Khayatkhoei, M. and Abdalmageed, W.
\newblock Emergent {{Asymmetry}} of {{Precision}} and {{Recall}} for {{Measuring Fidelity}} and {{Diversity}} of {{Generative Models}} in {{High Dimensions}}.
\newblock In \emph{Proceedings of the 40th {{International Conference}} on {{Machine Learning}}}, pp.\  16326--16343. PMLR, 2023.

\bibitem[Kim et~al.(2023)Kim, Jang, Kim, and Yoo]{kimTopPRRobustSupport2023}
Kim, P.~J., Jang, Y., Kim, J., and Yoo, J.
\newblock {{TopP}}\&{{R}}: {{Robust Support Estimation Approach}} for {{Evaluating Fidelity}} and {{Diversity}} in {{Generative Models}}.
\newblock In \emph{Advances in {{Neural Information Processing Systems}}}, volume~36, pp.\  7831--7866, 2023.

\bibitem[Kotelnikov et~al.(2023)Kotelnikov, Baranchuk, Rubachev, and Babenko]{kotelnikovTabDDPMModellingTabular2023}
Kotelnikov, A., Baranchuk, D., Rubachev, I., and Babenko, A.
\newblock {{TabDDPM}}: {{Modelling Tabular Data}} with {{Diffusion Models}}.
\newblock In \emph{Proceedings of the 40th {{International Conference}} on {{Machine Learning}}}, 2023.

\bibitem[Kynk{\"a}{\"a}nniemi et~al.(2019)Kynk{\"a}{\"a}nniemi, Karras, Laine, Lehtinen, and Aila]{kynkaanniemiImprovedPrecisionRecall2019}
Kynk{\"a}{\"a}nniemi, T., Karras, T., Laine, S., Lehtinen, J., and Aila, T.
\newblock Improved {{Precision}} and {{Recall Metric}} for {{Assessing Generative Models}}.
\newblock In \emph{Advances in {{Neural Information Processing Systems}}}, volume~32, 2019.

\bibitem[Liew et~al.(1985)Liew, Choi, and Liew]{liewDataDistortionProbability1985}
Liew, C.~K., Choi, U.~J., and Liew, C.~J.
\newblock A data distortion by probability distribution.
\newblock \emph{ACM Transactions on Database Systems}, 10\penalty0 (3):\penalty0 395--411, 1985.

\bibitem[McKenna et~al.(2022)McKenna, Mullins, Sheldon, and Miklau]{mckennaAIMAdaptiveIterative2022}
McKenna, R., Mullins, B., Sheldon, D., and Miklau, G.
\newblock {{AIM}}: An adaptive and iterative mechanism for differentially private synthetic data.
\newblock \emph{Proceedings of the VLDB Endowment}, 15\penalty0 (11):\penalty0 2599--2612, 2022.

\bibitem[Naeem et~al.(2020)Naeem, Oh, Uh, Choi, and Yoo]{naeemReliableFidelityDiversity2020}
Naeem, M.~F., Oh, S.~J., Uh, Y., Choi, Y., and Yoo, J.
\newblock Reliable {{Fidelity}} and {{Diversity Metrics}} for {{Generative Models}}.
\newblock In \emph{Proceedings of the 37th {{International Conference}} on {{Machine Learning}}}, pp.\  7176--7185. PMLR, 2020.

\bibitem[Park \& Kim(2023)Park and Kim]{parkProbabilisticPrecisionRecall2023}
Park, D. and Kim, S.
\newblock Probabilistic {{Precision}} and {{Recall Towards Reliable Evaluation}} of {{Generative Models}}.
\newblock In \emph{Proceedings of the {{IEEE}}/{{CVF International Conference}} on {{Computer Vision}}}, pp.\  20099--20109, 2023.

\bibitem[Pearce et~al.(2022)Pearce, Rashid, Kanervisto, Bignell, Sun, Georgescu, Macua, Tan, Momennejad, Hofmann, and Devlin]{pearceImitatingHumanBehaviour2022}
Pearce, T., Rashid, T., Kanervisto, A., Bignell, D., Sun, M., Georgescu, R., Macua, S.~V., Tan, S.~Z., Momennejad, I., Hofmann, K., and Devlin, S.
\newblock Imitating {{Human Behaviour}} with {{Diffusion Models}}.
\newblock In \emph{The {{Eleventh International Conference}} on {{Learning Representations}}}, 2022.

\bibitem[Qian et~al.(2023)Qian, Davis, and van~der Schaar]{qianSynthcityBenchmarkFramework2023}
Qian, Z., Davis, R., and van~der Schaar, M.
\newblock Synthcity: A benchmark framework for diverse use cases of tabular synthetic data.
\newblock In \emph{Thirty-Seventh {{Conference}} on {{Neural Information Processing Systems Datasets}} and {{Benchmarks Track}}}, volume~36, 2023.

\bibitem[Rubin(1993)]{rubin1993statistical}
Rubin, D.~B.
\newblock Discussion: {{Statistical}} disclosure limitation.
\newblock \emph{Journal of Official Statistics}, 9\penalty0 (2):\penalty0 461--468, 1993.

\bibitem[Ruff et~al.(2018)Ruff, Vandermeulen, Goernitz, Deecke, Siddiqui, Binder, M{\"u}ller, and Kloft]{ruffDeepOneClassClassification2018}
Ruff, L., Vandermeulen, R., Goernitz, N., Deecke, L., Siddiqui, S.~A., Binder, A., M{\"u}ller, E., and Kloft, M.
\newblock Deep {{One-Class Classification}}.
\newblock In \emph{Proceedings of the 35th {{International Conference}} on {{Machine Learning}}}, pp.\  4393--4402. PMLR, 2018.

\bibitem[Sajjadi et~al.(2018)Sajjadi, Bachem, Lucic, Bousquet, and Gelly]{sajjadiAssessingGenerativeModels2018}
Sajjadi, M. S.~M., Bachem, O., Lucic, M., Bousquet, O., and Gelly, S.
\newblock Assessing {{Generative Models}} via {{Precision}} and {{Recall}}.
\newblock In \emph{Advances in {{Neural Information Processing Systems}}}, volume~31, 2018.

\bibitem[Salimans et~al.(2016)Salimans, Goodfellow, Zaremba, Cheung, Radford, Chen, and Chen]{salimansImprovedTechniquesTraining2016}
Salimans, T., Goodfellow, I., Zaremba, W., Cheung, V., Radford, A., Chen, X., and Chen, X.
\newblock Improved {{Techniques}} for {{Training GANs}}.
\newblock In \emph{Advances in {{Neural Information Processing Systems}}}, volume~29, 2016.

\bibitem[Simon et~al.(2019)Simon, Webster, and Rabin]{simonRevisitingPrecisionRecall2019}
Simon, L., Webster, R., and Rabin, J.
\newblock Revisiting precision recall definition for generative modeling.
\newblock In \emph{Proceedings of the 36th {{International Conference}} on {{Machine Learning}}}, pp.\  5799--5808. PMLR, 2019.

\bibitem[Siry et~al.(2023)Siry, Webster, Simon, and Rabin]{siryTheoreticalEquivalenceSeveral2023}
Siry, R., Webster, R., Simon, L., and Rabin, J.
\newblock On the {{Theoretical Equivalence}} of {{Several Trade-Off Curves Assessing Statistical Proximity}}.
\newblock \emph{Journal of Machine Learning Research}, 24\penalty0 (185):\penalty0 1--34, 2023.

\bibitem[Southern et~al.(2023)Southern, Wayland, Bronstein, and Rieck]{southernCurvatureFiltrationsGraph2023}
Southern, J., Wayland, J., Bronstein, M., and Rieck, B.
\newblock Curvature {{Filtrations}} for {{Graph Generative Model Evaluation}}.
\newblock In \emph{Advances in {{Neural Information Processing Systems}}}, volume~36, pp.\  63036--63061, 2023.

\bibitem[Sykes et~al.(2024)Sykes, Simon, and Rabin]{sykesUnifyingExtendingPrecision2024}
Sykes, B., Simon, L., and Rabin, J.
\newblock Unifying and extending {{Precision Recall}} metrics for assessing generative models.
\newblock \url{http://arxiv.org/abs/2405.01611}, 2024.

\bibitem[Theis(2024)]{theisPositionWhatMakes2024}
Theis, L.
\newblock Position: {{What}} makes an image realistic?
\newblock In \emph{Proceedings of the 41st {{International Conference}} on {{Machine Learning}}}, pp.\  48062--48076. PMLR, 2024.

\bibitem[Theis et~al.(2016)Theis, {van den Oord}, and Bethge]{theisNoteEvaluationGenerative2016}
Theis, L., {van den Oord}, A., and Bethge, M.
\newblock A note on the evaluation of generative models.
\newblock In \emph{International {{Conference}} on {{Learning Representations}}}, pp.\  1--10, 2016.

\bibitem[{van Breugel} et~al.(2021){van Breugel}, Kyono, Berrevoets, and {van der Schaar}]{vanbreugelDECAFGeneratingFair2021}
{van Breugel}, B., Kyono, T., Berrevoets, J., and {van der Schaar}, M.
\newblock {{DECAF}}: {{Generating Fair Synthetic Data Using Causally-Aware Generative Networks}}.
\newblock In \emph{Advances in {{Neural Information Processing Systems}}}, volume~34, pp.\  22221--22233, 2021.

\bibitem[{van Breugel} et~al.(2023){van Breugel}, Seedat, Imrie, and van~der Schaar]{vanbreugelCanYouRely2023}
{van Breugel}, B., Seedat, N., Imrie, F., and van~der Schaar, M.
\newblock Can {{You Rely}} on {{Your Model Evaluation}}? {{Improving Model Evaluation}} with {{Synthetic Test Data}}.
\newblock In \emph{Thirty-Seventh {{Conference}} on {{Neural Information Processing Systems}}}, volume~36, 2023.

\bibitem[Xu et~al.(2018)Xu, Huang, Yuan, Guo, Sun, Wu, and Weinberger]{xuEmpiricalStudyEvaluation2018}
Xu, Q., Huang, G., Yuan, Y., Guo, C., Sun, Y., Wu, F., and Weinberger, K.
\newblock An empirical study on evaluation metrics of generative adversarial networks.
\newblock \url{http://arxiv.org/abs/1806.07755}, 2018.

\bibitem[Zhang et~al.(2023)Zhang, Wang, Yan, Jian, and Liu]{zhangGenerativeTablePretraining2023}
Zhang, T., Wang, S., Yan, S., Jian, L., and Liu, Q.
\newblock Generative {{Table Pre-training Empowers Models}} for {{Tabular Prediction}}.
\newblock In \emph{The 2023 {{Conference}} on {{Empirical Methods}} in {{Natural Language Processing}}}, 2023.

\end{thebibliography}
\bibliographystyle{icml2025}

\newpage
\appendix
\onecolumn

\section{Metrics in This Paper}\label{sec:metric-details}

\paragraph{Notation}
Recall that all metrics we consider are computed on 
embeddings of datapoints. We denote the embedded value
of a datapoint by $\phi$, and denote an embedded dataset
by $\Phi$. We differentiate real and generated data with
subscripts: $\Phi_r$ is real data, $\Phi_g$ is generated
data. We use $|\Phi|$ to denote the size of the dataset
$\Phi$.

\paragraph{Source Code Repositories}
We used to following source code repositories for the metrics in this work:
\begin{itemize}
    \item I-Prec, I-Rec, Density, Coverage: \url{https://github.com/clovaai/generative-evaluation-prdc}, MIT License
    \item IAP, IBR: \url{https://github.com/ahmedmalaa/evaluating-generative-models}, MIT license or BSD 3-clause license (repository has MIT, files say BSD 3-clause)
    \item C-Prec, C-Rec: \url{https://github.com/FasilCheema/GenerativeMetrics}, GPL-3.0 license
    \item symPrec, symRec: \url{https://github.com/mahyarkoy/emergent_asymmetry_pr}, MIT license
    \item P-Prec, P-Rec: \url{https://github.com/kdst-team/Probablistic_precision_recall}, no license
\end{itemize}

\paragraph{Improved Precision / Recall}

\citet{kynkaanniemiImprovedPrecisionRecall2019} motivate the improved
precision / recall metrics as improvements to the curve-valued metric
of \citet{sajjadiAssessingGenerativeModels2018}. In particular, improved 
precision and recall are single numbers, making comparisons with them easier,
and they come with a practical algorithm to compute the metrics.

The basis for both improved precision and recall is approximating the support of the 
real or synthetic data distribution with a set of hyperspheres around each point,
with radius set to the $k$th nearest neighbour of each point. 
\begin{equation}
    S(\Phi) = \bigcup_{\phi\in \Phi} B(\phi, \NND_k(\phi, \Phi))
\end{equation}
where $B(\phi, r)$ is a hypersphere of radius $r$ centered at $\phi$, and 
$\NND_k(\phi, \Phi)$ is the distance to the $k$th nearest neighbour\footnote{
Excluding $\phi$ itself.} of $\phi$ in $\Phi$.
Improved precision then simply counts the fraction of synthetic points that are
in the support of the real data, while improved recall counts real points
that are in the support of the synthetic data.
\begin{align}
    \text{I-Prec}(\Phi_r, \Phi_g) &= \frac{1}{|\Phi_g|}
    \sum_{\phi_g\in \Phi_g} \ind_{\phi_g \in S(\Phi_r)}, \\
    \text{I-Rec}(\Phi_r, \Phi_g) &= \frac{1}{|\Phi_r|}
    \sum_{\phi_r\in \Phi_r} \ind_{\phi_r \in S(\Phi_g)}.
\end{align}

The hyperparameters of the metric are $k$, and the size of the synthetic 
dataset $|\Phi_g|$. \citet{kynkaanniemiImprovedPrecisionRecall2019}
choose $k = 3$ and use $|\Phi_r| = |\Phi_g| = 50000$ for image data.
We follow their choice of $k$.

\paragraph{Density / Coverage}
\citet{naeemReliableFidelityDiversity2020} point out that improved precision and
recall are vulnerable to outliers, since the hypersphere around an outlier is likely
to be large, and the metrics simply count how many points are in at least one
hypersphere. To fix these, \citet{naeemReliableFidelityDiversity2020} 
propose density and coverage. 

Density counts, for each synthetic datapoint, how many hyperspheres that the 
synthetic datapoint belongs to, and normalises the sum over all synthetic 
datapoints.
\begin{equation}
    \text{Density} = \frac{1}{k|\Phi_g|}\sum_{\phi_g\in \Phi_g}
    \sum_{\phi_r\in \Phi_r} \ind_{\phi_g \in B(\phi_r, \NND_k(\Phi_r))}.
\end{equation}

Coverage counts the fraction of real datapoints that have at least 
one synthetic point in their hypersphere.
\begin{equation}
    \text{Coverage} = \frac{1}{|\Phi_r|}\sum_{\phi\in \Phi_r} 
    \ind_{\exists \phi_g\in \Phi_g\colon \phi_g \in 
    B(\phi_r, \NND_k(\phi_r, \Phi_r))}.
\end{equation}

The hyperparameters for density and coverage are $k$ and 
$|\Phi_g|$. \citet{naeemReliableFidelityDiversity2020} choose them according to 
an analytical expression for $\E[\text{coverage}]$ for identical real and 
synthetic distributions. They first set $|\Phi_r| = |\Phi_g| = 10000$,
and then set $k = 5$ to obtain $\E[\text{coverage}] > 0.95$.
Since we do not always use $|\Phi_r| = |\Phi_g| = 10000$,
we solve $k$ from the expression of $\E[\text{coverage}]$
given by \citep[Lemma 2]{naeemReliableFidelityDiversity2020},
which supports $|\Phi_r| \neq |\Phi_g|$. We set a maximum
$k$ of 20.

\paragraph{$\alpha$-precision / $\beta$-recall}
Instead of placing hyperspheres around each datapoint like the previous metrics, $\alpha$-precision and $\beta$-recall estimate the supports
with a single hypersphere that contains an $\alpha$ (or $\beta$) fraction 
of the datapoints~\citep{alaaHowFaithfulYour2022}. Using a single hypersphere to approximate the support clearly
requires the embedding space to have suitable geometry, which is why 
\citet{alaaHowFaithfulYour2022} use a DeepSVDD 
neural network~\citep{ruffDeepOneClassClassification2018}, which explicitly 
attempts to embed the datapoints inside a hypersphere, as their embedding. They train this network on the real data in tabular
settings.

Let $S_\alpha(\Phi_r)$ be the minimum-volume hypersphere that contains an $\alpha$
fraction of the real embeddings. Then 
\begin{equation}
    P_\alpha = \frac{1}{|\Phi_g|}\sum_{\phi_g\in \Phi_g}
    \ind_{\phi_g \in S_\alpha(\Phi_r)}.
\end{equation}
Let $S_\beta(\Phi_g)$ be the minimum-volume hypersphere that contains 
a $\beta$-fraction of the synthetic embeddings. Let
\begin{equation}
    \phi_{g,\beta}^* = \NN_1(\phi_r, \Phi_g \cap S_\beta(\Phi_g))
\end{equation}
where $\NN_1(\phi, \Phi)$ is the nearest neighbour\footnote{Including 
$\phi$ itself, so if $\phi\in \Phi, \NN_1(\phi, \Phi) = \phi$.} 
of $\phi$ in $\Phi$. Then
\begin{equation}
    R_\beta = \frac{1}{|\Phi_r|}\sum_{\phi_r\in \Phi_r}
    \ind_{\phi_{g,\beta}^*
    \in B(\phi_r, \NND_k(\phi_r, \Phi_r))}.
\end{equation}

$P_\alpha$ and $R_\beta$ have a separate value for each 
$\alpha, \beta \in [0, 1]$. To aggregate these values, 
\citet{alaaHowFaithfulYour2022} define integrated $\alpha$-precision and 
$\beta$-recall as
\begin{align}
    \mathrm{IAP} &= 1 - 2\int_0^1 |P_\alpha - \alpha|\dx \alpha, \\
    \mathrm{IBR} &= 1 - 2\int_0^1 |R_\beta - \beta|\dx \beta.
\end{align}
These integrated metrics are what we evaluate in our experiments.

$\alpha$-precision and $\beta$-recall have the architecture
choices and training details of the embedding network as 
hyperparameters. $\beta$-recall additionally has $k$.
We use the code of \citet{alaaHowFaithfulYour2022} to compute
both metrics, so we use their choices for the embedding network
and $k$. We use the same normalisation we use with 
other metrics before applying the neural network embedding.

\paragraph{Precision / Recall Cover}
Precision cover and recall cover~\citep{cheemaPrecisionRecallCover2023} are based 
on the intuition that metrics should not care whether a part of the support
is covered if that part has a very small probability. For example, parts of 
the real data distribution that have very low probability need not be covered 
by the synthetic data generator.

\begin{equation}
    \text{C-Prec} = \frac{1}{|\Phi_g|}\sum_{\phi_g \in \Phi_g}
    \ind_{|\{\phi_r\in \Phi_r 
    \mid \phi_r \in B(\phi_g, \NND_{k'}(\phi_g, \Phi_g))\}| \leq k},
\end{equation}
\begin{equation}
    \text{C-Rec} = \frac{1}{|\Phi_r|}\sum_{\phi_r \in \Phi_r}
    \ind_{|\{\phi_g\in \Phi_g 
    \mid \phi_g \in B(\phi_r, \NND_{k'}(\phi_r, \Phi_r))\}| \leq k}.
\end{equation}

Precision cover and recall cover have the hyperparemeters $k$ and $k'$.
\citet{cheemaPrecisionRecallCover2023} set 
$k' = \lceil \ln |\Phi_r| + 6 \rceil$ and 
$k = \lceil k' / 3 \rceil$, which we follow in our 
experiments.

\paragraph{Symmetric Precision / Recall}
\citet{khayatkhoeiEmergentAsymmetryPrecision2023} point out that previous
metrics behave asymmetrically when the real distribution is on the surface of 
a hypersphere, and the synthetic distribution is on the surface of another
hypersphere with the same center, but either a larger or smaller radius.
The asymmetric behavior is a problem, since the surfaces of the hyperspheres 
are disjoint, so both fidelity and diversity measures should be close to zero when
the difference in radii is large enough.

To fix the asymmetry, \citet{khayatkhoeiEmergentAsymmetryPrecision2023} first
define complement versions of improved precision and recall
\begin{align}
    \text{cPrecision} &= \frac{1}{\Phi_g}\sum_{\phi_g\in \Phi_g}
    \ind_{|B(\phi_g, \NND_k(\phi_g, \Phi_g)) \cap \Phi_r| \geq 1}, \\
    \text{cRecall} &= \frac{1}{\Phi_r}\sum_{\phi_r\in \Phi_r}
    \ind_{|B(\phi_r, \NND_k(\phi_r, \Phi_r)) \cap \Phi_g| \geq 1}. \\
\end{align}
cPrecision counts the fraction of synthetic data points whose $k$-nearest 
neighbourhoods contain at least one real datapoint, and cRecall reverses the 
roles of real and synthetic data. Note that cRecall is the same quantity as 
coverage.

cPrecision and cRecall have the problematic asymmetric behaviour, but in opposite
direction from improved precision and recall, so 
\citet{khayatkhoeiEmergentAsymmetryPrecision2023} define symmetric metrics that
fix the asymmetry.
\begin{align}
    \text{symPrec} &= \min(\text{cPrecision}, \text{I-precision}), \\
    \text{symRec} &= \min(\text{cRecall}, \text{I-Recall}).
\end{align}

The hyperparameters for symPrecision and symRecall are 
$k$, and $|\Phi_g|$. \citet{khayatkhoeiEmergentAsymmetryPrecision2023}
use $|\Phi_g| = |\Phi_r| = 10000$, and set $k = 5$. 
We follow their choice of $k$.

\paragraph{Probabilistic Precision / Recall}
\citet{parkProbabilisticPrecisionRecall2023} aim to solve the outlier
robustness problem of improved precision and recall by approximating
the supports of the real and synthetic data probabilistically. Instead of 
computing a binary value of whether a datapoint is in the approximate support 
or not, they compute compute an estimate for the probability that the point is
in the support, and average these probability estimates to obtain the aggregate
metrics called probabilistic precision and recall.

\citet{parkProbabilisticPrecisionRecall2023} call the function estimating the 
probability of $\phi$ to be in the support of $\Phi$ the probabilistic
scoring rule (PSR), and define it as
\begin{equation}
    \mathrm{PSR}_{\Phi}(\phi') = 1 - \prod_{\phi \in \Phi}
    (1 - f(\phi, \phi', R_{\Phi})).
\end{equation}
The function $f(\phi, \phi', R_\Phi)$ is a simple estimate of the probability 
that $\phi'$ is in the support around $\phi$:
\begin{equation}
    f(\phi, \phi', R) = 
    \begin{cases}
        1 - \frac{||\phi - \phi'||_2}{R} & \text{ if } ||\phi - \phi'||_2 \leq R \\
        0 & \text{ otherwise}
    \end{cases}
\end{equation}
where
\begin{equation}
    R_{\Phi} = \frac{a}{|\Phi|}\sum_{\phi\in \Phi}\NND_k(\phi, \Phi)
\end{equation}
and $a > 0$ is a hyperparameter.

Probabilistic precision and recall are then averages of the estimated probabilities:
\begin{align}
    \text{P-Prec} &= \frac{1}{|\Phi_g|}\sum_{\phi\in \Phi_g} 
    \mathrm{PSR}_{\Phi_r}(\phi_g) \\
    \text{P-Rec} &= \frac{1}{|\Phi_r|}\sum_{\phi\in \Phi_r} 
    \mathrm{PSR}_{\Phi_g}(\phi_r).
\end{align}

The hyperparameters of P-precision and P-recall are 
$a$, $k$, which control the choice of $R$, and 
$|\Phi_g|$. \citet{parkProbabilisticPrecisionRecall2023}
set $a = 1.2$, $k = 4$ and $|\Phi_g| = |\Phi_r| = 10000$.
We follow their choice of $a$ and $k$.

\section{Sanity Check Details}\label{sec:sanity-check-details}

\subsection{General Success Criteria}
\begin{figure*}[h!]
    \begin{subfigure}{1.0\textwidth}
        \centering
        \includegraphics[width=\textwidth]{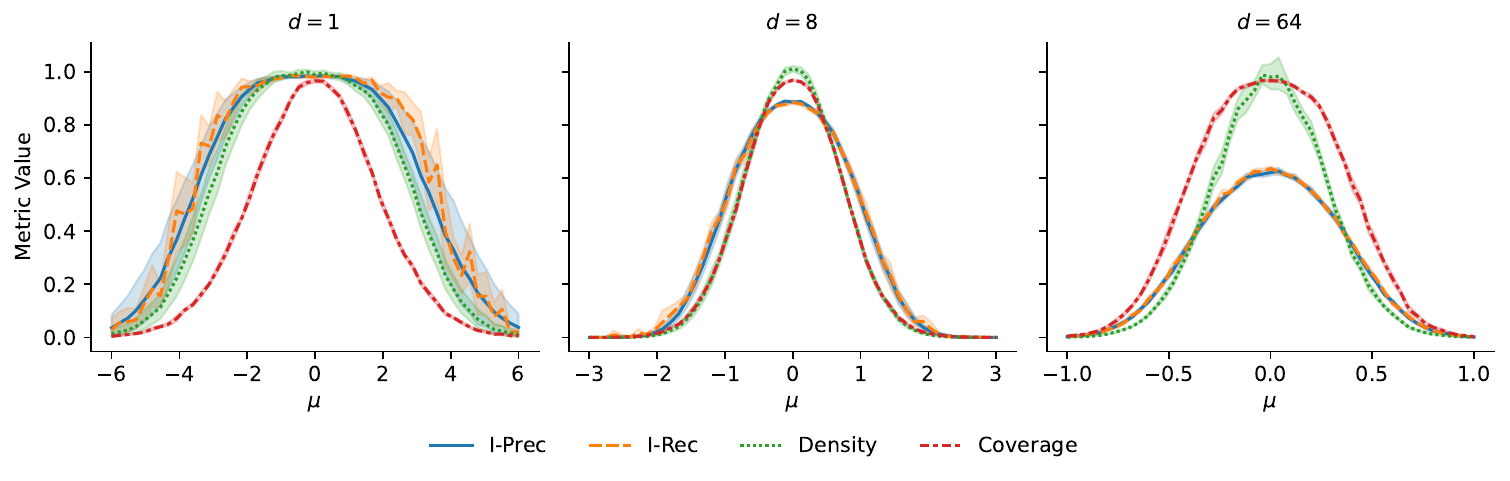}
        \caption{}
        \label{fig:gaussian-mean-difference-0}
    \end{subfigure}
    \begin{subfigure}{1.0\textwidth}
        \centering
        \includegraphics[width=\textwidth]{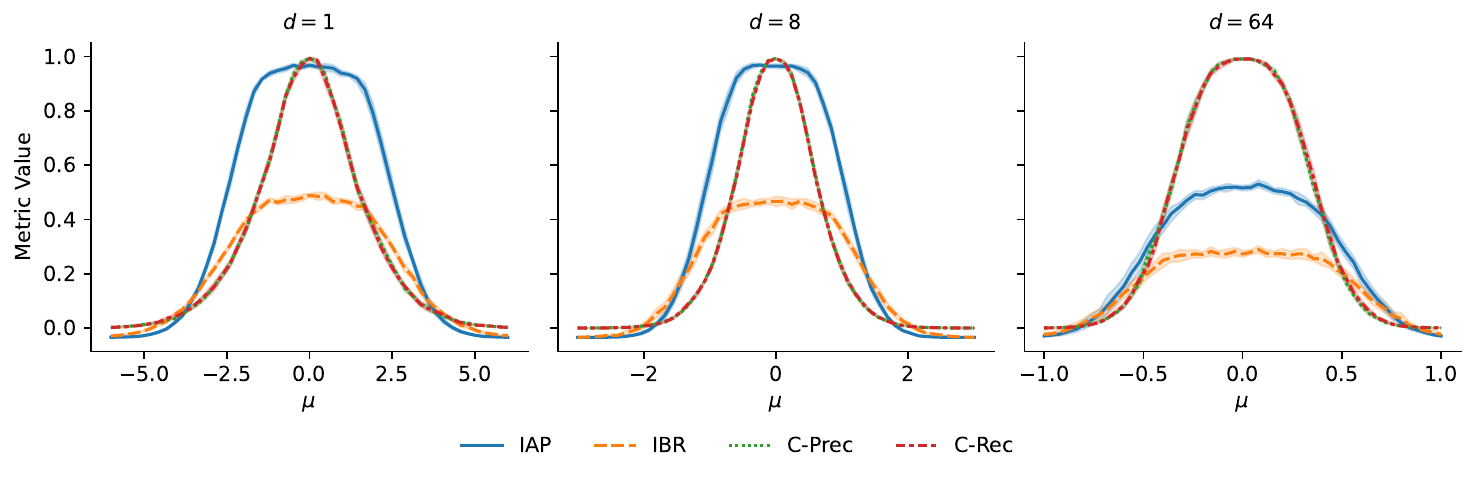}
        \caption{}
        \label{fig:gaussian-mean-difference-1}
    \end{subfigure}
    \begin{subfigure}{1.0\textwidth}
        \centering
        \includegraphics[width=\textwidth]{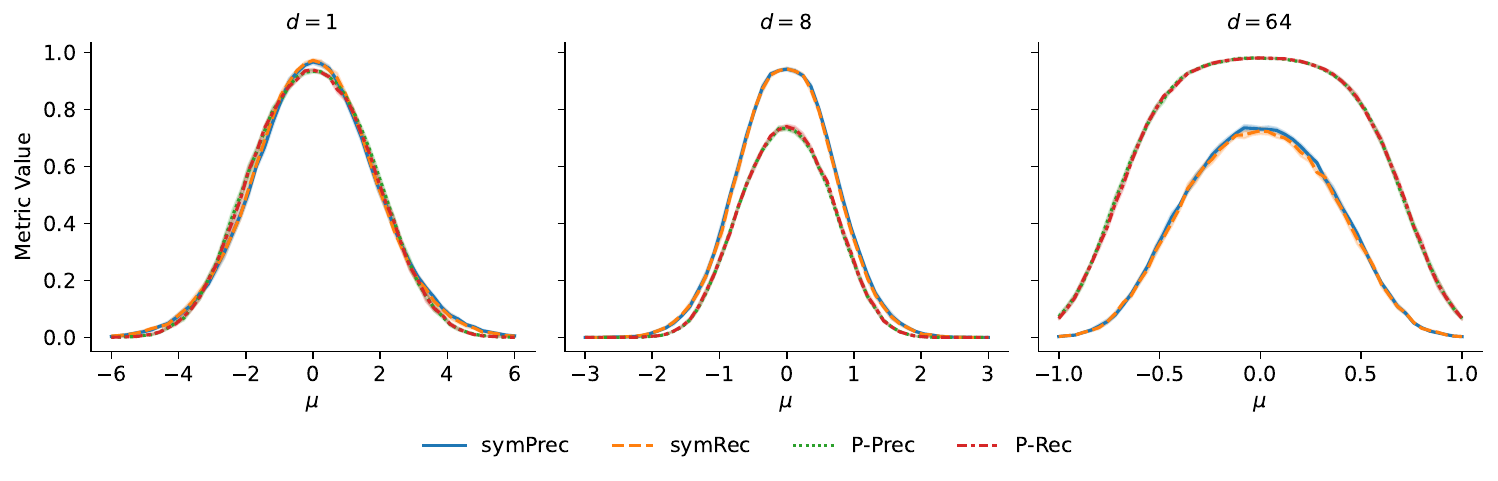}
        \caption{}
        \label{fig:gaussian-mean-difference-2}
    \end{subfigure}
    \caption{
        Gaussian mean difference check: two $d$-dimensional Gaussian distributions with real 
        mean 0, synthetic mean $\mu$, covariance $I_d$. 
    }
    \label{fig:gaussian-mean-difference}
\end{figure*}

We check the whether each metric passed or failed each check programmatically,
checking whether the values of the metric meet given criteria. We detail these
criteria when introducing each sanity check later in this section.
Each criteria looks at specific points on the curves we plot in 
Section~\ref{sec:sanity-check-figures}, see for example 
Figure~\ref{fig:gaussian-mean-difference}. The curves represent the mean metric
values over 10 repeats. Some criteria only look at 
the value at a few significant points, often the left and right extremes and 
the middle. With these criteria, we check that the metric value at the 
given point is within 0.05 of a given value, unless otherwise specified.
More complex criteria look at the overall shape of the curve
by comparing values at several points. When there are multiple plots 
for different variations of the setting, such as multiple values of 
$d$ in Figure~\ref{fig:gaussian-mean-difference}, a metric must pass the 
criteria for all variations. Next, we give the our detailed 
definitions of the overall shape criteria.

\paragraph{Bell shape}
Informally, a bell-shaped curve has low values on the left, increases
until a peak at a given midpoint, and then decreases again towards the right.
Our specific criteria for a bell-shaped curve are:
\begin{itemize}
    \item The difference between the values at both left and right extremes and
    at the midpoint is at least 0.2.
    \item The difference between the midpoint value and the maximum value
    is at most 0.1.
    \item The difference between the left extreme value or the right 
    extreme value and the minimum value is at most 0.1.
\end{itemize}

\paragraph{Low-to-high shape}
Informally, a low-to-high shape is any curve that has the lowest values
at the left extreme and highest values at the right extreme, with a clear
difference between the left and right extremes.
Our specific criteria are:
\begin{itemize}
    \item The difference between the right extreme value and the left
    extreme value is at least 0.2.
    \item The difference between the left extreme value and the minimum value
    is at most 0.1.
    \item The difference between the maximum value and the right extreme value
    is at most 0.1.
\end{itemize}

\paragraph{High-to-low shape}
Informally, high-to-low shape is similar to the low-to-high shape, but the 
values go from high on the left to low on the right. Our specific criteria 
are:
\begin{itemize}
    \item The difference between the left extreme value and the right
    extreme value is at least 0.2.
    \item The difference between the right extreme value and the minimum value
    is at most 0.1.
    \item The difference between the maximum value and the left extreme value
    is at most 0.1.
\end{itemize}

\paragraph{High-to-low shape with middle drop}
Informally, this shape requires a curve to clearly decrease between the 
left and right extremes like the high-to-low shape, but also clearly 
decrease at a given point in the middle. Our specific criteria are:
\begin{itemize}
    \item All of the criteria for a high-to-low shape.
    \item The difference between the left extreme value and a value at a 
    given quantile of the x-axis is at least 0.1.
\end{itemize}

\paragraph{Horizontal line shape}
This shape represents an approximately constant curve. Our specific criterion
is that the difference between the maximum value and the minimum value is 
at most 0.05.

\paragraph{Converging line shape}
This shape represents a curve that converges towards some value, and 
is then approximately constant. The convergence must happen before a 
given quantile of the x-axis. The specific criterion is that
the difference between the maximum and minimum values that are to the right of
the given quantile x-axis quantile is at most 0.05.

\subsection{Gaussian Mean Difference (+ outlier)}
\begin{figure}
    \centering
    \includegraphics[width=1.0\linewidth]{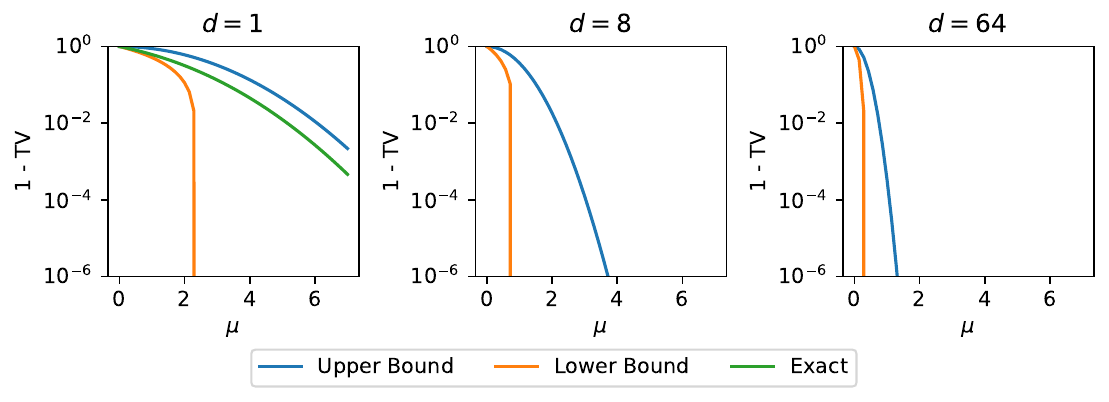}
    \caption{
        Total variation distance between a standard Gaussian and a Gaussian
        with mean $\mu 1_d$. The lower and upper bounds are based on 
        Hellinger distance.
    }
    \label{fig:gaussian-mean-difference-tv-visualisation}
\end{figure}

In this check, the real distribution is a $d$-dimensional Gaussian
standard Gaussian distribution, and the synthetic distribution is 
a similar Gaussian with mean $\mu 1_d$, where $\mu \in \R$ and 
$1_d$ is a $d$-dimensional vector of all ones. We vary $d \in \{1, 8, 64\}$
and 
\begin{equation}
    \mu \in \begin{cases}
    [-6, 6], & d = 1 \\
    [-3, 3], & d = 8 \\
    [-1, 1], & d = 64.
    \end{cases}
\end{equation}
These ranges were selected to ensure that the total variation distance
between the real and synthetic distributions at the extreme ends of 
the $\mu$ interval is at least $0.99$, see 
Figure~\ref{fig:gaussian-mean-difference-tv-visualisation}. We set 
the real and synthetic dataset sizes to 1000.

In the ``with outlier'' case, we add one outlier point to either the 
real or synthetic data at the largest $\mu$ value.

\paragraph{Success Criteria}
As $\mu$ increases, the real and synthetic distribution are initially
almost disjoint, then become more similar until they are identical at 
$\mu = 0$, and then become less similar until they are again almost 
disjoint at the right extreme. As such, to pass D1b (purpose), we require 
all metrics to have a bell-shaped curve centered at $\mu = 0$. 
For D4 (bounds), we require all metrics to be close to 0 at the 
left and right extremes, and close to 1 at $\mu = 0$.

The criteria for the ``with outlier'' case are the same, since a single
outlier should not significantly change synthetic data evaluations.

\subsection{Gaussian Standard Deviation Difference}
\begin{figure}
    \centering
    \includegraphics[width=1.0\linewidth]{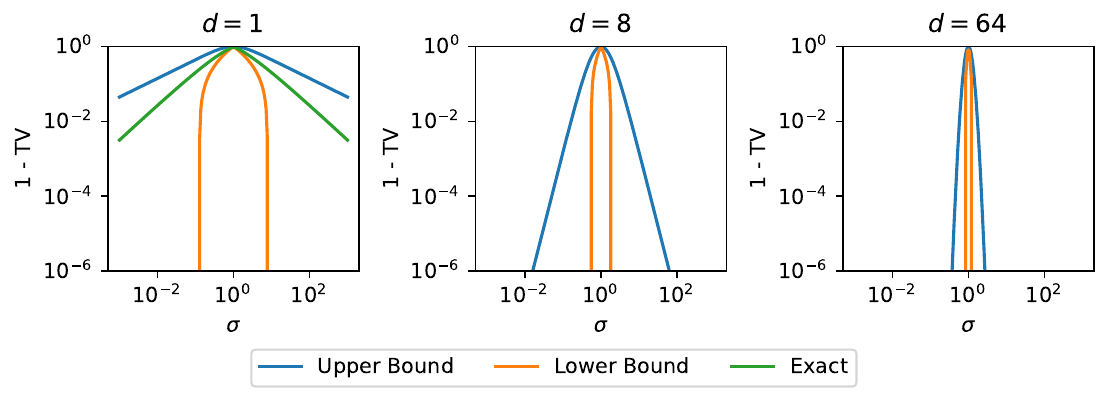}
    \caption{
        Total variation distance between a standard Gaussian and a Gaussian
        with covariance $\sigma^2 I_d$. The lower and upper bounds are based on 
        Hellinger distance.
    }
    \label{fig:gaussian-std-difference-tv-visualisation}
\end{figure}

In this check, the real distribution is a $d$-dimensional Gaussian
standard Gaussian distribution, and the synthetic distribution is 
a similar Gaussian with covariance $\sigma^2 I_d$, where $\sigma > 0$ and
$I_d$ is the $d\times d$ identity matrix. We vary $d \in \{1, 8, 64\}$
and logarithmically vary $\sigma$ in the interval
\begin{equation}
    \sigma \in \begin{cases}
    [10^{-3}, 10^3], & d = 1 \\
    [10^{-1}, 10^1], & d = 8 \\
    [10^{-0.5}, 10^{0.5}], & d = 64.
    \end{cases}
\end{equation}
As in the Gaussian mean difference check, we selected the $\sigma$ intervals
to ensure that the total variation distance between the 
real and synthetic distributions at the extreme ends is at least 0.99,
see Figure~\ref{fig:gaussian-std-difference-tv-visualisation}. We set 
the real and synthetic dataset sizes to 1000.

\paragraph{Success Criteria}
In this check, the synthetic distribution is very narrow with the smallest 
$\sigma$ values, but contained in the real distribution, so fidelity metrics
should have high values and diversity metrics should have low values.
As $\sigma$ increases, the metrics become more similar, and become identical
at $\sigma = 1$, at which point all metrics should have high values.
As $\sigma$ increases further, the synthetic distribution becomes wider,
so fidelity metrics should drop. Diversity metrics can either 
fall to low value or stay at high values, as described in 
Section~\ref{sec:success-criteria}.

Our specific criteria are the following:
\begin{itemize}
    \item For D1b (purpose):
    \begin{itemize}
        \item Fidelity metrics should have a high-to-low shape.
        \item Low metrics should have a bell-shape with midpoint $\sigma = 1$,
        high metrics should have a low-to-high shape.
    \end{itemize}
    \item For D4 (bounds):
    \begin{itemize}
        \item Fidelity metrics should have values close to 1, 1, 0
        at the left extreme, $\sigma = 1$, and at the right extreme, respectively.
        \item Diversity metrics should have values close to 0, 1 at the
        at the left extreme and $\sigma = 1$, respectively. Low metrics
        should have values close to 0 and high metrics should have 
        values close to 1 at the right extreme.
    \end{itemize}
\end{itemize}

\subsection{Sequential / Simultaneous Mode Dropping}
This checks evaluates how the metrics detect dropped modes. The real data
is a 10-component Gaussian mixture in $d \in \{1, 8, 64\}$ dimensions, 
with the component means evenly spaced between 0 and $10\cdot 1_d$, 
and the component standard deviations are
\begin{equation}
    \sigma = \begin{cases}
        \frac{1}{6}, & d = 1 \\
        \frac{1}{3}, & d = 8 \\
        1, & d = 64.
    \end{cases}
\end{equation}
We set the real and synthetic dataset sizes to 1000.

We drop up to 9 of the modes in the synthetic distribution either 
simultaneously or sequentially,
as done by \citet{naeemReliableFidelityDiversity2020}. In simultaneous 
dropping, the weights of the 9 modes are gradually decreased, until their 
weight is zero and the synthetic distribution is just a single mode.
In sequential dropping, we completely remove 0 to 9 of the modes
from the synthetic distribution.

\paragraph{Success Criteria}
With both simultaneous and sequential dropping, the real and synthetic
distributions are initially identical. Since the synthetic distribution
is contained in the real distribution regardless of the dropped modes,
fidelity metrics have high values in all cases. Diversity metrics should be 
sensitive to mode dropping, and decrease as mode modes are dropped.
Our specific criteria are:
\begin{itemize}
    \item For D1b (purpose):
    \begin{itemize}
        \item Fidelity metrics should have horizontal lines.
        \item Diversity metrics should have a high-to-low with middle drop shape,
        with a drop of at least 0.1, at the 0.95 quantile in the simultaneous 
        case, and at the 0.5 quantile for the sequential case.
    \end{itemize}
    \item For D4 (bounds): 
    \begin{itemize}
        \item Fidelity metrics should be close to one at both left and 
        right extremes.
        \item Diversity metrics should be close to one at the left 
        extreme.
    \end{itemize}
\end{itemize}

\subsection{Mode Dropping + Invention}
\begin{figure}
    \centering
    \includegraphics[width=0.5\linewidth]{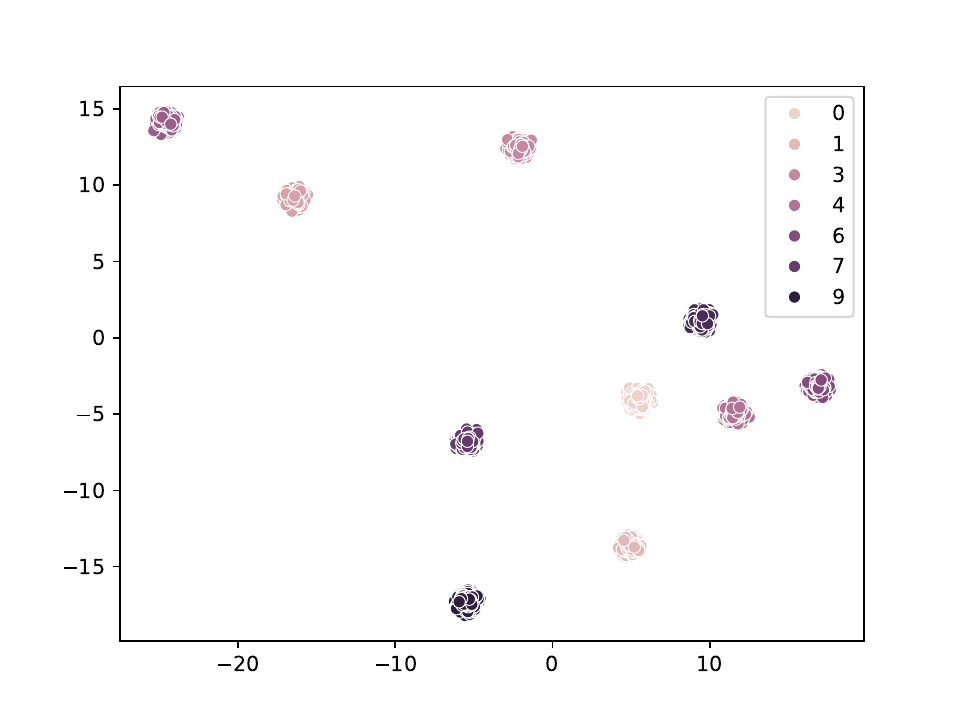}
    \caption{
        Samples of each component in the mode dropping and invention sanity 
        check.
    }
    \label{fig:gaussian-mixture-mode-dropping-invention-components}
\end{figure}

This check from \citet{kynkaanniemiImprovedPrecisionRecall2019} evaluates how 
the metrics detect both mode dropping and 
invention. We start be defining 10 two-dimensional Gaussian component 
distributions, which we show in 
Figure~\ref{fig:gaussian-mixture-mode-dropping-invention-components}.
The standard deviation of each component is 0.25, and the means were 
randomly sampled from a Gaussian with mean 0 and standard deviation 10.
These same components are used for all repeats of the experiment.
The real distribution is a mixture of components 0 to 4, and the synthetic
distribution has between 1 and 10 components, included in order. As a 
result, the 5-component synthetic distribution is identical to the real
distribution. With less than 5 modes, the synthetic distribution is dropping
modes, and with more than 5 modes, it is inventing modes.
We set the real and synthetic dataset sizes to 1000.

\paragraph{Success Criteria}
\begin{itemize}
    \item For D1b (purpose):
    \begin{itemize}
        \item Fidelity metrics should have a high-to-low shape.
        \item Diversity metrics should have a low-to-high shape.
    \end{itemize}
    \item For D4 (bounds):
    \begin{itemize}
        \item Both fidelity and diversity metrics should be close to 
        1 at 5 synthetic components.
        \item Fidelity metrics should be close to 1 with 1 synthetic component.
        \item Diversity metrics should be close to 1 with 10 synthetic components.
    \end{itemize}
\end{itemize}

\subsection{Hypersphere Surface}
This check from \citet{khayatkhoeiEmergentAsymmetryPrecision2023} highlights
a problematic behavior in some metrics. The real distribution
is a uniform distribution on the surface of a $d$-dimensional hypersphere
with radius 1, and the synthetic distribution is a similar uniform
distribution on the surface of a hypersphere of radius $r \in [0.1, 1.9]$.
We set the real and synthetic dataset sizes to 1000, and vary 
$d \in \{2, 8, 128\}$.

\paragraph{Success Criteria}
With $r \neq 1$, the real and synthetic distributions are completely disjoint,
so all metrics should have low values. In practice, detecting very small 
differences in $r$ is very hard for a general-purpose metric, so we 
only require that the metrics detect a difference at the extreme values of 
$r$. Specifically, for D1b (purpose), all metrics should have a bell shape
with midpoint 1. For D4 (bounds), all metrics should have be close to 
zero at extreme values of $r$, and close to one at $r = 1$.

\subsection{Hypercube, Varying Sample / Synthetic Size}
This check from \citet{cheemaPrecisionRecallCover2023} evaluates 
how the metrics have with different sizes of real and synthetic data.
The real distribution is uniform distribution on a $d$-dimensional 
hypercube with side length 1.
The synthetic distribution is a similar uniform distribution, but the 
hypercube has been translated equally on all axes so that the 
$d$-dimensional volume of the overlapping area between the cubes is 0.2.

We vary either both the real and synthetic dataset size with both of them
equal, or only the synthetic dataset size, in which case 
the real dataset size $|\Phi_r| = 1000$. The changing size is varied
logarithmically in $[10^2, 10^4]$. We also vary $d \in \{1, 8, 64\}$.

\paragraph{Success Criteria}
According to \citet{cheemaPrecisionRecallCover2023}, both fidelity and 
diversity metrics should have a value of 0.2, the volume of the overlap
between the cubes, in this setup, since that is the value of a theoretical
notion of precision and recall that can be computed in this case.
We expect the metrics to get close to this value with the largest 
dataset size for D1b (purpose). 
We also expect the 
metric values to converge as the size increases (converging line shape), 
and require convergence by size 1000. This is counted under D2 (hyperparameters)
when only the synthetic data size changes, and under D3 (data)
when the sizes of both datasets change.

\subsection{Sphere vs. Torus}
This check from \citet{cheemaPrecisionRecallCover2023} evaluates how the 
metrics distinguish two disjoint distributions with a non-trivial geometry.
One of the distributions is a uniform distribution on a sphere with radius
0.8. The other distribution is on a torus with major radius 1, minor radius
0.1, which surrounds the sphere, but is disjoint from it. We look at both 
making the sphere distribution the real distribution, and making the torus
distribution real.

\begin{figure}
    \centering
    \includegraphics[width=0.5\linewidth]{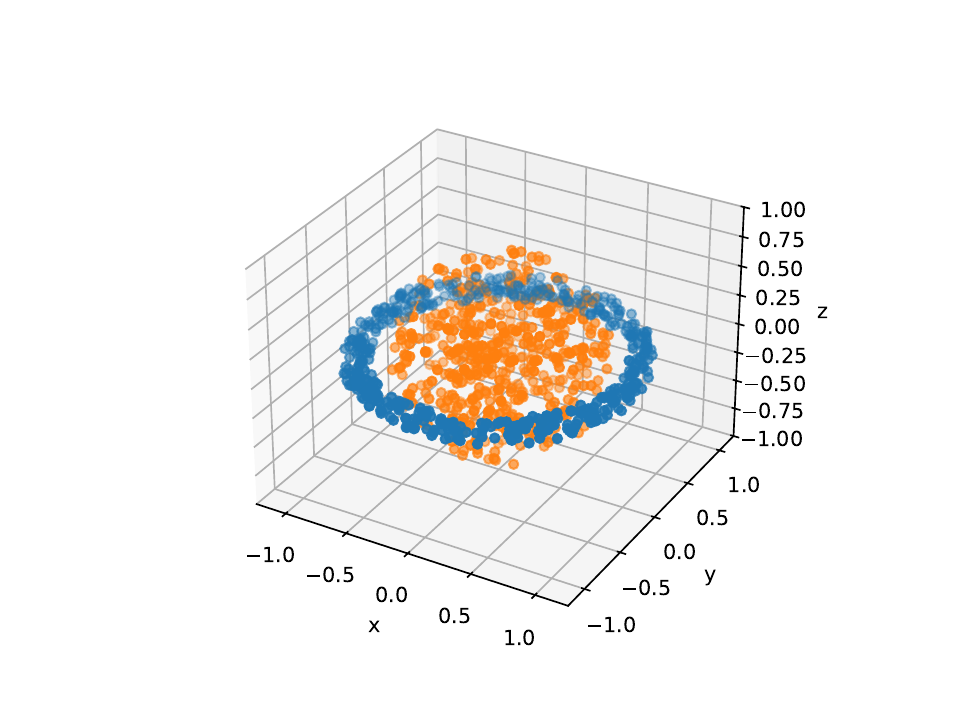}
    \caption{
        Samples from the sphere and torus distributions 
        in the sphere vs. torus sanity check.
    }
    \label{fig:sphere-torus-samples}
\end{figure}

We sample the sphere distribution by sampling a cube containing the sphere,
and rejecting samples outside the sphere. We sample the torus by 
first uniformly sampling an angle $\theta \in [0, 2\pi]$, and sampling a 
point uniformly on the circle that is formed by intersecting the torus
with the $y = 0$ plane on the positive $x$ side. The point is then
rotated by $\theta$ around the $z$ axis to obtain the final sample.
Figure~\ref{fig:sphere-torus-samples} shows samples from both distributions.

We set the real dataset size $|\Phi_r| = 1000$. We logarithmically vary 
the synthetic dataset size $|\Phi_g| \in [10^2, 10^4]$.

\paragraph{Success Criteria}
The two distributions in this check are disjoint, so all metrics should have 
low values. We do not place requirements at small synthetic dataset sizes,
but we do require the metrics to pass regardless of which distribution is 
considered the real one. Specifically, for 
D1b (purpose), we require the metrics to have a converging line shape,
with convergence at latest at 1000 synthetic datapoints. For D4 (bounds),
we require the metrics to be close to 0 at the maximum number of synthetic 
datapoints.

\subsection{Mode Collapse}
\begin{figure}
    \centering
    \includegraphics[width=1.0\linewidth]{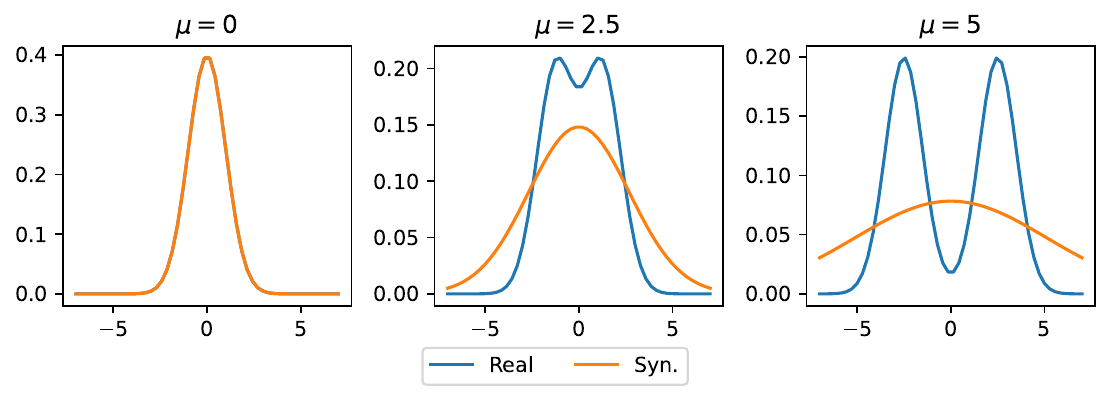}
    \caption{
        Comparison of pdfs for the real and synthetic distributions in the 
        one vs. two modes check.
    }
    \label{fig:one-vs-two-modes-pdfs}
\end{figure}

This check from \citet{alaaHowFaithfulYour2022} evaluates the mode resolution
of the metrics: how well do they distinguish a real distribution of two modes
from a synthetic distribution that models both modes with a single wide mode.
The real distribution is an equal mixture of two $d$-dimensional 
($d\in \{1, 8, 64\}$) Gaussians
with means $-\frac{1}{2}\mu 1_d$ and $\frac{1}{2}\mu 1_d$ and unit variance,
where $\mu \in [0, 5]$ is varied.
The synthetic distribution is a Gaussian with mean 0 and 
covariance $(1 + \mu^2) I_d$. The size of both the real and synthetic 
dataset is 1000. See Figure~\ref{fig:one-vs-two-modes-pdfs} for 
a visualisation of the distributions with $d = 1$.

\paragraph{Success Criteria}
When $\mu = 0$, the distributions are identical. As $\mu$ increases, the 
modes of the real distribution become further separated, while the synthetic 
distribution widens to cover both modes. As a result, metric values should 
be high at low $\mu$ values, and decrease when $\mu$ is increased for fidelity
metrics. Diversity metrics can either stay high or decrease depending on 
whether they are high or low metrics.

Our specific criteria are:
\begin{itemize}
    \item For D1b (purpose):
    \begin{itemize}
        \item Fidelity metrics should have a high-to-low shape.
        \item Diversity metrics should have a horizontal line shape for 
        high metrics, and a high-to-low shape for low metrics.
    \end{itemize}
    \item For D4 (bounds): values should be close to one when $\mu = 0$ 
    for all metrics.
\end{itemize}

\subsection{Scaling One Dimension}
This check evaluates how robust the metrics are to differing scales of the 
data variables. The real data distribution is a two-dimensional standard 
Gaussian with the second dimension scaled by multiplying it with $s$. 
The synthetic data distribution is a similarly scaled Gaussian, but with 
mean $(6, 0)$. The scale is logarithmically varied in $s \in [10^{-3}, 10^3]$.
The size of both datasets is 1000.

\paragraph{Success Criteria}
The two distributions are almost disjoint in the first dimension, so all metrics
should have low values. Specifically, we require a horizontal line shape 
for D5 (invariance), and a values close to zero on both left and right extremes
for desiderata D4 (bounds).

\subsection{Gaussian Mean Difference + Pareto}
The aim of this check is to evaluate how well the metrics
handle a heavy-tailed power-law distribution, which are common
in tabular data. The setup is the same as in 
the Gaussian mean difference check with $d = 1$, but 
both the real and synthetic data contain an additional 
Pareto-distributed variable, which which has an identical 
distribution between the real and synthetic data.
The specific Pareto distribution is type 1\footnote{
The pdf is $p(x) = \frac{\alpha}{x^{\alpha + 1}}$ on $x \geq 1$.
} with shape 
$\alpha = 1.01$ and scale 1.

\paragraph{Success Criteria}
Since the presence of the Pareto-distributed variable 
with an identical distribution in the real and synthetic 
data should not affect data quality, the success criteria
are the same as in the Gaussian mean difference check.

\subsection{One disjoint Dimension + Many Identical Dimensions}
This check evaluates how the metrics handle a difference in
just one of a large number of dimensions. The real 
distribution is a $d + 1$-dimensional standard Gaussian,
and the synthetic distribution is a similar Gaussian, except
the mean of the first dimension is 6. The number of 
identical dimensions $d$ is varied logarithmically in
$d\in [1, 10^3]$. The sizes of the real and synthetic 
datasets are 1000.

\paragraph{Success Criteria}
Since the two distributions are almost disjoint due to the 
difference in the first dimension, all metrics should have
low values regardless of the number of extra dimensions.
Specifically, for D1b (purpose), the values should be a 
horizontal line, and for D4 (bounds), they should be 
close to 0 at both left and right extremes.

\subsection{Discrete Numerical vs. Continuous Numerical}
This check evaluates how the metrics detect a difference 
between a continuous distribution and a discrete 
numerical distribution. This difference can occur in both
directions in tabular data. Many numerical values, such 
as a count of something, or age in years, are always 
integers in real data, but a synthetic data generator 
based on transforming a continuous distribution,
like a GAN or a diffusion model, will likely model them
as a continuous value.\footnote{
The continuous values can of course be rounded to integers,
but evaluation metrics should detect a lack of rounding.
}
In the other direction, some synthetic data 
generators~\citep{mckennaAIMAdaptiveIterative2022}
must discretise continuous variables, and can only output 
the discretised values. Evaluation metrics should detect
differences in both directions.

One distribution is a standard Gaussian multiplied by a scale $s$, and the 
other distribution is the same Gaussian and multiplier, but 
its values have been rounded to integers.
We look at both the case where the continuous distribution is 
real, and the case where the discrete distribution is real.
The scale $s$ is varied logarithmically in $s\in [1, 10^3]$.
The sizes of the real and synthetic datasets are 1000.

\paragraph{Success Criteria}
The two distributions in this check are almost completely 
disjoint, since one of them is continuous and one of them
is discrete.
However, when the discrete distribution is real, the 
continuous synthetic distribution covers all of its values,
so high-type diversity metrics can have high values in this 
case. Conversely, when the continuous distribution is real,
all values from the discrete synthetic distribution have 
reasonable density in the real distribution, so fidelity 
metrics should have high values.

Our specific criteria are:
\begin{itemize}
    \item For desiderata D1b (purpose), all metrics should be 
    horizontal lines regardless of which distribution is 
    considered real.
    \item For D4 (bounds):
    \begin{itemize}
        \item If the discrete distribution is real:
        \begin{itemize}
            \item Fidelity metrics should be close to 0 at 
            small and large discretisation scales.
            \item High diversity metrics should be close to 1
            and low diversity metrics should be close to 0
            at small and large discretisation scales.
        \end{itemize}
        \item If the continuous distribution is real:
        \begin{itemize}
            \item Fidelity metrics should be close to 1 at
            small and large discretisation scales.
            \item Diversity metrics should be close to 0
            at small and large discretisation scales.
        \end{itemize}
    \end{itemize}
\end{itemize}

\section{Sanity Check Result Plots}\label{sec:sanity-check-figures}

\begin{figure*}
    \centering
    \begin{subfigure}{1.0\textwidth}
        \centering
        \includegraphics[width=\textwidth]{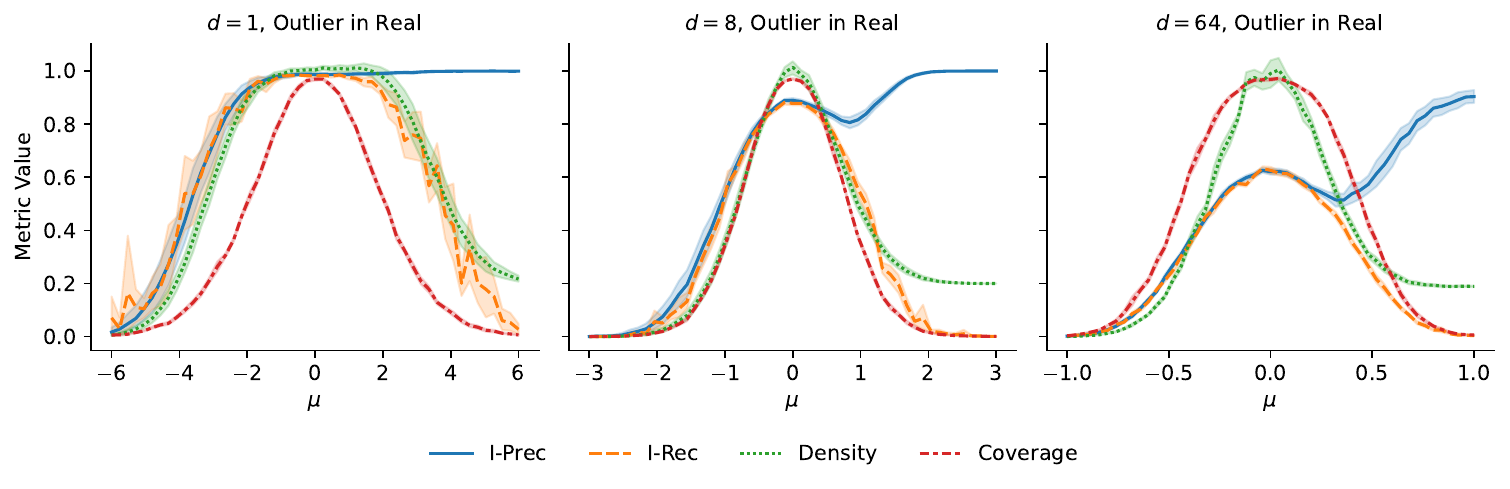}
        \caption{}
        \label{fig:gaussian-mean-difference-with-real-outlier-0}
    \end{subfigure}
    \begin{subfigure}{1.0\textwidth}
        \centering
        \includegraphics[width=\textwidth]{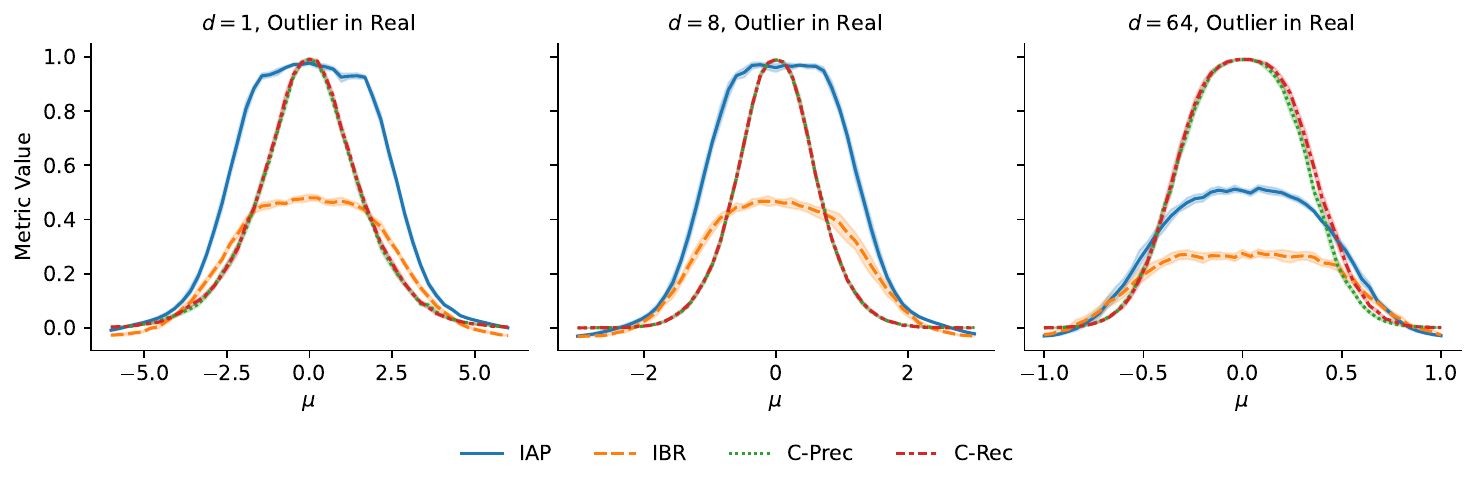}
        \caption{}
        \label{fig:gaussian-mean-difference-with-real-outlier-1}
    \end{subfigure}
    \begin{subfigure}{1.0\textwidth}
        \centering
        \includegraphics[width=\textwidth]{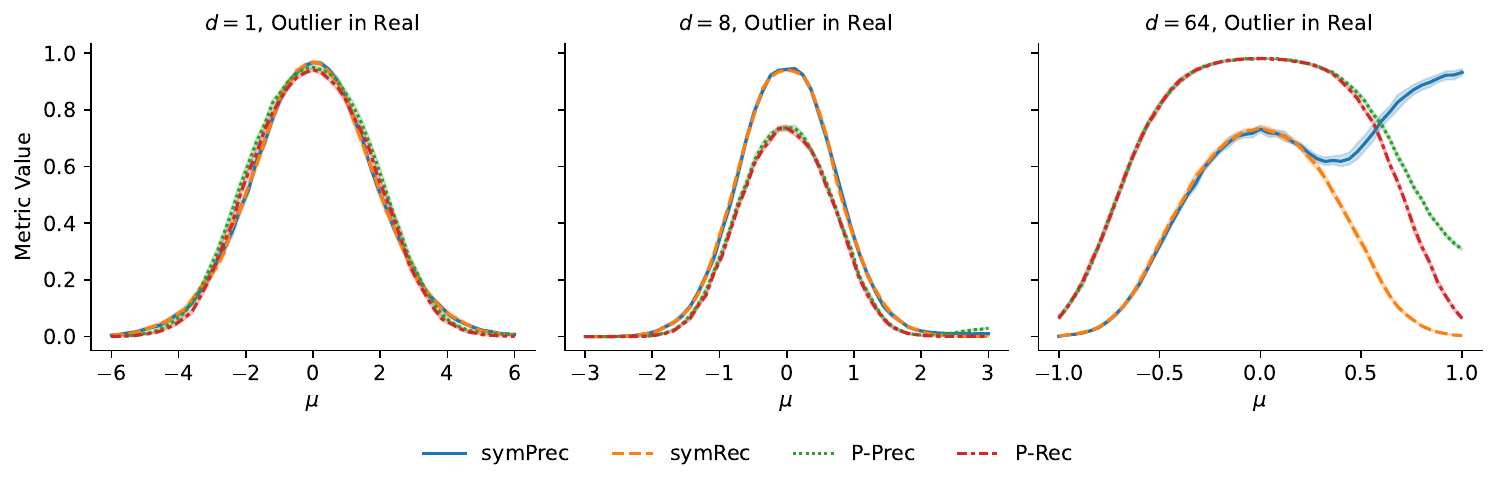}
        \caption{}
        \label{fig:gaussian-mean-difference-with-real-outlier-2}
    \end{subfigure}
    \caption{
        Gaussian mean difference + outlier (in real data) check: two $d$-dimensional Gaussian distributions with real 
        mean 0, synthetic mean $\mu$, covariance $I_d$, with an outlier
        at the largest $\mu$ displayed on the x-axis in the real data. 
    }
    \label{fig:gaussian-mean-difference-with-real-outlier}
\end{figure*}

\begin{figure*}
    \centering
    \begin{subfigure}{1.0\textwidth}
        \centering
        \includegraphics[width=\textwidth]{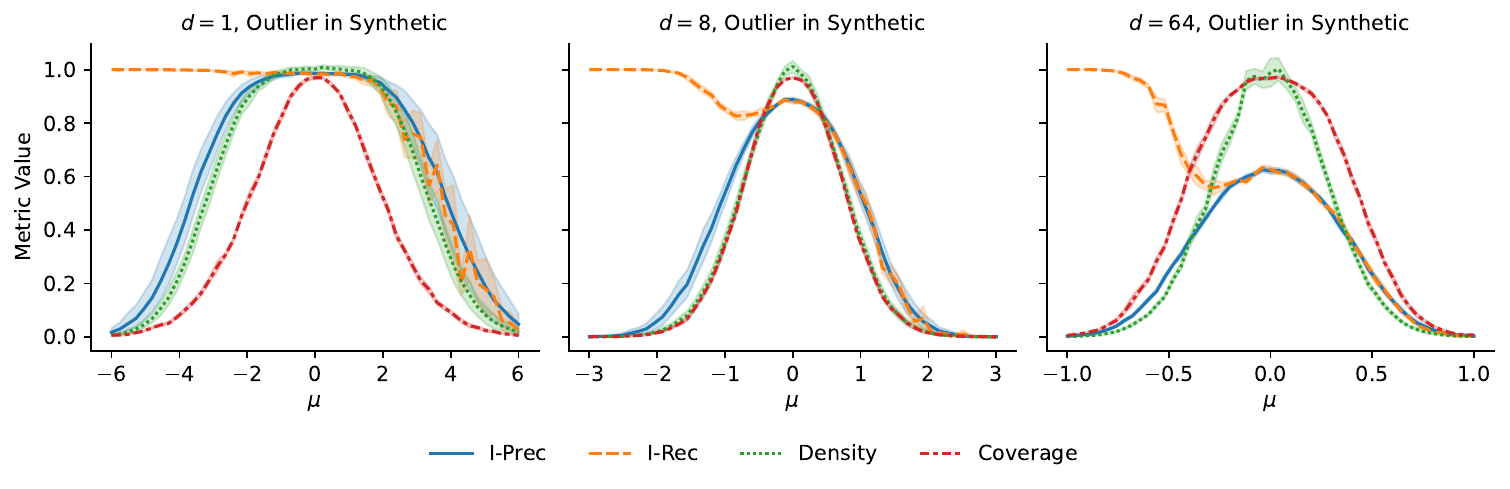}
        \caption{}
        \label{fig:gaussian-mean-difference-with-syn-outlier-0}
    \end{subfigure}
    \begin{subfigure}{1.0\textwidth}
        \centering
        \includegraphics[width=\textwidth]{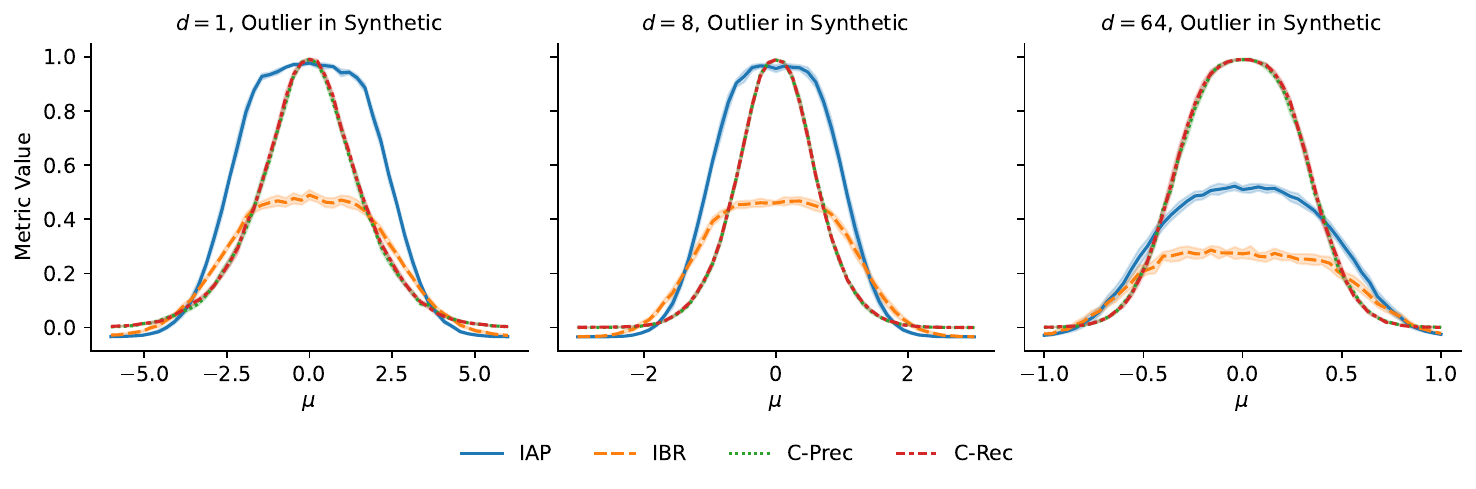}
        \caption{}
        \label{fig:gaussian-mean-difference-with-syn-outlier-1}
    \end{subfigure}
    \begin{subfigure}{1.0\textwidth}
        \centering
        \includegraphics[width=\textwidth]{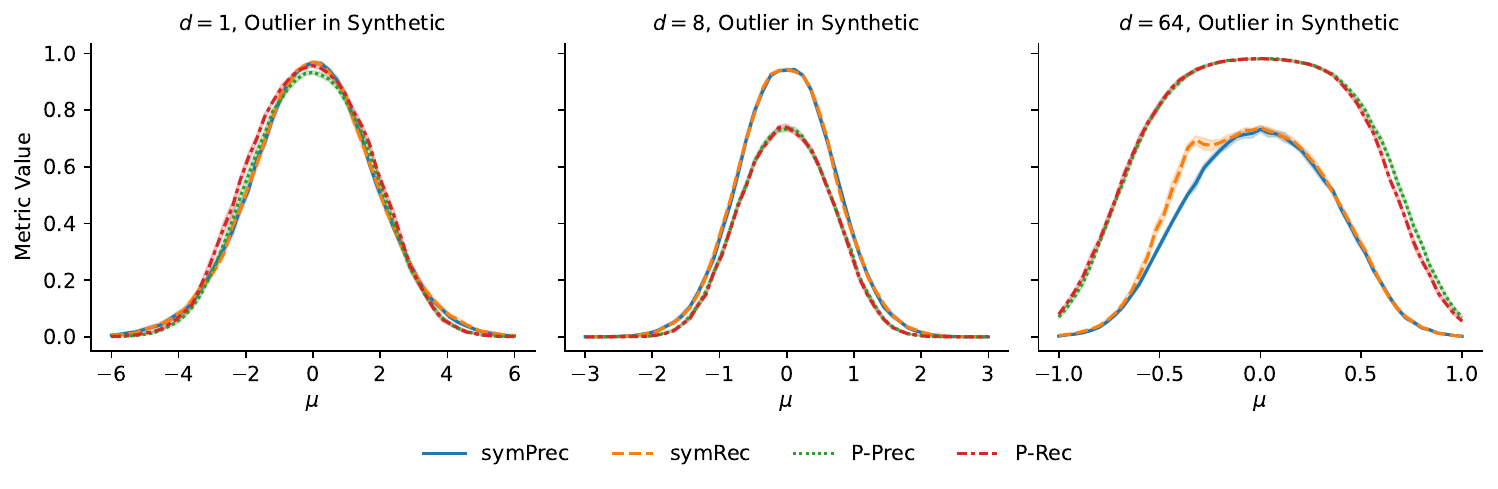}
        \caption{}
        \label{fig:gaussian-mean-difference-with-syn-outlier-2}
    \end{subfigure}
    \caption{
        Gaussian mean difference + outlier (in synthetic data) check: two $d$-dimensional Gaussian distributions with real 
        mean 0, synthetic mean $\mu$, covariance $I_d$, with an outlier
        at the largest $\mu$ displayed on the x-axis in the synthetic data. 
    }
    \label{fig:gaussian-mean-difference-with-syn-outlier}
\end{figure*}

\begin{figure*}
    \begin{subfigure}{1.0\textwidth}
        \centering
        \includegraphics[width=\textwidth]{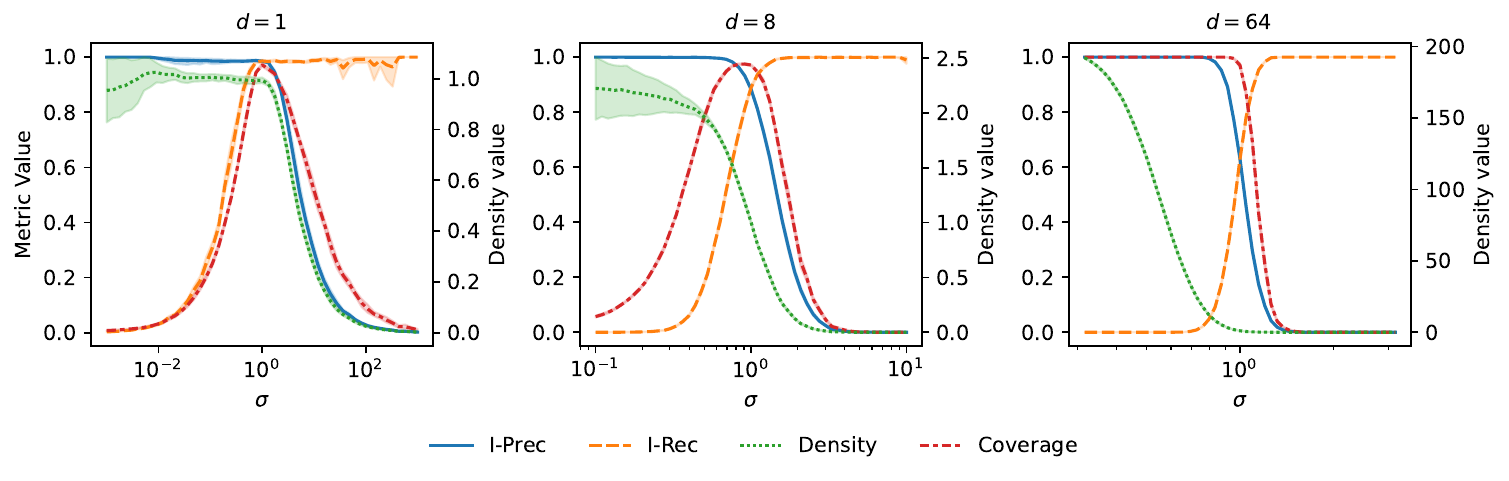}
        \caption{}
        \label{fig:gaussian-std-difference-0}
    \end{subfigure}
    \begin{subfigure}{1.0\textwidth}
        \centering
        \includegraphics[width=\textwidth]{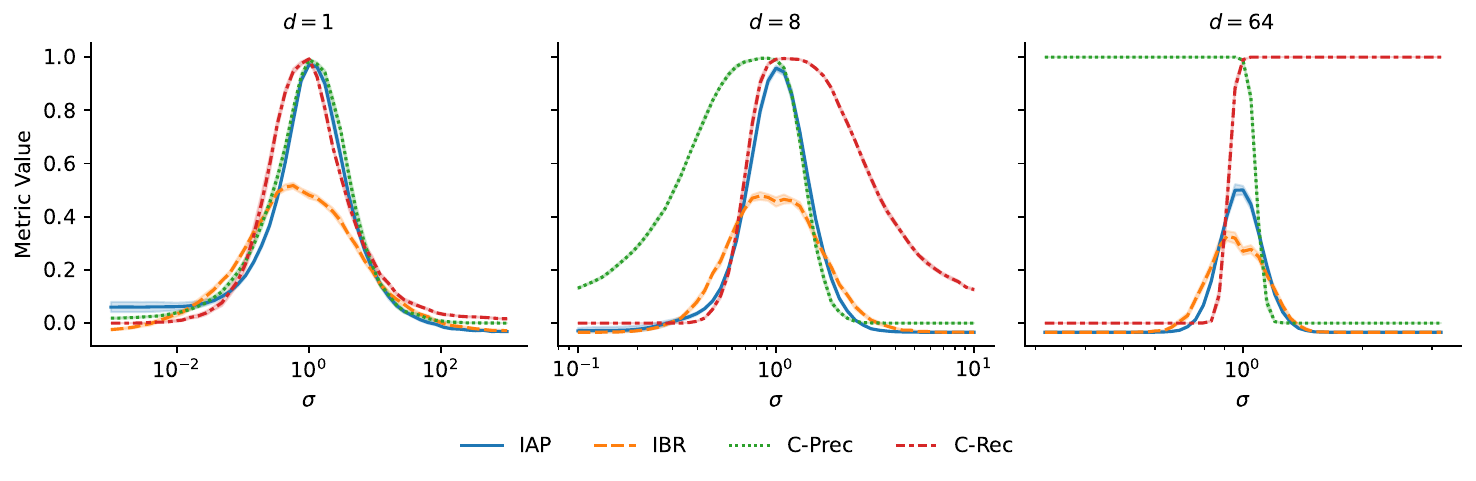}
        \caption{}
        \label{fig:gaussian-std-difference-1}
    \end{subfigure}
    \begin{subfigure}{1.0\textwidth}
        \centering
        \includegraphics[width=\textwidth]{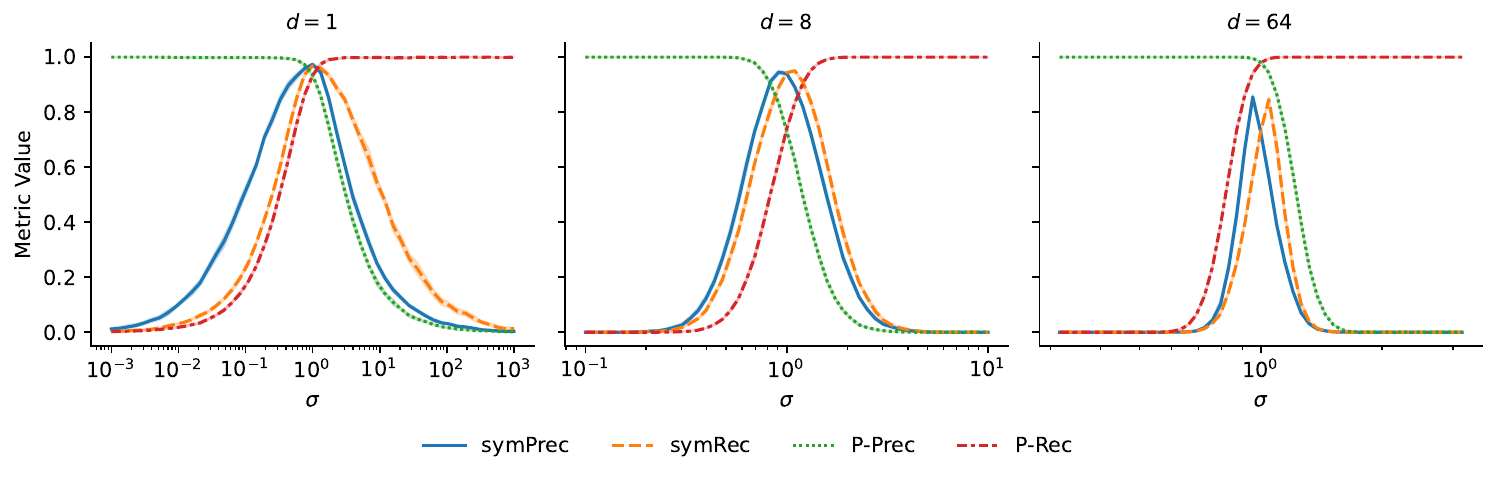}
        \caption{}
        \label{fig:gaussian-std-difference-2}
    \end{subfigure}
    \caption{
        Gaussian standard deviation difference check: two $d$-dimensional Gaussian with mean 0, real standard 
        deviation 1, and differing synthetic standard deviation. Density is 
        plotted on a separate y-axis due to having a very different range from the 
        other metrics.
    }
    \label{fig:gaussian-std-difference}
\end{figure*}

\begin{figure*}
    \begin{subfigure}{1.0\textwidth}
        \centering
        \includegraphics[width=\textwidth]{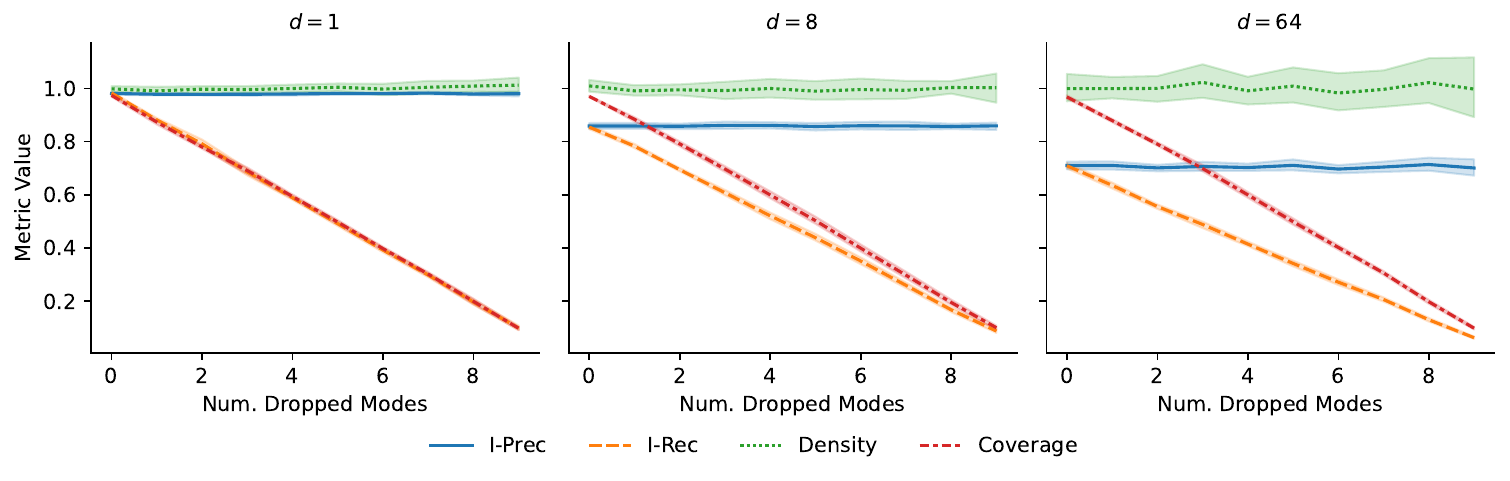}
        \caption{}
        \label{fig:mode-dropping-sequential-0}
    \end{subfigure}
    \begin{subfigure}{1.0\textwidth}
        \centering
        \includegraphics[width=\textwidth]{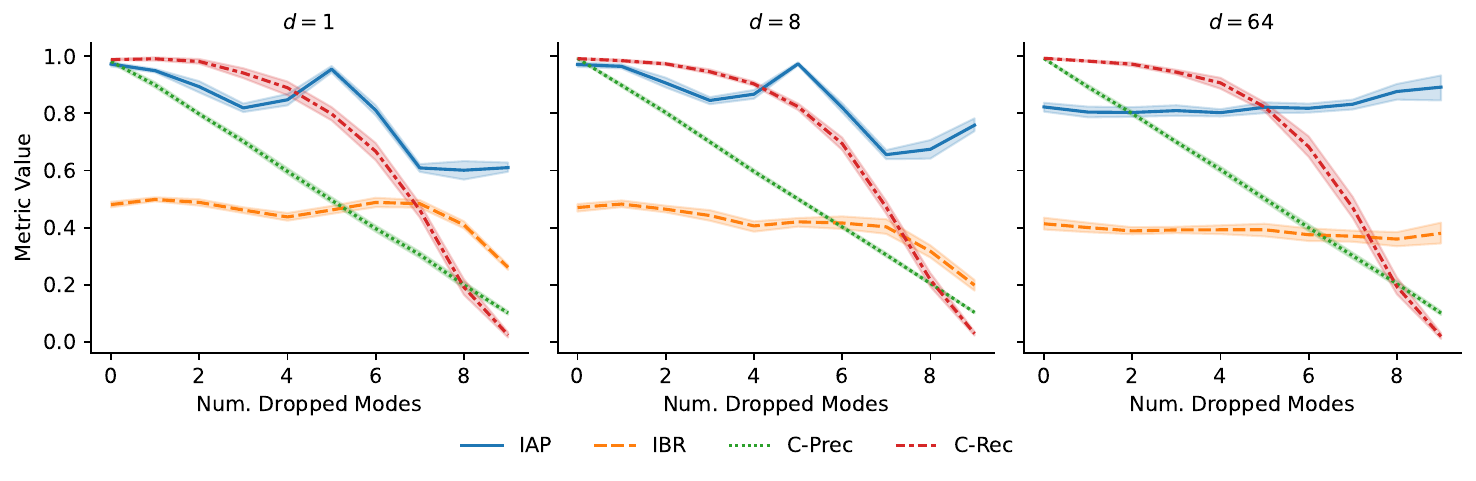}
        \caption{}
        \label{fig:mode-dropping-sequential-1}
    \end{subfigure}
    \begin{subfigure}{1.0\textwidth}
        \centering
        \includegraphics[width=\textwidth]{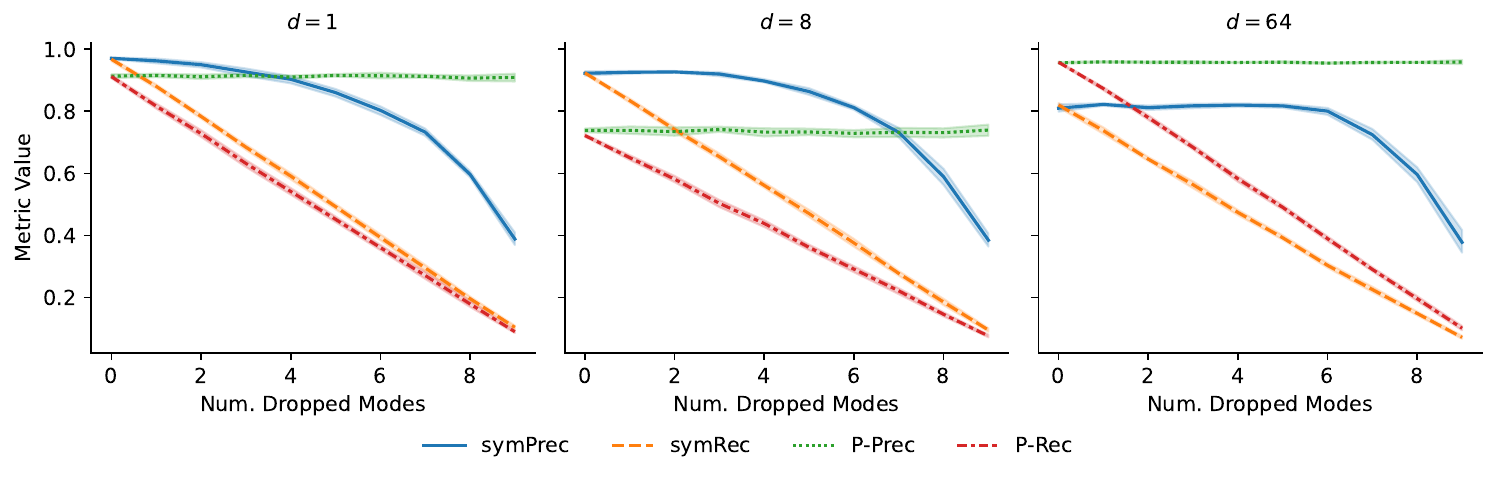}
        \caption{}
        \label{fig:mode-dropping-sequential-2}
    \end{subfigure}
    \caption{
        Sequential mode dropping check: two $d$-dimensional mixtures of Gaussians, where 
        the real data has 10 modes, and the synthetic data drops 0 to 9
        of these modes. Fidelity metrics should not drop, since the synthetic
        is always from a part of the real data distribution, but diversity 
        metrics should drop due to the dropped modes. 
    }
    \label{fig:mode-dropping-sequential}
\end{figure*}

\begin{figure*}
    \begin{subfigure}{1.0\textwidth}
        \centering
        \includegraphics[width=\textwidth]{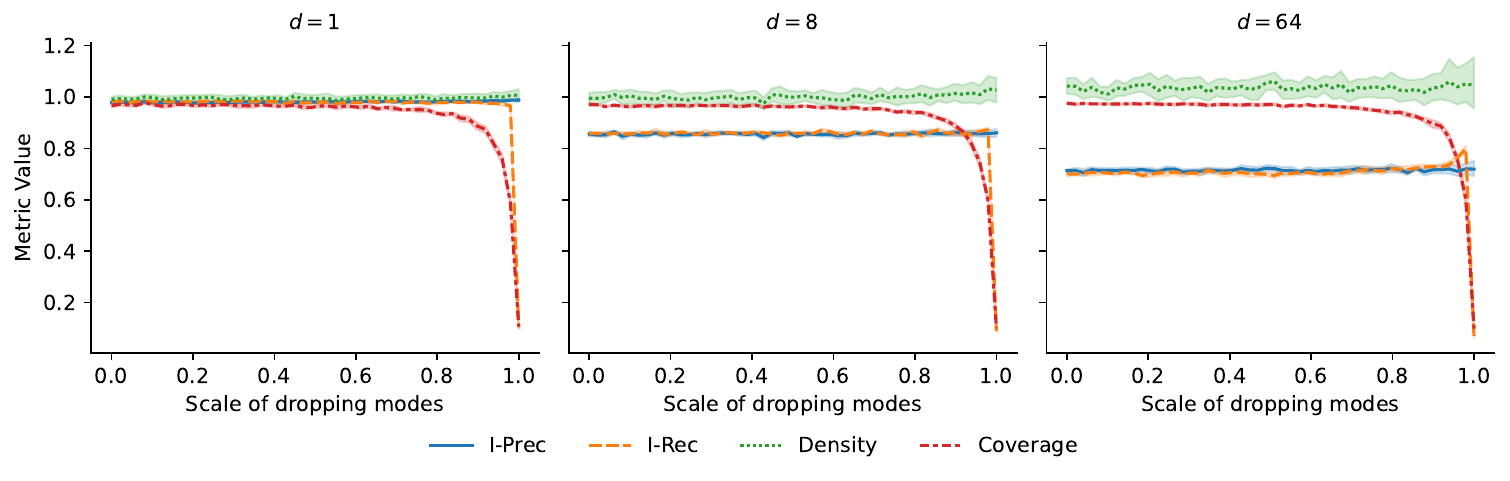}
        \caption{}
        \label{fig:mode-dropping-simultaneous-0}
    \end{subfigure}
    \begin{subfigure}{1.0\textwidth}
        \centering
        \includegraphics[width=\textwidth]{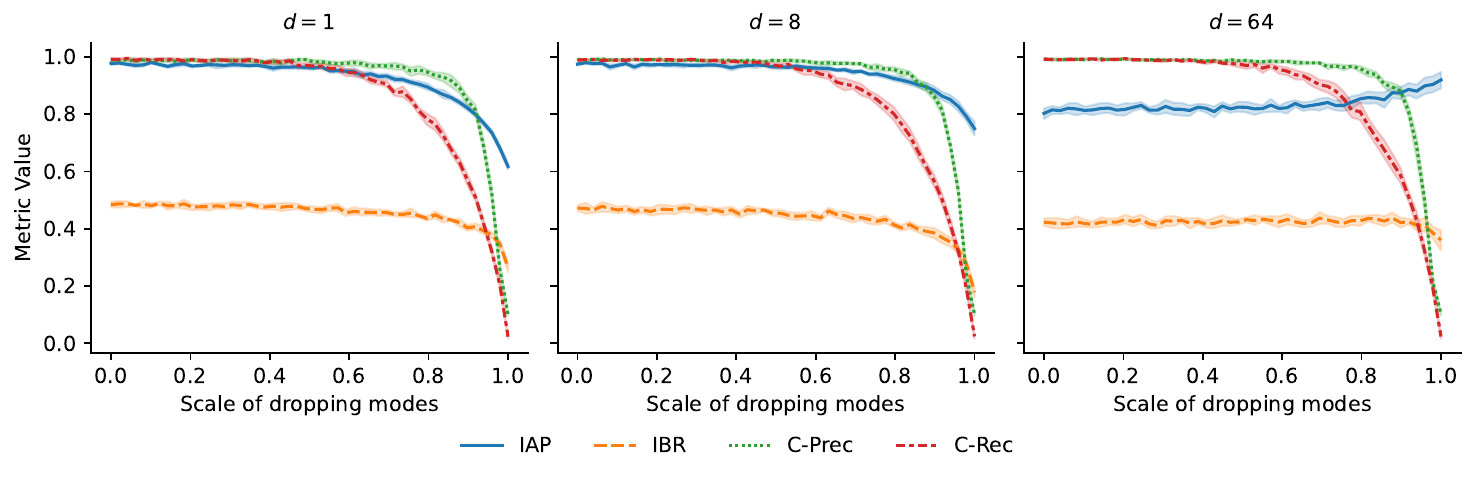}
        \caption{}
        \label{fig:mode-dropping-simultaneous-1}
    \end{subfigure}
    \begin{subfigure}{1.0\textwidth}
        \centering
        \includegraphics[width=\textwidth]{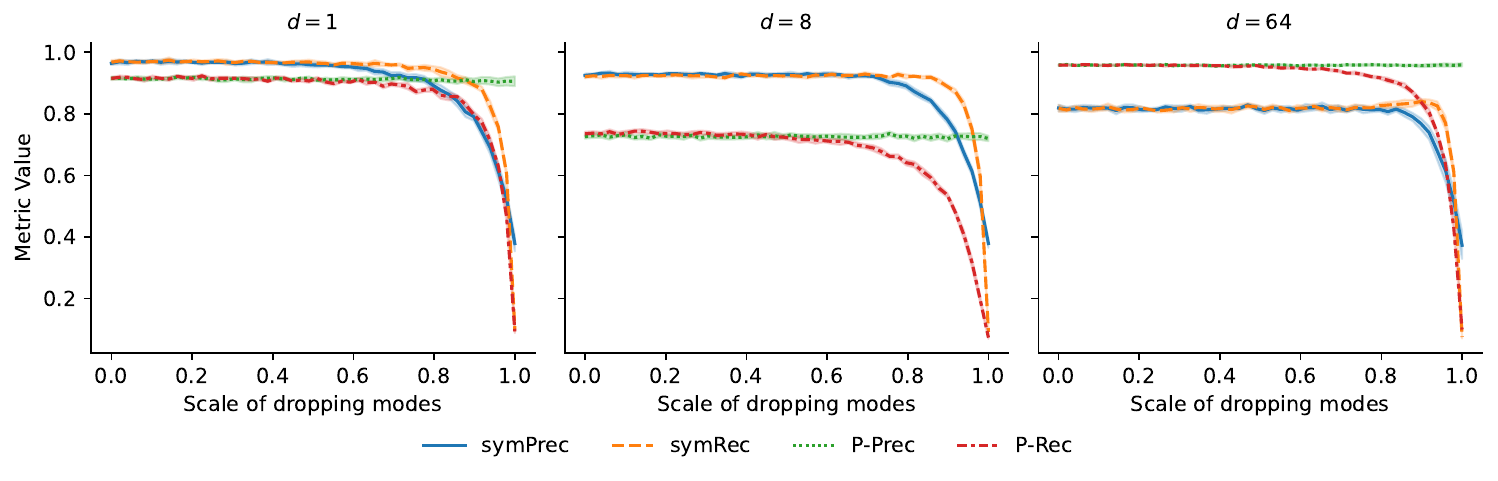}
        \caption{}
        \label{fig:mode-dropping-simultaneous-2}
    \end{subfigure}
    \caption{
        Simultaneous mode dropping check: two $d$-dimensional mixtures of Gaussians, where the 
        real data has 10 modes, and the synthetic data scales down the density 
        of 9 modes by the number on the x-axis. Fidelity metrics should not drop, 
        since the synthetic
        is always from a part of the real data distribution, but diversity 
        metrics should drop due to the dropped modes. 
    }
    \label{fig:mode-dropping-simultaneous}
\end{figure*}

\begin{figure*}
    \begin{subfigure}{0.33\textwidth}
        \centering 
        \includegraphics[width=\textwidth]{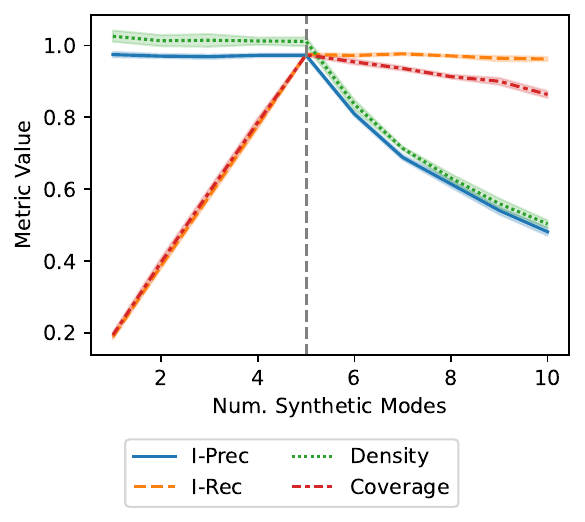}
        \caption{}
        \label{fig:mode-dropping-invention-0}
    \end{subfigure}
    \begin{subfigure}{0.33\textwidth}
        \centering 
        \includegraphics[width=\textwidth]{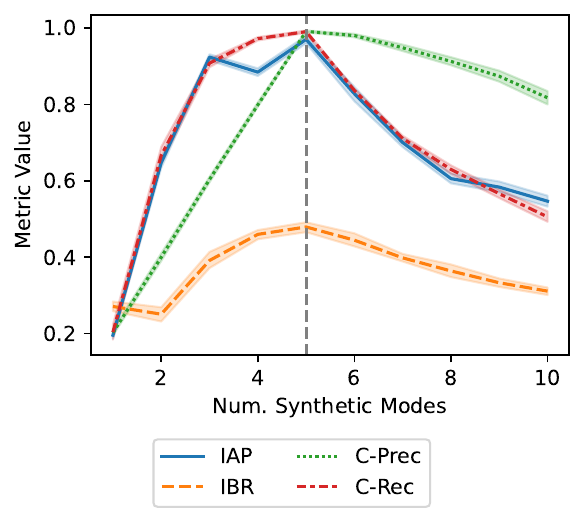}
        \caption{}
        \label{fig:mode-dropping-invention-1}
    \end{subfigure}
    \begin{subfigure}{0.33\textwidth}
        \centering 
        \includegraphics[width=\textwidth]{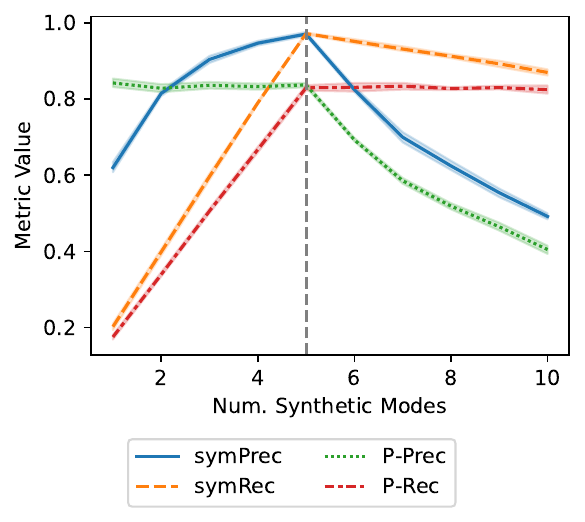}
        \caption{}
        \label{fig:mode-dropping-invention-2}
    \end{subfigure}
    \caption{
        Mode dropping + invention check: 2-dimensional mixtures of Gaussians, where the real data 
        has 5 modes, and the synthetic data has 1-10 modes. The first 
        5 are the same as the real data, and the last 5 are invented modes.
        Fidelity metrics should be close to 1 until 5 modes, and then drop, 
        while diversity metrics should increase with 1 to 5 modes, and
        be close to 1 with 5 to 10 modes. 
    }
    \label{fig:mode-dropping-invention}
\end{figure*}

\begin{figure*}
    \begin{subfigure}{1.0\textwidth}
        \centering
        \includegraphics[width=\textwidth]{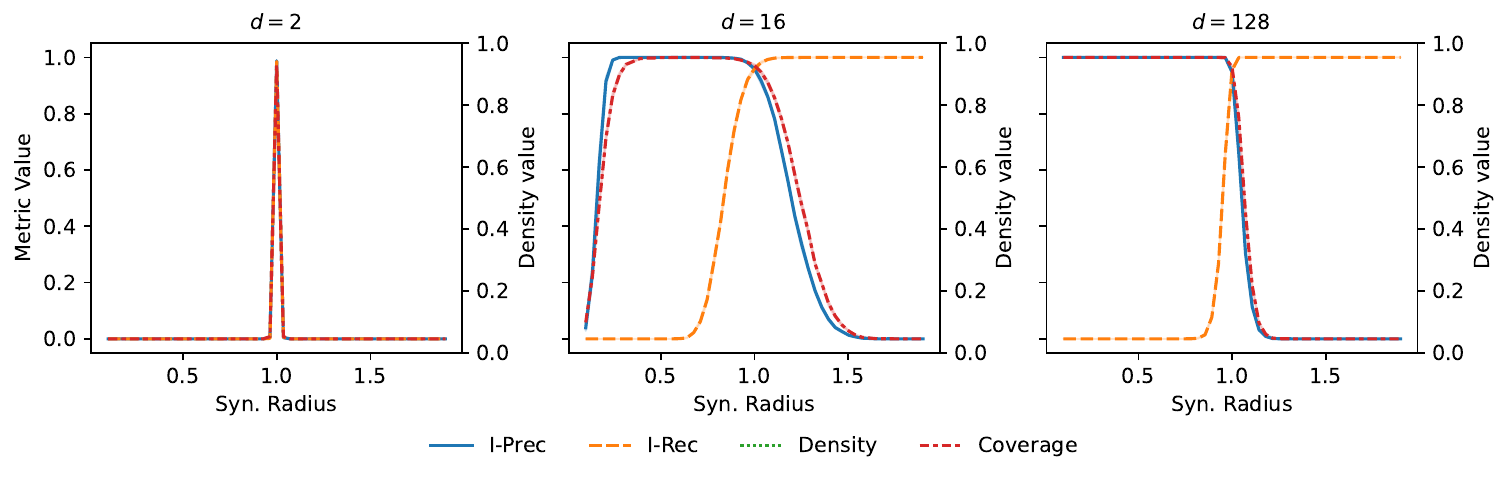}
        \caption{}
        \label{fig:uniform-hypersphere-surface-0}
    \end{subfigure}
    \begin{subfigure}{1.0\textwidth}
        \centering
        \includegraphics[width=\textwidth]{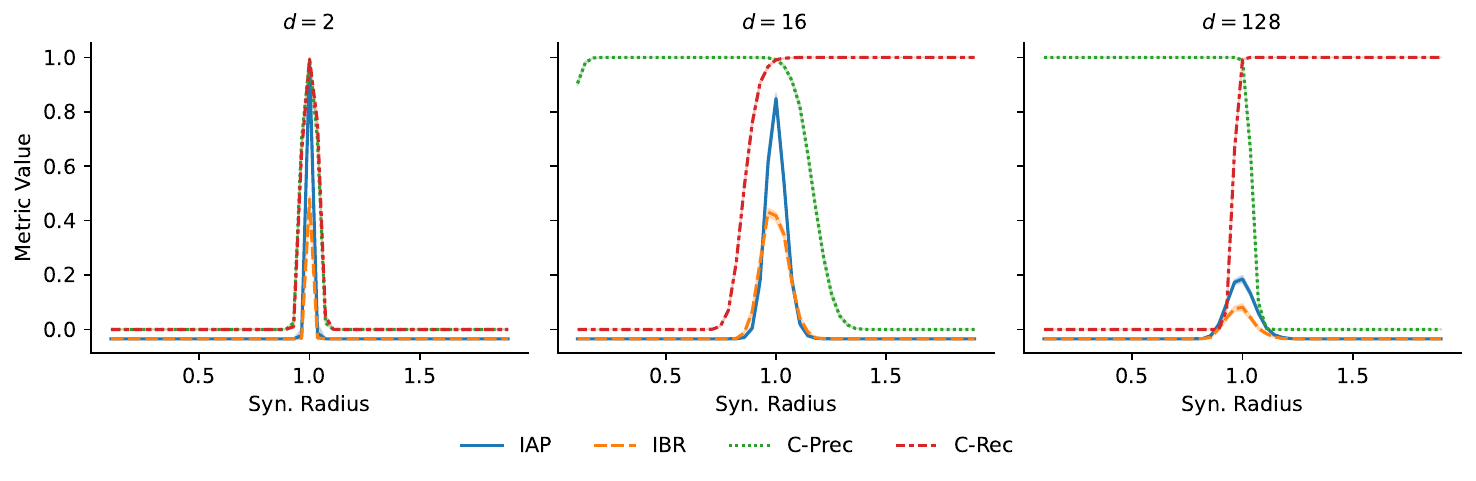}
        \caption{}
        \label{fig:uniform-hypersphere-surface-1}
    \end{subfigure}
    \begin{subfigure}{1.0\textwidth}
        \centering
        \includegraphics[width=\textwidth]{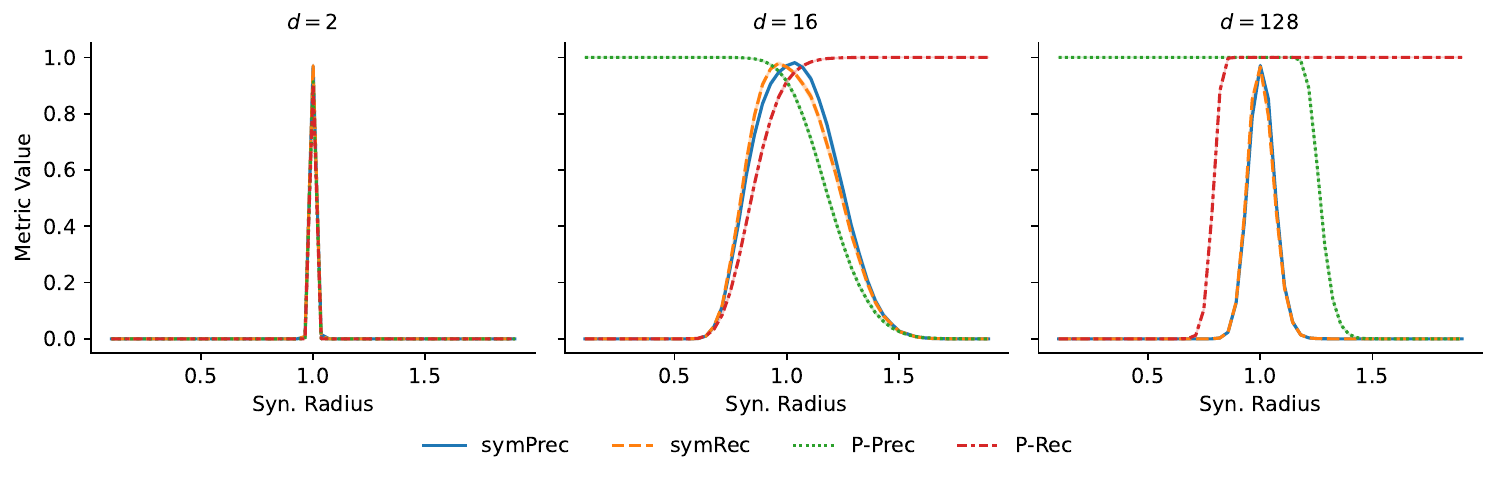}
        \caption{}
        \label{fig:uniform-hypersphere-surface-2}
    \end{subfigure}
    \caption{
        Hypersphere surface check: two uniform distributions on the surfaces
        of two hyperspheres, with radius 1 for real data and 
        the radius on the x-axis for the synthetic data. When the synthetic
        radius is not 1, the distributions are completely disjoint, so 
        all metrics should have the value 0 away from synthetic radius 1.
    }
    \label{fig:uniform-hypersphere-surface}
\end{figure*}

\begin{figure*}
    \begin{subfigure}{1.0\textwidth}
        \centering
        \includegraphics[width=\textwidth]{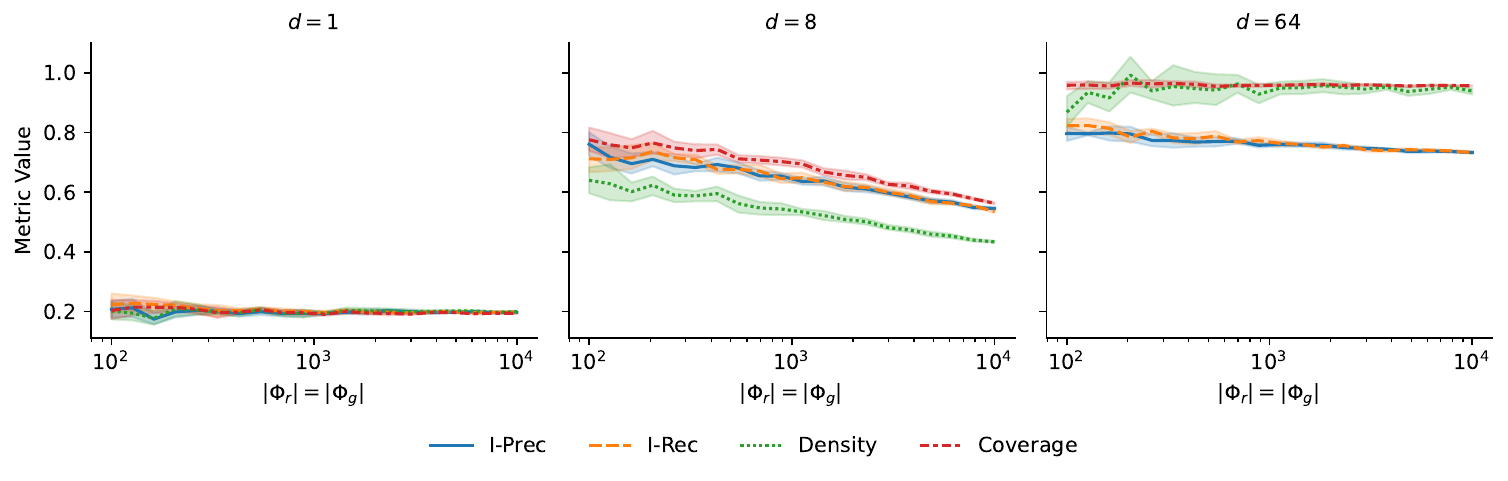}
        \caption{}
        \label{fig:uniform-hypercube-varying-dataset-size-0}
    \end{subfigure}
    \begin{subfigure}{1.0\textwidth}
        \centering
        \includegraphics[width=\textwidth]{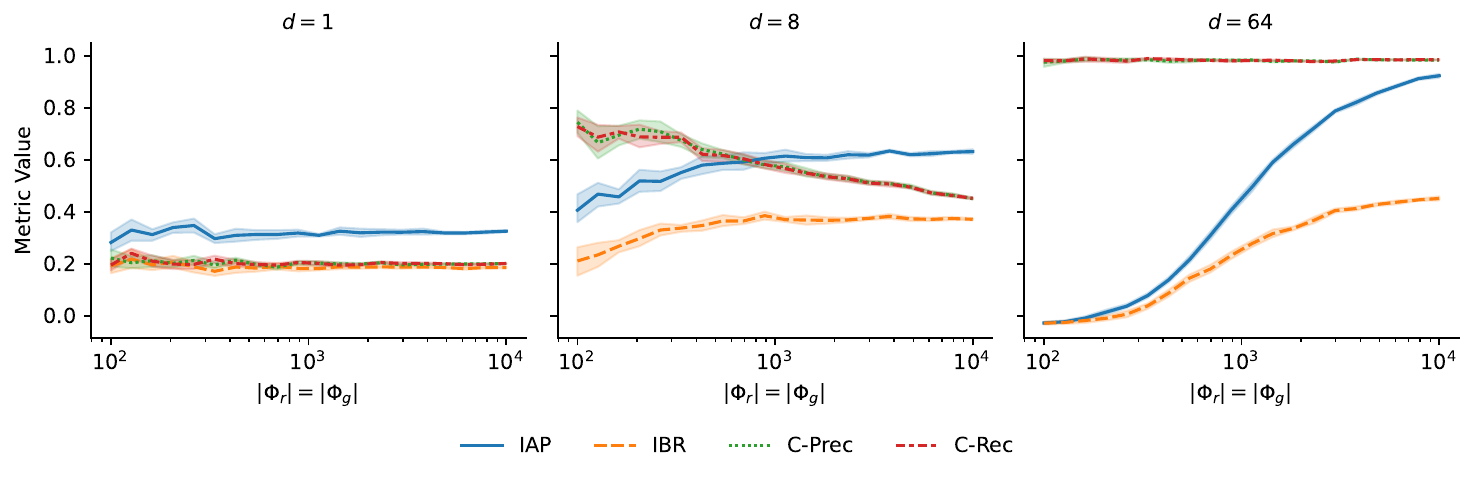}
        \caption{}
        \label{fig:uniform-hypercube-varying-dataset-size-1}
    \end{subfigure}
    \begin{subfigure}{1.0\textwidth}
        \centering
        \includegraphics[width=\textwidth]{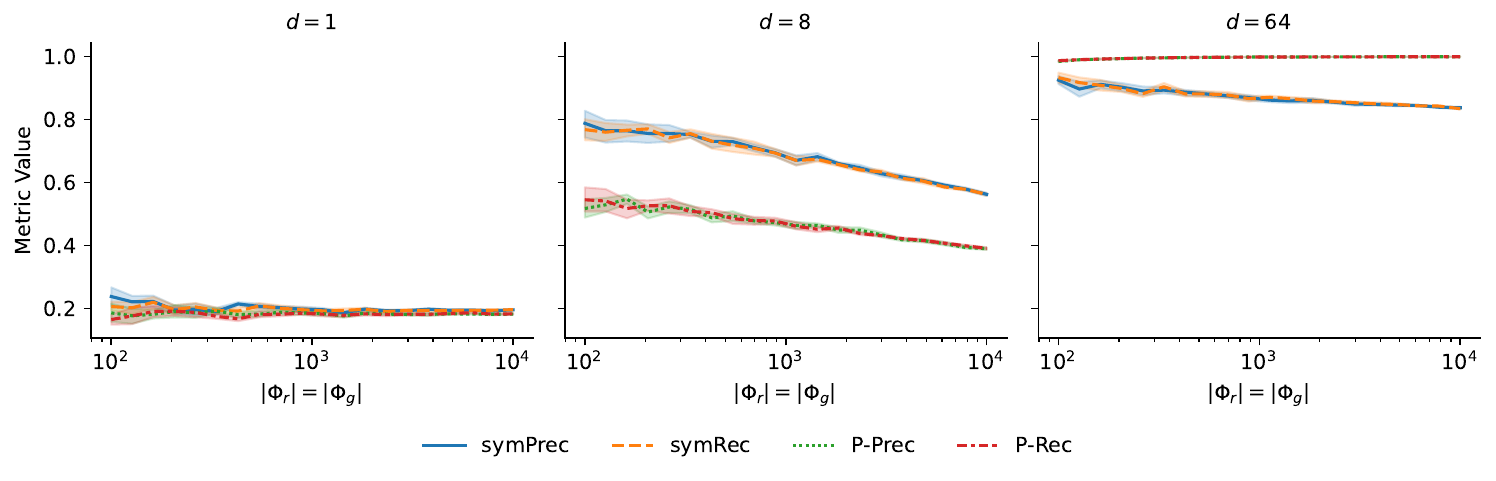}
        \caption{}
        \label{fig:uniform-hypercube-varying-dataset-size-2}
    \end{subfigure}
    \caption{
        Hypercube, varying sample size check: two uniform distributions on 
        overlapping hypercubes in $d$ dimensions,
        varying both the synthetic dataset size $|\Phi_g|$ and the real dataset
        size $|\Phi_r|$.
    }
    \label{fig:uniform-hypercube-varying-dataset-size}
\end{figure*}

\begin{figure*}
    \begin{subfigure}{1.0\textwidth}
        \centering
        \includegraphics[width=\textwidth]{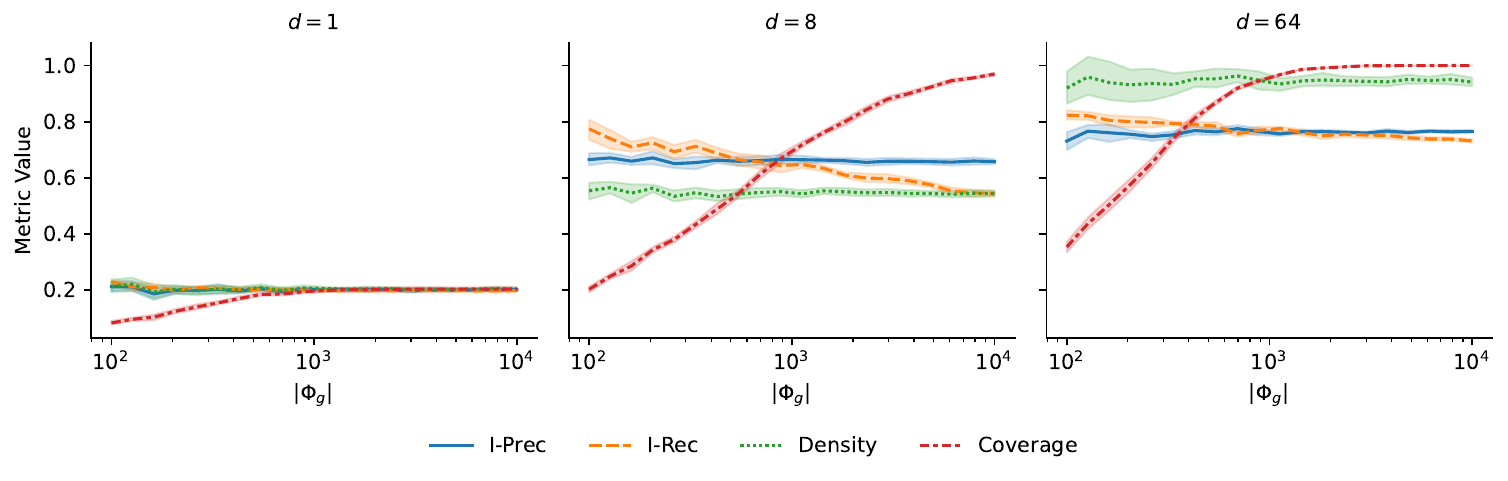}
        \caption{}
        \label{fig:uniform-hypercube-varying-syn-dataset-size-0}
    \end{subfigure}
    \begin{subfigure}{1.0\textwidth}
        \centering
        \includegraphics[width=\textwidth]{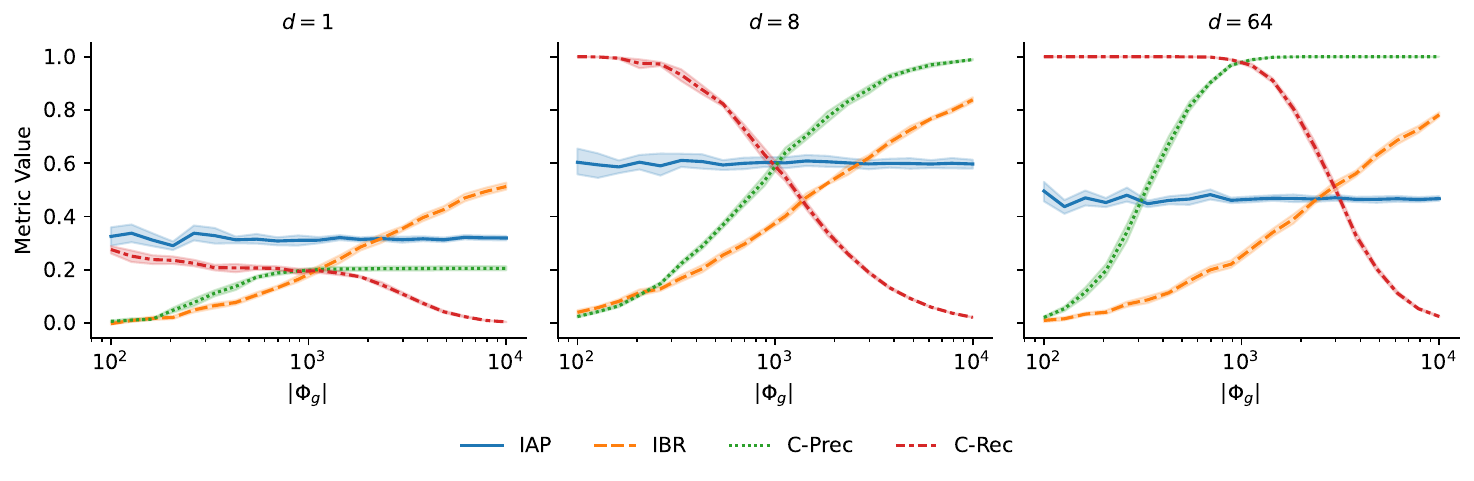}
        \caption{}
        \label{fig:uniform-hypercube-varying-syn-dataset-size-1}
    \end{subfigure}
    \begin{subfigure}{1.0\textwidth}
        \centering
        \includegraphics[width=\textwidth]{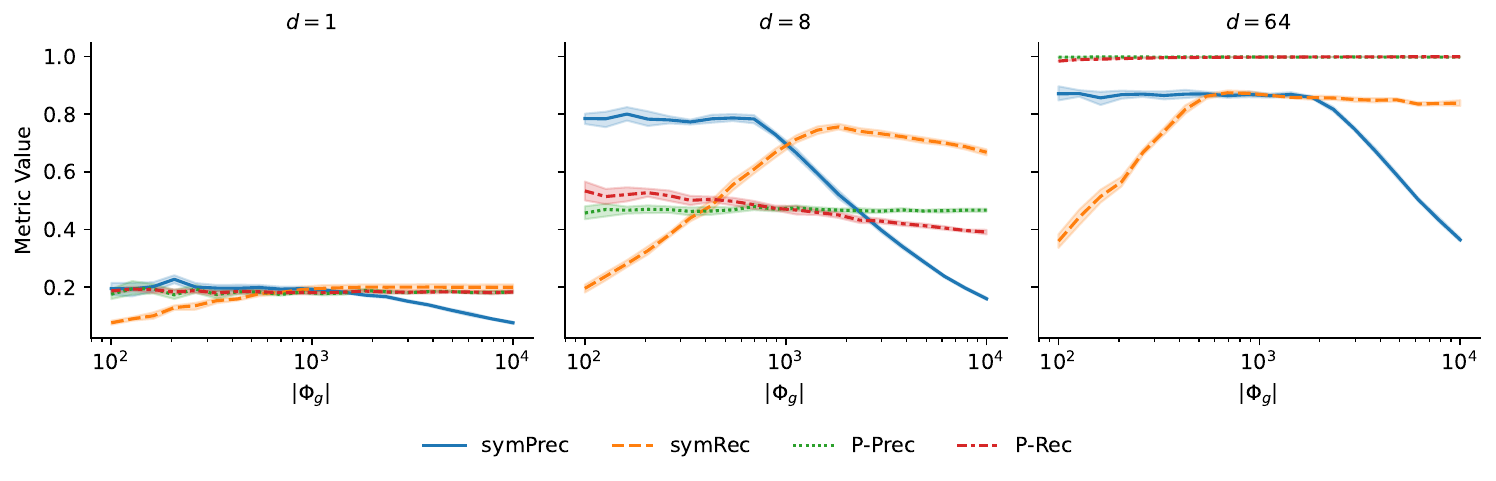}
        \caption{}
        \label{fig:uniform-hypercube-varying-syn-dataset-size-2}
    \end{subfigure}
    \caption{
        Hypercube, varying synthetic size check: two uniform distributions on overlapping 
        hypercubes in $d$-dimensions, varying the synthetic 
        dataset size $|\Phi_g|$, while the real dataset size if fixed at
        $|\Phi_r| = 1000$.
    }
    \label{fig:uniform-hypercube-varying-syn-dataset-size}
\end{figure*}

\begin{figure*}
    \begin{subfigure}{0.32\textwidth}
        \centering
        \includegraphics[width=\textwidth]{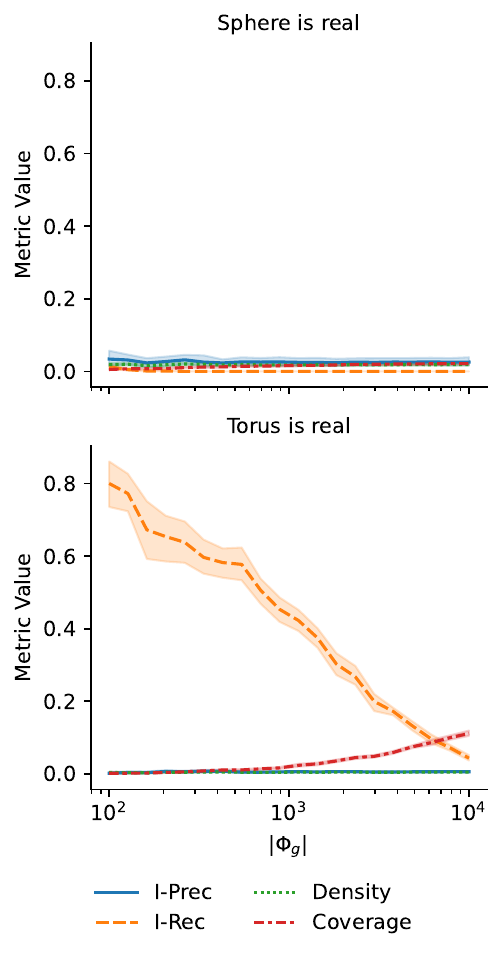}
        \caption{}
        \label{fig:sphere-torus-0}
    \end{subfigure}
    \begin{subfigure}{0.32\textwidth}
        \centering
        \includegraphics[width=\textwidth]{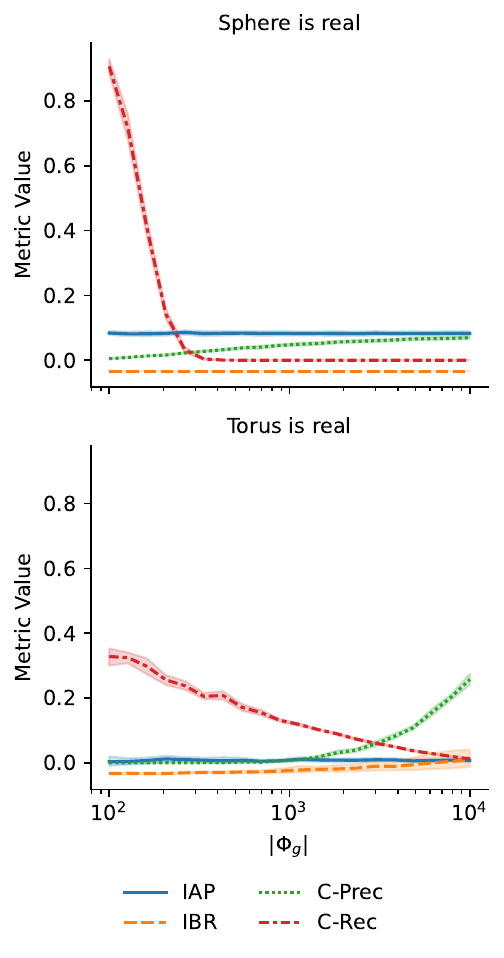}
        \caption{}
        \label{fig:sphere-torus-1}
    \end{subfigure}
    \begin{subfigure}{0.32\textwidth}
        \centering
        \includegraphics[width=\textwidth]{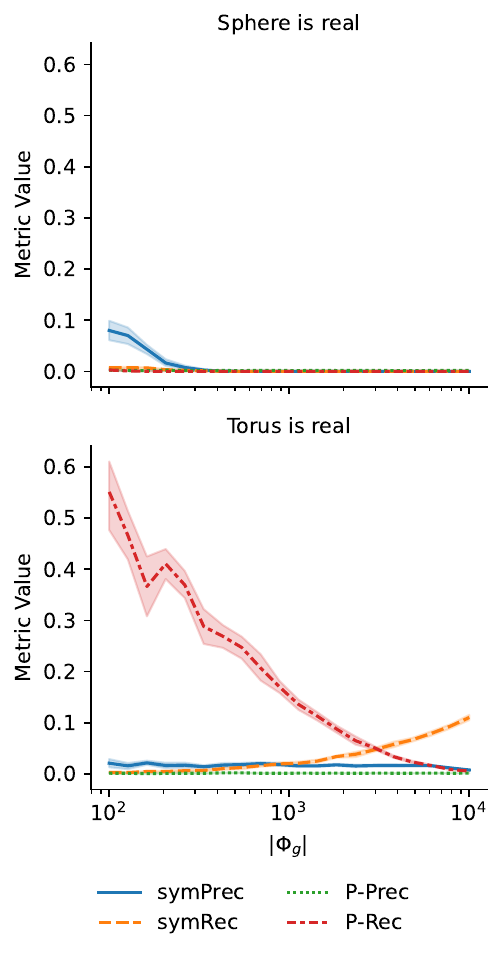}
        \caption{}
        \label{fig:sphere-torus-2}
    \end{subfigure}
    \caption{
        Sphere vs. torus check: distributions on a sphere and a disjoint torus
        surrounding the sphere, with varying synthetic dataset
        size. The real dataset size is fixed at $|\Phi_r| = 1000$.
    }
    \label{fig:sphere-torus}
\end{figure*}

\begin{figure*}
    \begin{subfigure}{1.0\textwidth}
        \centering
        \includegraphics[width=\textwidth]{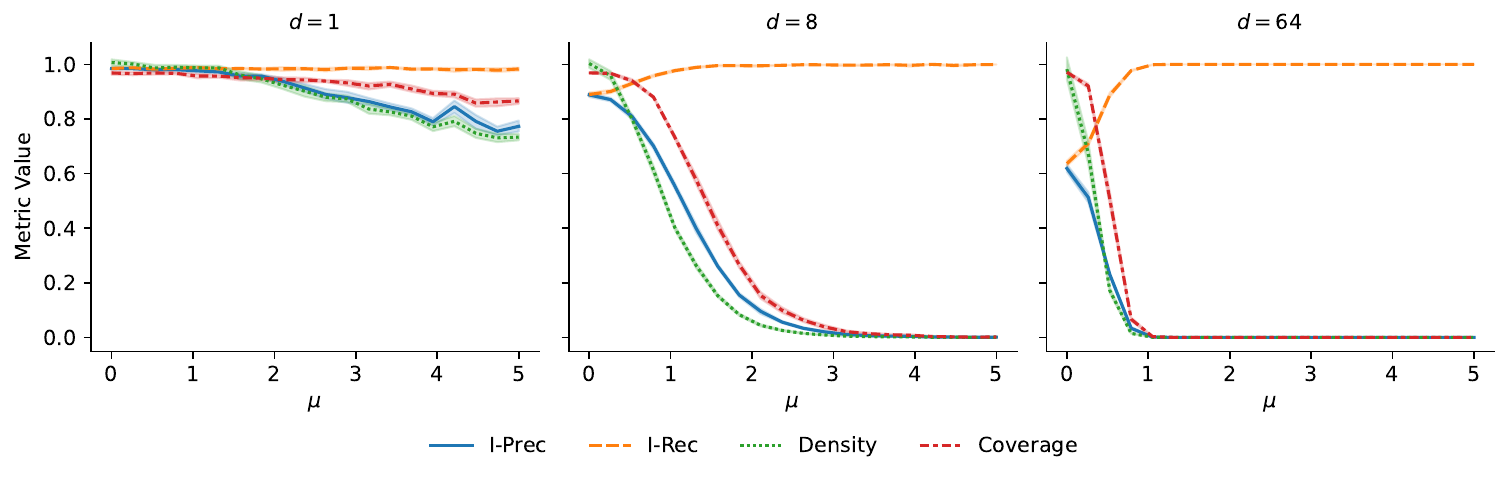}
        \caption{}
        \label{fig:one-vs-two-modes-0}
    \end{subfigure}
    \begin{subfigure}{1.0\textwidth}
        \centering
        \includegraphics[width=\textwidth]{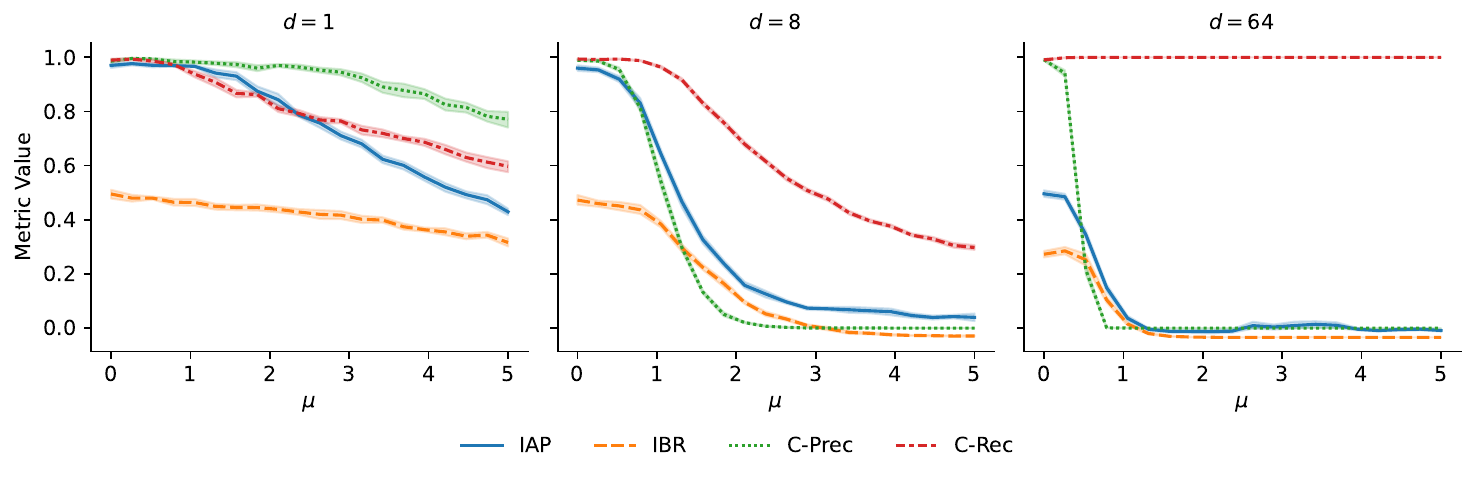}
        \caption{}
        \label{fig:one-vs-two-modes-1}
    \end{subfigure}
    \begin{subfigure}{1.0\textwidth}
        \centering
        \includegraphics[width=\textwidth]{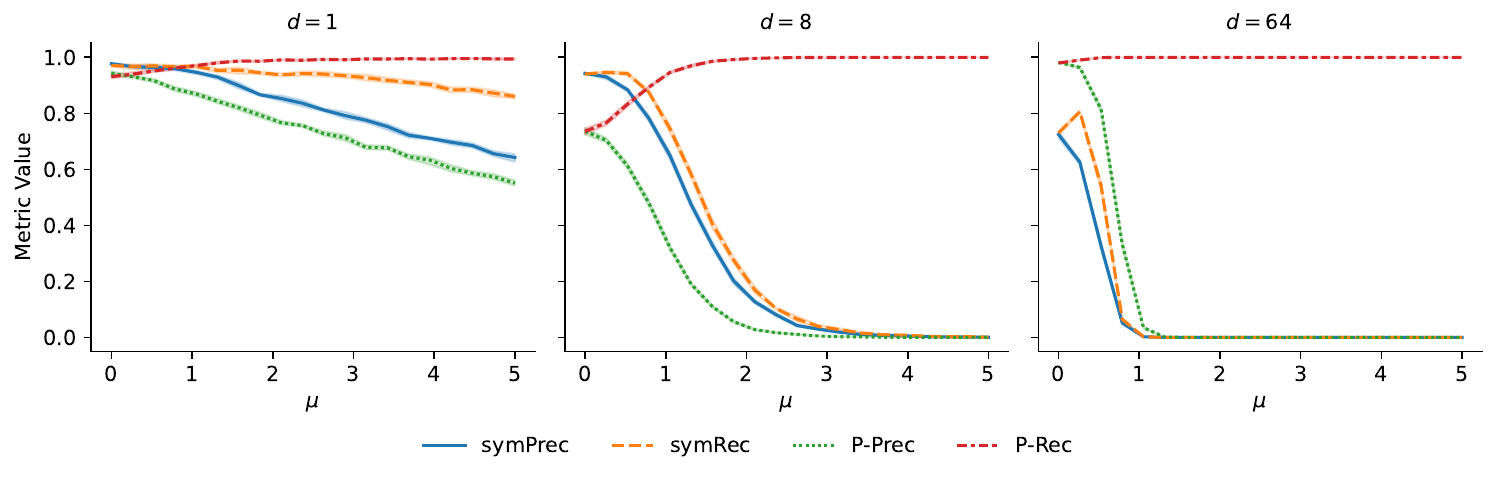}
        \caption{}
        \label{fig:one-vs-two-modes-2}
    \end{subfigure}
    \caption{
        Mode collapse check: the real distribution being a two-component 
        Gaussian mixture, and the synthetic distribution is a 
        Gaussian that covers the real distribution. $\mu$ is the 
        separation between the two components of the real distribution.
    }
    \label{fig:one-vs-two-modes}
\end{figure*}

\begin{figure*}
    \begin{subfigure}{0.32\textwidth}
        \centering
        \includegraphics[width=\textwidth]{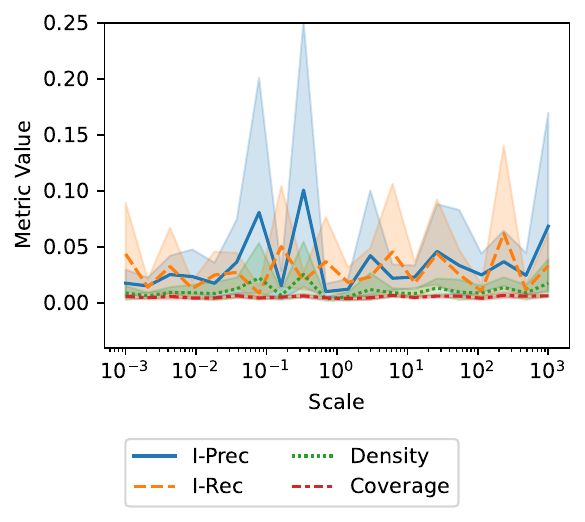}
        \caption{}
        \label{fig:gaussian-scaling-one-dimension-0}
    \end{subfigure}
    \begin{subfigure}{0.32\textwidth}
        \centering
        \includegraphics[width=\textwidth]{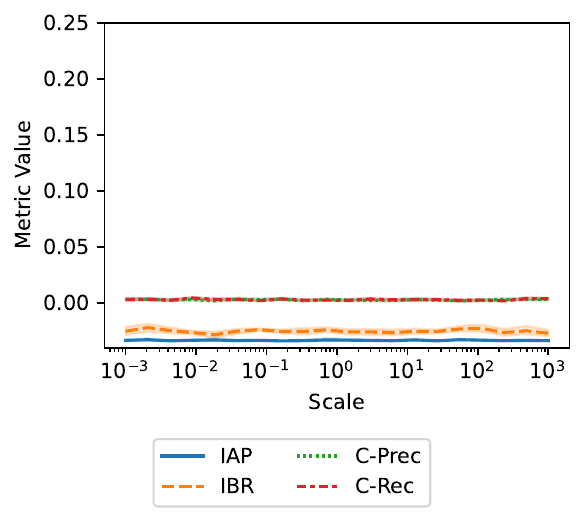}
        \caption{}
        \label{fig:gaussian-scaling-one-dimension-1}
    \end{subfigure}
    \begin{subfigure}{0.32\textwidth}
        \centering
        \includegraphics[width=\textwidth]{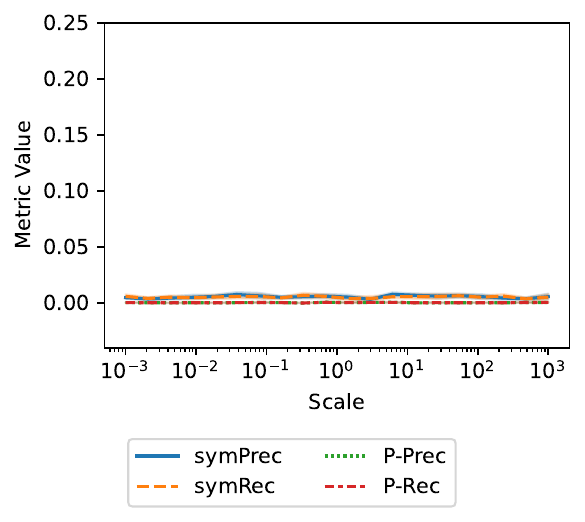}
        \caption{}
        \label{fig:gaussian-scaling-one-dimension-2}
    \end{subfigure}
    \caption{
        Scaling one dimension check: two 2-dimensional Gaussian distributions
        that are almost disjoint on the first dimension, 
        identical on the second dimension, and the second 
        dimension is scaled by a varying multiplier.
        Negative values of IAP and IBR are due to numerical
        integration errors.
    }
    \label{fig:gaussian-scaling-one-dimension}
\end{figure*}

\begin{figure*}
    \begin{subfigure}{0.32\textwidth}
        \centering
        \includegraphics[width=\textwidth]{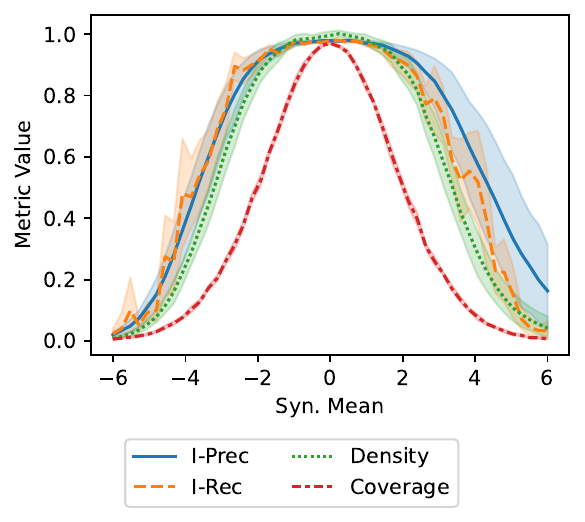}
        \caption{}
        \label{fig:gaussian-mean-difference-with-pareto-0}
    \end{subfigure}
    \begin{subfigure}{0.32\textwidth}
        \centering
        \includegraphics[width=\textwidth]{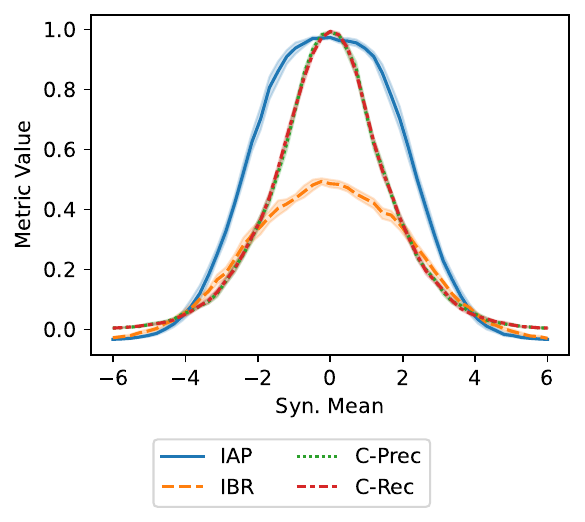}
        \caption{}
        \label{fig:gaussian-mean-difference-with-pareto-1}
    \end{subfigure}
    \begin{subfigure}{0.32\textwidth}
        \centering
        \includegraphics[width=\textwidth]{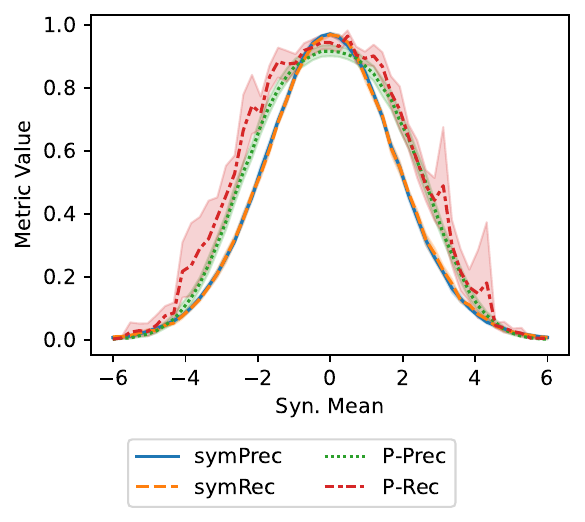}
        \caption{}
        \label{fig:gaussian-mean-difference-with-pareto-2}
    \end{subfigure}
    \caption{
        Gaussian mean difference + Pareto check: two-dimensional data, where the first 
        dimension is Gaussian with mean different means 
        in the real and synthetic data, and the second
        dimension is a Pareto distribution that is identical
        between real and synthetic data.
    }
    \label{fig:gaussian-mean-difference-with-pareto}
\end{figure*}

\begin{figure*}
    \begin{subfigure}{0.32\textwidth}
        \centering
        \includegraphics[width=\textwidth]{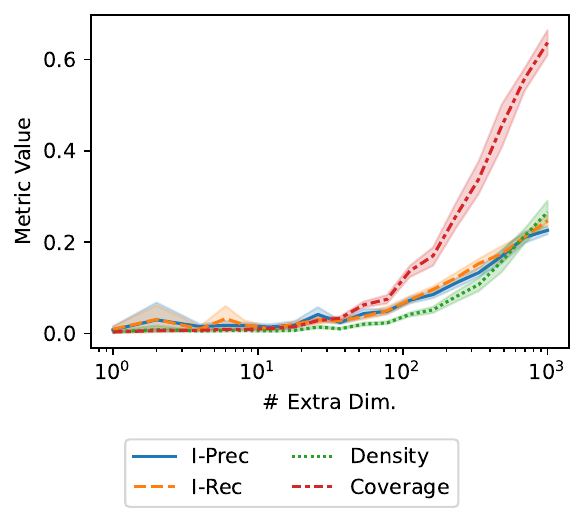}
        \caption{}
        \label{fig:gaussian-high-dim-one-disjoint-dim-0}
    \end{subfigure}
    \begin{subfigure}{0.32\textwidth}
        \centering
        \includegraphics[width=\textwidth]{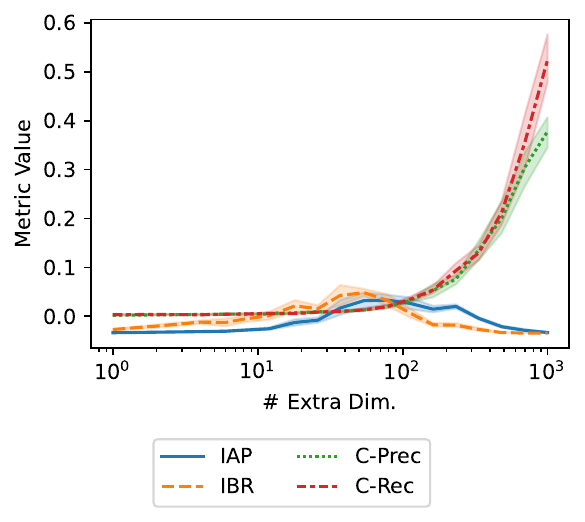}
        \caption{}
        \label{fig:gaussian-high-dim-one-disjoint-dim-1}
    \end{subfigure}
    \begin{subfigure}{0.32\textwidth}
        \centering
        \includegraphics[width=\textwidth]{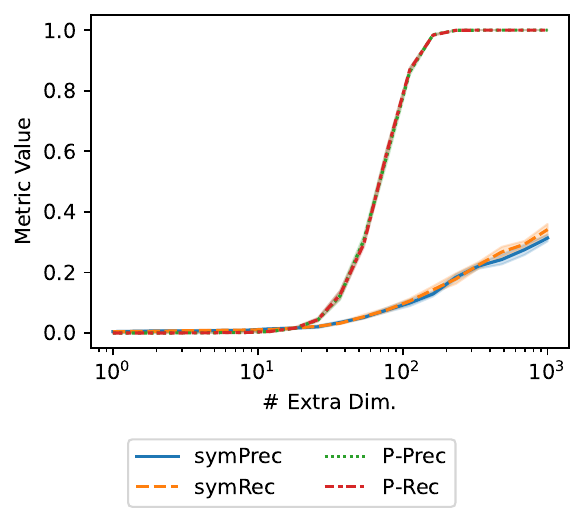}
        \caption{}
        \label{fig:gaussian-high-dim-one-disjoint-dim-2}
    \end{subfigure}
    \caption{
        One disjoint dimension + many identical dimensions check: two Gaussian distributions that are 
        almost disjoint in the first dimensions and 
        identical in the rest of the dimensions, with 
        varying number of the identical extra dimensions.
    }
    \label{fig:gaussian-high-dim-one-disjoint-dim}
\end{figure*}

\begin{figure*}
    \begin{subfigure}{0.32\textwidth}
        \centering
        \includegraphics[width=\textwidth]{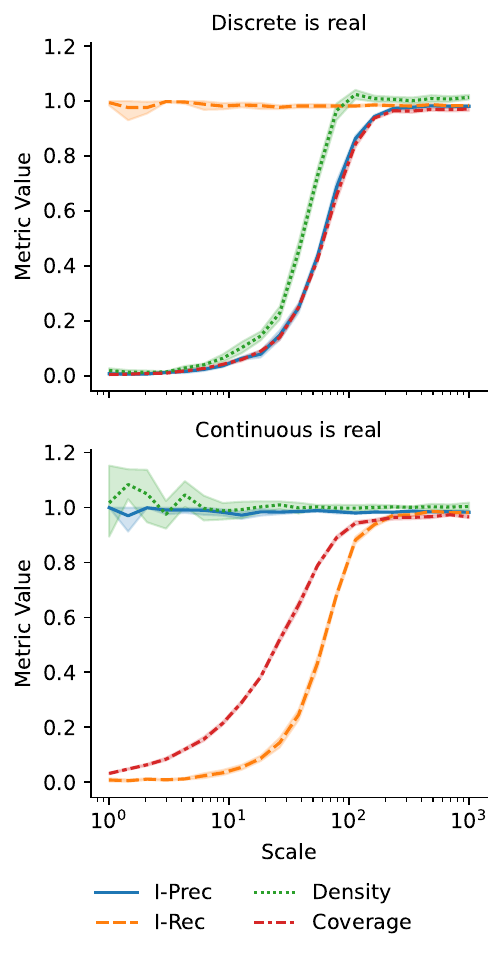}
        \caption{}
        \label{fig:discrete-numerical-vs-continuous-numerical-0}
    \end{subfigure}
    \begin{subfigure}{0.32\textwidth}
        \centering
        \includegraphics[width=\textwidth]{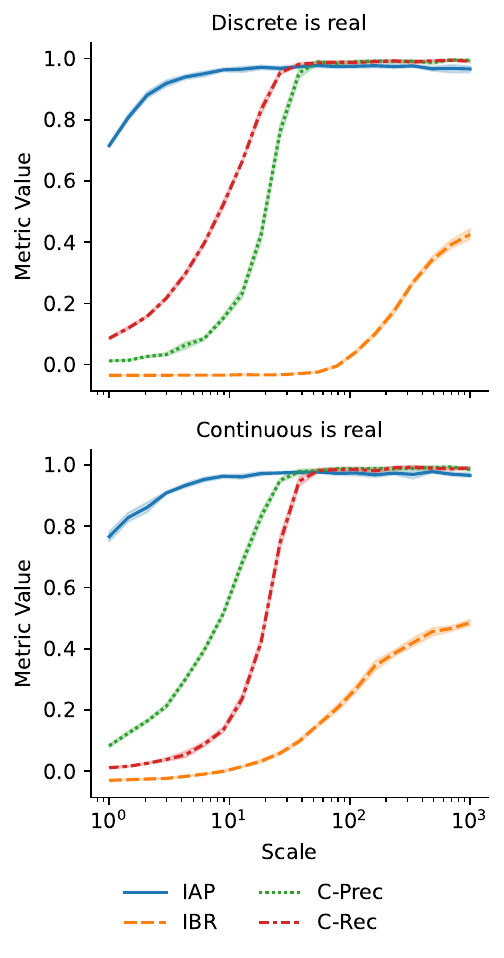}
        \caption{}
        \label{fig:discrete-numerical-vs-continuous-numerical-1}
    \end{subfigure}
    \begin{subfigure}{0.32\textwidth}
        \centering
        \includegraphics[width=\textwidth]{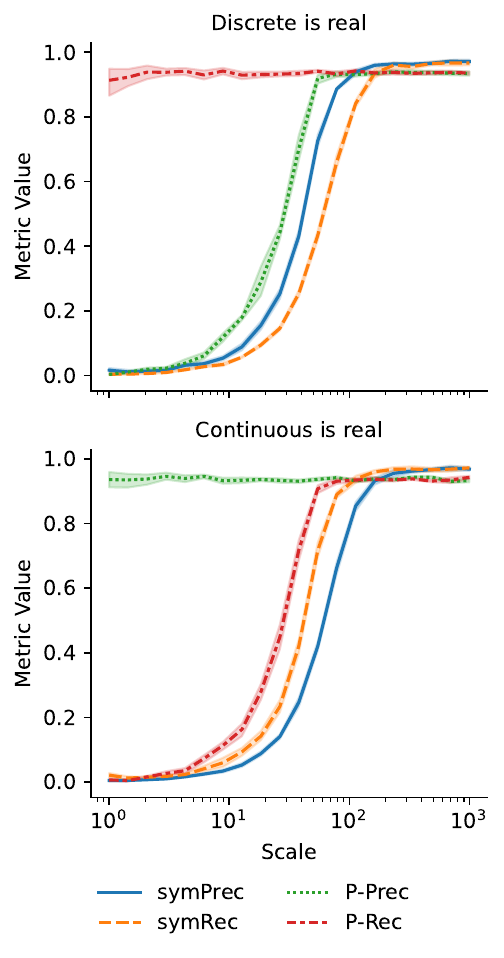}
        \caption{}
        \label{fig:discrete-numerical-vs-continuous-numerical-2}
    \end{subfigure}
    \caption{
        Discrete numerical vs. continuous numerical check: two identical scaled Gaussian distributions, 
        one of which is discretised by rounding to 
        integers. 
    }
    \label{fig:discrete-numerical-vs-continuous-numerical}
\end{figure*}

\end{document}